\def\G{\Gamma}
\def\sg{\partial}
\def\ul{\underline}
\def\ignore#1{}
\def\remark#1{\smskip\pn{\bf Remark #1}\quad} 

\newcount\sectnum
\newcount\subsectnum
\newcount\eqnumber

\global\eqnumber=1\sectnum=0


\def\lab{(\the\sectnum.\the\eqnumber)}



\def\show#1{#1}



\def\magnify{\magnification=1200}

\def\smskip{\vskip 5 pt}
\def\medskip{\vskip 10 pt}
\def\bigskip{\vskip 15 pt}
\def\pn{\par\noindent}
\def\br{\break}
\def\un{\underline}
\def\ov{\overline}
\def\Bl{\Bigl}
\def\Br{\Bigr}
\def\bl{\bigl} 
\def\br{\bigr} 
\def\lf{\left}
\def\ri{\right}

\def\argmin{\mathop{\arg \min}}
\def\tendsd{\downarrow}
\def\mod{\rm mod}
\def\tends{\rightarrow}
\def\thereis{\exists}
\def\implies{\Rightarrow}
\def\implied{\Leftarrow}
\def\frac#1#2{{#1\over #2}}

\def\ubar{\underline}
\def\ol#1{\overline{#1}}
\def\kth{$k^{\rm th}$ }
\def\ith{$i^{\rm th}$ }
\def\jth{$j^{\rm th}$ }
\def\lth{$\ell^{\rm th}$ }
\def\grad{\nabla}
\def\tendsd{\downarrow}
\def\tr{ ^{\prime}}
\def\half{{\scriptstyle {1\over 2}}}
\def\Ascr{{\cal A}}
\def\Fscr{{\cal F}}
\def\Pscr{{\cal P}}
\def\Zscr{{\cal Z}}
\def\Nscr{{\cal N}}
\def\Cscr{{\cal C}}

\def\a{\alpha}
\def\be{\beta}
\def\b{\beta}
\def\l{\lambda}
\def\g{\gamma}
\def\m{\mu}
\def\p{\pi}
\def\r{\rho}
\def\e{\epsilon}
\def\t{\tau}
\def\d{\delta}
\def\s{\sigma}
\def\f{\phi}
\def\o{\omega}
\def\D{\Delta}
\def\P{\Pi}
\def\G{\Gamma}

\def\re{\Re}
\def\rn{\Re^n}

\def\thereis{\exists}
\def\gr{\nabla}
\def\lt{\lim_{t\tends\infty}}
\def\lty{\lim_{t\tends\infty}}
\def\noteq{\ne}
\def\mapr{:\re\mapsto\re}
\def\mapn{:\rn\mapsto\re}

\def\bn{\hfil\break}
\def\bnt{\hfil\break\indent} 
\def\tl{\tilde}
\def\ab{\allowbreak}
\def\old#1{}
\def\leaderfill{\leaders\hbox to 1em{\hss.\hss}\hfill}

\def\ie{i.e.}
\def\eg{e.g.}

\parindent=2pc
\baselineskip=15pt
\vsize=8.7 true in
\voffset=0.125 true in
\parskip=3pt
\def\singlespace{\baselineskip 12 pt}
\def\onehalfspace{\baselineskip 18 pt}
\def\doublespace{\baselineskip 24 pt}
\def\normalspace{\baselineskip 15 pt}


\def\colvect#1#2{\lf(\matrix{#1\cr#2\cr}\ri)}
\def\rowvect#1#2{\lf(\matrix{#1&#2\cr}\ri)}
\def\twomat#1#2#3#4{\lf(\matrix{#1&#2\cr #3&#4\cr}\ri)}
\def\ncolvect#1#2{\lf(\matrix{#1\cr\vdots\cr#2\cr}\ri)}
\def\nrowvect#1#2{\lf(\matrix{#1&\ldots&#2\cr}\ri)}

\def\minprob#1#2#3{$$\eqalign{&\hbox{minimize\ \ }#1\cr &\hbox{subject to\ \
}#2\cr}\ifnum 0=#3{}\else\eqno(#3)\fi$$}        
\def\minprobn#1#2#3{$$\eqalign{&\hbox{minimize\ \ }#1\cr &\hbox{subject to\ \
}#2\cr}\eqno(#3)$$}     
\def\maxprob#1#2#3{$$\eqalign{&\hbox{maximize\ \ }#1\cr &\hbox{subject to\ \
}#2\cr}\ifnum 0=#3{}\else\eqno(#3)\fi$$}        
\def\maxprobn#1#2#3{$$\eqalign{&\hbox{maximize\ \ }#1\cr &\hbox{subject to\ \
}#2\cr}\eqno(#3)$$}     
\def\aligntwo#1#2#3#4#5{$$\eqalign{#1&#2\cr #3&#4\cr}
\ifnum 0=#5{}\else\eqno(#5)\fi$$}
\def\alignthree#1#2#3#4#5#6#7{$$\eqalign{#1&#2\cr #3&#4\cr #5&#6\cr}
\ifnum 0=#7{}\else\eqno(#7)\fi$$}


\def\eqnum{\eqno{\hbox{(\the\sectnum.\the\eqnumber)}\global\advance\eqnumber
by1}}

\def\eqnu{\eqno{\hbox{(\the\sectnum.\the\eqnumber)}\global\advance\eqnumber
by1}}

\newcount\examplnumber
\def\examplnum{\global\advance\examplnumber by1}

\newcount\figrnumber
\def\figrnum{\global\advance\figrnumber by1}

\newcount\propnumber
\def\propnum{\global\advance\propnumber by1}

\newcount\defnumber
\def\defnum{\global\advance\defnumber by1}

\newcount\lemmanumber
\def\lemmanum{\global\advance\lemmanumber by1}

\newcount\assumptionnumber
\def\assumptionnum{\global\advance\assumptionnumber by1}

\newcount\conditionnumber
\def\conditionnum{\global\advance\conditionnumber by1}

\def\exampl{\the\sectnum.\the\examplnumber}
\def\figr{\the\sectnum.\the\figrnumber}
\def\propn{\the\sectnum.\the\propnumber}
\def\defn{\the\sectnum.\the\defnumber}
\def\lemman{\the\sectnum.\the\lemmanumber}
\def\assumptionn{\the\sectnum.\the\assumptionnumber}
\def\condn{\the\sectnum.\the\conditionnumber}

\def\section#1{\goodbreak\vskip 3pc plus 6pt minus 3pt\leftskip=-2pc
   \global\advance\sectnum by 1\eqnumber=1
\global\examplnumber=1\figrnumber=1\propnumber=1\defnumber=1\lemmanumber=1\assumptionnumber=1 \conditionnumber =1%
   \line{\hfuzz=1pc{\hbox to 3pc{\bf 
   \vtop{\hfuzz=1pc\hsize=38pc\hyphenpenalty=10000\noindent\uppercase{\the\sectnum.\quad #1}}\hss}}
			\hfill}
			\leftskip=0pc\nobreak\tenf
			\vskip 1pc plus 4pt minus 2pt\noindent\ignorespaces}



\def\sect#1{\noindent\leftskip=-2pc\tenf
   \goodbreak\vskip 1pc plus 4pt minus 2pt
                \global\advance\subsectnum by 1\eqnumber=1
   \line{\hfuzz=1pc{\hbox to 3pc{\bf 
   \vtop{\hfuzz=1pc\hsize=38pc\hyphenpenalty=10000\noindent\uppercase{{\bf #1}}}\hss}}
                        \hfill}
   \leftskip=0pc\nobreak\tenf
                        \vskip 1pc plus 4pt minus 2pt\nobreak\noindent\ignorespaces}

\def\subsection#1{\noindent\leftskip=0pc\tenf
   \goodbreak\vskip 1pc plus 4pt minus 2pt
   \line{\hfuzz=1pc{\hbox to 3pc{\bf 
   \vtop{\hfuzz=1pc\hsize=38pc\hyphenpenalty=10000\noindent{\bf #1}}\hss}}
                        \hfill}
   \leftskip=0pc\nobreak\tenf
                        \vskip 1pc plus 4pt minus 2pt\nobreak\noindent\ignorespaces}
\def\subsubsection#1{\goodbreak\vskip 1pc plus 4pt minus 2pt
   \hfuzz=3pc\leftskip=0pc\noindent\tenit #1 \nobreak\tenf\vskip 6pt plus 1pt
                                minus 1pt\nobreak\ignorespaces\leftskip=0pc}
%
\def\textlist#1{\par{\bf #1}\ }

\def\beginexample#1{\noindent\goodbreak\vskip 6pt plus 1pt minus 1pt
\noindent
  \hbox {\bf Example #1\hss}
  \nobreak\vskip 4pt plus 1pt minus 1pt \nobreak\noindent\ninef
  \global\advance
                \leftskip by\parindent\pn}
\def\endexample{\vskip 12pt\tenf\par
  \global\advance\leftskip by -\parindent
  }

\def\beginexercise#1{\noindent\goodbreak\vskip 6pt plus 1pt minus 1pt \noindent\global\normalbaselineskip=12pt
  \hbox {\bf Exercise #1\hss}
  \nobreak\vskip 4pt plus 1pt minus 1pt 
  \nobreak\noindent\ninef\global\advance\leftskip
                        by\parindent\pn}
\def\endexercise{\vskip 12pt\tenf\par
  \global\advance\leftskip by -\parindent
  }

\def\beginsection#1{\noindent\goodbreak\vskip 6pt plus 1pt minus 1pt \noindent\global\normalbaselineskip=12pt
  \hbox {\it #1\hss}
  \vskip 0.1pt plus 1pt minus 1pt \nobreak\noindent\ninef\global\advance
                \leftskip by\parindent\noindent\pn}
\def\endsection{\vskip 12pt\tenf\par
  \global\advance\leftskip by -\parindent
}

\def\enddisplaylist{\vskip 12pt\par}


\def\lemma#1{\smskip\pn{\bf Lemma #1}\quad}
\def\theorem#1{\smskip\pn{\bf Theorem #1}\quad}
\def\proposition#1{\smskip\pn{\bf Proposition #1}\quad}
\def\proof{\smskip\pn{\bf Proof:}\quad} 
\def\definition#1{\smskip\pn{\bf Definition #1}\quad} 
\def\assumption#1{\smskip\pn{\bf Assumption #1}\quad}
\def\corollary#1{\smskip\pn{\bf Corollary #1}\quad}
\def\exercise#1{\smskip\pn{\bf Exercise #1}\quad}

\def\figure#1{\smskip\pn{\bf Figure #1}\quad}
\def\QED{\quad{\bf Q.E.D.} \par\bigskip} \def\qed{\quad{\bf
Q.E.D.} \par\bigskip}
\def\ref{\smskip\pn}

\def\chapter#1#2{{\bf \centerline{\helbigbig
{#1}}}\bigskip\bigskip{\bf \centerline{\helbigbig
{#2}}}\bigskip\bigskip} 

\def\longchapter#1#2#3{{\bf \centerline{\helbigbig
{#1}}}\bigskip\bigskip{\bf \centerline{\helbigbig
{#2}}}\bigskip{\bf \centerline{\helbigbig
{#3}}}\bigskip\bigskip} 

\def\papertitle#1#2{{\bf \centerline{\helbigb
{#1}}}\bigskip\bigskip{\centerline{
by}}\bigskip{\bf \centerline{
{#2}}}\bigskip\bigskip} 

\def\longpapertitle#1#2#3{{\bf \centerline{\helbigb
{#1}}}\bigskip{\bf \centerline{\helbigb
{#2}}}\bigskip\bigskip{\centerline{
by}}\bigskip{\bf \centerline{
{#3}}}\bigskip\bigskip} 


\def\nitem#1{\smskip\item{#1}}
\def\nitemitem#1{\smskip\itemitem{#1}}
\def\endlist{\smskip}

\newcount\alphanum
\newcount\romnum

\def\alphaenumerate{\ifcase\alphanum \or (a)\or (b)\or (c)\or (d)\or (e)\or
(f)\or (g)\or (h)\or (i)\or (j)\or (k)\fi}
\def\romenumerate{\ifcase\romnum \or (i)\or (ii)\or (iii)\or (iv)\or (v)\or
(vi)\or (vii)\or (viii)\or (ix)\or (x)\or (xi)\fi}

\def\alist{\begingroup\vskip10pt\alphanum=1
\parskip=2pt\parindent=0pt \leftskip=3pc
\everypar{\llap{{\rm\alphaenumerate\hskip1em}}\advance\alphanum by1}}
\def\endalist{\vskip1pc\endgroup\parskip=0pt
  \parindent=\bigskipamount \leftskip=0pc\tenf
                        \noindent\ignorespaces}

\def\nolist{\begingroup\vskip10pt\alphanum=0
\parskip=2pt\parindent=0pt \leftskip=3pc
\everypar{\llap{\global\advance\alphanum by1(\the\alphanum)\hskip1em}}}
\def\endnolist{\vskip1pc\endgroup\parskip=0pt\leftskip=0pc\tenf
                        \noindent\ignorespaces}

\def\romlist{\begingroup\vskip10pt\romnum=1
\parskip=2pt\parindent=0pt \leftskip=5pc
\everypar{\llap{{\rm\romenumerate\hskip1em}}\advance\romnum by1}}
\def\endromlist{\vskip1pc\endgroup\parskip=0pt\leftskip=0pc\tenf
                        \noindent\ignorespaces}


\long\def\fig#1#2#3{\vbox{\vskip1pc\vskip#1
\prevdepth=12pt \baselineskip=12pt
\vskip1pc
\hbox to\hsize{\hfill\vtop{\hsize=25pc\noindent{\eightbf Figure #2\ }
{\eightpoint#3}}\hfill}}}

\long\def\widefig#1#2#3{\vbox{\vskip1pc\vskip#1
\prevdepth=12pt \baselineskip=12pt
\vskip1pc
\hbox to\hsize{\hfill\vtop{\hsize=28pc\noindent{\eightbf Figure #2\ }
{\eightpoint#3}}\hfill}}}

\long\def\table#1#2{\vbox{\vskip0.5pc
\prevdepth=12pt \baselineskip=12pt
\hbox to\hsize{\hfill\vtop{\hsize=25pc\noindent{\eightbf Table #1\ }
{\eightpoint#2}}\hfill}}}

\def\rightleftheadline#1#2{ifodd\pageno\rightheadline{#1}
\else\leftheadline{#2}\fi} 
\def\rightheadline#1{\headline{\tenrm\hfil #1}}
\def\leftheadline#1{\headline{\tenrm#1\hfil}}


\long\def\leftfig#1#2{\vbox{\smskip\hsize=220pt
\vtop{{\noindent {\bf #1}}}
\smskip
\noindent
\vbox{{\noindent #2}}
}}

\long\def\rightfig#1#2#3{\vbox{\smskip\vskip#1
\prevdepth=12pt \baselineskip=12pt
\hsize=210pt
\smskip
\vbox{\noindent{\eightbold #2}
\hskip1em{\eightpoint#3}}
}}

\long\def\concept#1#2#3#4#5{\bigskip\hrule
\vbox{\hbox{\leftfig{#1}{#2} \hskip3em
\rightfig{#3}{#4}{#5}} \smskip}
\hrule\bigskip}


\long\def\bconcept#1#2#3#4#5#6#7{
\vbox{
\hbox to \hsize{\vtop{\par #1}}
\concept{#2}{#3}{#4}{#5}{#6}
\hbox to \hsize{\vtop{\par #7}}
\smskip}
}


\long\def\boxconcept#1#2#3#4#5{
\vbox{\hbox{\leftfig{#1}{#2} \hskip3em
\rightfig{#3}{#4}{#5}} \smskip}
}


\def\boxit#1{\vbox{\hrule\hbox{\vrule\kern3pt
                                \vbox{\kern3pt#1\kern3pt}\kern3pt\vrule}\hrule}}
\def\centerboxit#1{$$\vbox{\hrule\hbox{\vrule\kern3pt
                                \vbox{\kern3pt#1\kern3pt}\kern3pt\vrule}\hrule}$$}

\long\def\boxtext#1#2{$$\boxit{\vbox{\hsize #1\noindent\strut #2\strut}}$$}

%
%
%

\def\picture #1 by #2 (#3){
  \vbox to #2{
    \hrule width #1 height 0pt depth 0pt
    \vfill
    \special{picture #3} 
    }
  }

\def\scaledpicture #1 by #2 (#3 scaled #4){{
  \dimen0=#1 \dimen1=#2
  \divide\dimen0 by 1000 \multiply\dimen0 by #4
  \divide\dimen1 by 1000 \multiply\dimen1 by #4
  \picture \dimen0 by \dimen1 (#3 scaled #4)}
  }

%
%

\long\def\captfig#1#2#3#4#5{\vbox{\vskip1pc
\hbox to\hsize{\hfill{\picture #1 by #2 (#3)}\hfill}
\prevdepth=9pt \baselineskip=9pt
\vskip1pc
\hbox to\hsize{\hfill\vtop{\hsize=24pc\noindent{\eightbold Figure #4}
\hskip1em{\eightpoint#5}}\hfill}}}

%
%
%

\def\illustration #1 by #2 (#3){
  \vskip#2\hskip#1\special{illustration #3} 
    }

\def\scaledillustration #1 by #2 (#3 scaled #4){{
  \dimen0=#1 \dimen1=#2
  \divide\dimen0 by 1000 \multiply\dimen0 by #4
  \divide\dimen1 by 1000 \multiply\dimen1 by #4
  \illustration \dimen0 by \dimen1 (#3 scaled #4)}
  }


\newbox\graybox
\newdimen\xgrayspace
\newdimen\ygrayspace
%
%
%
%
%
%
%
%
%

\def\Textshade#1#2#3#4#5#6{%
    \xgrayspace=#4pt%
    \ygrayspace=#4pt%
    \def\grayshade{#3}%
    \def\linewidth{#5}%
    \def\theradius{#6}%
    \setbox\graybox=\hbox{\surroundboxa{#2}}%
    \hbox{%
    \hbox to 0pt{%
    \PScommands
}%
    \box\graybox}}%
%
%
\long%

\long%
\def\Parashade#1#2#3#4#5#6#7{%
    \xgrayspace=#4pt%
    \ygrayspace=#4pt%
    \def\grayshade{#3}%
    \def\linewidth{#5}%
    \def\theradius{#6}%
    \def\thevskip{#7pt}%
    \setbox\graybox=\hbox{\surroundboxb{#2}}%
    \vskip\thevskip%
    \hbox{%
    \hbox to 0pt{%
    \PScommands
}%
     \box\graybox}%
     \vskip\thevskip%
}%
%
%
%
\long\def\surroundboxa#1{\leavevmode\hbox{\vtop{%
\vbox{\kern\ygrayspace%
\hbox{\kern\xgrayspace#1%
      \kern\xgrayspace}}\kern\ygrayspace}}}
%
%
\long\def\surroundboxb#1{\leavevmode\hbox{\vtop{%
\vbox{\kern\ygrayspace%
\hbox{\kern\xgrayspace\vbox{\advance\hsize-2\xgrayspace#1}%
      \kern\xgrayspace}}\kern\ygrayspace}}}
%
%
%
\long\def\PScommands{%
\special{rawpostscript
/sharpbox{%
           newpath
           xmin ymin moveto
           xmin ymax lineto
           xmax ymax lineto
           xmax ymin lineto
           xmin ymin lineto
           closepath 
          }bind def
}%
\special{rawpostscript
/sharpboxnb{%
           newpath
           xmin ymin moveto
           xmin ymax lineto
           xmax ymax lineto
           xmax ymin lineto
          }bind def
}%
\special{rawpostscript
/sharpboxnt{%
           newpath
           xmin ymax moveto
           xmin ymin lineto
           xmax ymin lineto
           xmax ymax lineto
          }bind def
}%
\special{rawpostscript
/roundbox{%
           newpath
           xmin radius add ymin moveto
           xmax ymin xmax ymax radius arcto
           xmax ymax xmin ymax radius arcto
           xmin ymax xmin ymin radius arcto
           xmin ymin xmax ymin radius arcto 16 {pop} repeat
           closepath
          }bind def
}%
\special{rawpostscript
/sharpcorners{%
               sharpbox gsave grayshade setgray fill grestore 
               linewidth setlinewidth stroke
              }bind def
}%
\special{rawpostscript
/sharpcornersnt{%
               sharpboxnt gsave grayshade setgray fill grestore 
               linewidth setlinewidth stroke
              }bind def
}%
\special{rawpostscript
/sharpcornersnb{%
               sharpboxnb gsave grayshade setgray fill grestore 
               linewidth setlinewidth stroke
              }bind def
}%
\special{rawpostscript
/roundcorners{%
               roundbox gsave grayshade setgray fill grestore 
               linewidth setlinewidth stroke
              }bind def
}%
\special{rawpostscript
/plainbox{%
           sharpbox grayshade setgray fill 
          }bind def
}%
%
\special{rawpostscript
/roundnoframe{%
               roundbox grayshade setgray fill 
              }bind def
}%
\special{rawpostscript
/sharpnoframe{%
               sharpbox grayshade setgray fill 
              }bind def
}%
}%
%
%

\def\pshade#1{\Parashade{sharpcorners}{#1}{0.95}{10}{0.5}{10}{10}}


\def\boxit#1{\vbox{\hrule\hbox{\vrule\kern3pt
                                \vbox{\kern3pt#1\kern3pt}\kern3pt\vrule}\hrule}}

\def\boxitnb#1{\vbox{\hrule\hbox{\vrule\kern3pt
                                \vbox{\kern3pt#1\kern3pt}\kern3pt\vrule}}}

\def\boxitnt#1{\vbox{\hbox{\vrule\kern3pt
                                \vbox{\kern3pt#1\kern3pt}\kern3pt\vrule}\hrule}}

\long\def\boxtext#1#2{$$\boxit{\vbox{\hsize #1\noindent\strut #2\strut}}$$}
\long\def\boxtextnb#1#2{$$\boxitnb{\vbox{\hsize #1\noindent\strut #2\strut}}$$}
\long\def\boxtextnt#1#2{$$\boxitnt{\vbox{\hsize #1\noindent\strut #2\strut}}$$}

\def\texshopbox#1{\boxtext{462pt}{\vskip-1.5pc\pshade{\vskip-1.0pc#1\vskip-2.0pc}}}
\def\texshopboxnt#1{\boxtextnt{462pt}{\vskip-1.5pc\pshade{\vskip-1.0pc#1\vskip-2.0pc}}}
\def\texshopboxnb#1{\boxtextnb{462pt}{\vskip-1.5pc\pshade{\vskip-1.0pc#1\vskip-2.0pc}}}


\font\hel=cmr10%
\font\helb=cmbx10%
\font\heli=cmti10%
\font\helbi=cmsl10%
\font\ninehel=cmr9%
\font\nineheli=cmti9%
\font\ninehelb=cmbx9%
\font\helbig=cmr10 scaled 1500%
\font\helbigbig=cmr10 scaled 2500%
\font\helbigb=cmbx10 scaled 1500%
\font\helbigbigb=cmbx10 scaled 2500%
\font\bigi=cmti10 scaled \magstep5%
\font\eightbold=cmbx8%
\font\ninebold=cmbx9%
\font\boldten=cmbx10%

\def\tenf{\hel}%
\def\tenit{\heli}%
\def\ninef{\ninehel}%
\def\nineb{\ninehelb}%
\def\nineit{\nineheli}%
\def\smit{\nineheli}%
\def\smbf{\ninehelb}%


\font\tenrm=cmr10%
\font\teni=cmmi10%
\font\tensy=cmsy10%
\font\tenbf=cmbx10%
\font\tentt=cmtt10%
\font\tenit=cmti10%
\font\tensl=cmsl10%

\def\tenpoint{\def\rm{\fam0\tenrm}%
\textfont0=\tenrm%
\textfont1=\teni%
\textfont2=\tensy%
\textfont\itfam=\tenit%
\textfont\slfam=\tensl%
\textfont\ttfam=\tentt%
\textfont\bffam=\tenbf%
\scriptfont0=\sevenrm%
\scriptfont1=\seveni%
\scriptfont2=\sevensy%
\scriptscriptfont0=\sixrm%
\scriptscriptfont1=\sixi%
\scriptscriptfont2=\sixsy%
\def\it{\fam\itfam\tenit}%
\def\tt{\fam\ttfam\tentt}%
\def\sl{\fam\slfam\tensl}%
\scriptfont\bffam=\sevenbf%
\scriptscriptfont\bffam=\sixbf%
\def\bf{\fam\bffam\tenbf}%
\normalbaselineskip=18pt%
\normalbaselines\rm}%

\font\ninerm=cmr9%
\font\ninebf=cmbx9%
\font\nineit=cmti9%
\font\ninesy=cmsy9%
\font\ninei=cmmi9%
\font\ninett=cmtt9%
\font\ninesl=cmsl9%

\def\ninepoint{\def\rm{\fam0\ninerm}%
\textfont0=\ninerm%
\textfont1=\ninei%
\textfont2=\ninesy%
\textfont\itfam=\nineit%
\textfont\slfam=\ninesl%
\textfont\ttfam=\ninett%
\textfont\bffam=\ninebf%
\scriptfont0=\sixrm%
\scriptfont1=\sixi%
\scriptfont2=\sixsy%
\def\it{\fam\itfam\nineit}%
\def\tt{\fam\ttfam\ninett}%
\def\sl{\fam\slfam\ninesl}%
\scriptfont\bffam=\sixbf%
\scriptscriptfont\bffam=\fivebf%
\def\bf{\fam\bffam\ninebf}%
\normalbaselineskip=16pt%
\normalbaselines\rm}%

\font\eightrm=cmr8%
\font\eighti=cmmi8%
\font\eightsy=cmsy8%
\font\eightbf=cmbx8%
\font\eighttt=cmtt8%
\font\eightit=cmti8%
\font\eightsl=cmsl8%

\def\eightpoint{\def\rm{\fam0\eightrm}%
\textfont0=\eightrm%
\textfont1=\eighti%
\textfont2=\eightsy%
\textfont\itfam=\eightit%
\textfont\slfam=\eightsl%
\textfont\ttfam=\eighttt%
\textfont\bffam=\eightbf%
\scriptfont0=\sixrm%
\scriptfont1=\sixi%
\scriptfont2=\sixsy%
\scriptscriptfont0=\fiverm%
\scriptscriptfont1=\fivei%
\scriptscriptfont2=\fivesy%
\def\it{\fam\itfam\eightit}%
\def\tt{\fam\ttfam\eighttt}%
\def\sl{\fam\slfam\eightsl}%
\scriptscriptfont\bffam=\fivebf%
\def\bf{\fam\bffam\eightbf}%
\normalbaselineskip=14pt%
\normalbaselines\rm}%

\font\sevenrm=cmr7%
\font\seveni=cmmi7%
\font\sevensy=cmsy7%
\font\sevenbf=cmbx7%
\font\seventt=cmtt8 at 7pt%
\font\sevenit=cmti8 at 7pt%
\font\sevensl=cmsl8 at 7pt%

\def\sevenpoint{%
   \def\rm{\sevenrm}\def\bf{\sevenbf}%
   \def\smc{\sevensmc}\baselineskip=12pt\rm}%

\font\sixrm=cmr6%
\font\sixi=cmmi6%
\font\sixsy=cmsy6%
\font\sixbf=cmbx6%
\font\sixtt=cmtt8 at 6pt%
\font\sixit=cmti8 at 6pt%
\font\sixsl=cmsl8 at 6pt%
\font\sixsmc=cmr8 at 6pt%

\def\sixpoint{%
   \def\rm{\sixrm}\def\bf{\sixbf}%
   \def\smc{\sixsmc}\baselineskip=12pt\rm}%

\fontdimen13\tensy=2.6pt%
\fontdimen14\tensy=2.6pt%
\fontdimen15\tensy=2.6pt%
\fontdimen16\tensy=1.2pt%
\fontdimen17\tensy=1.2pt%
\fontdimen18\tensy=1.2pt%

\def\tenf{\tenpoint}%
\def\ninef{\ninepoint}%
\def\eightf{\eightpoint}%



 \ifx\MYUNDEFINED\BoxedEPSF
   \let\temp\relax
 \else
   \message{}
   \message{ !!! BoxedEPS %
         or BoxedArt macros already defined !!!}
   \let\temp\endinput
 \fi
  \temp
 
 \chardef\EPSFCatAt\the\catcode`\@
 \catcode`\@=11

 \chardef\C@tColon\the\catcode`\:
 \chardef\C@tSemicolon\the\catcode`\;
 \chardef\C@tQmark\the\catcode`\?
 \chardef\C@tEmark\the\catcode`\!
 \chardef\C@tDqt\the\catcode`\"

 \def\PunctOther@{\catcode`\:=12
   \catcode`\;=12 \catcode`\?=12 \catcode`\!=12 \catcode`\"=12}
 \PunctOther@

 \let\wlog@ld\wlog 
 \def\wlog#1{\relax} 

 \newif\ifIN@
 \newdimen\XShift@ \newdimen\YShift@ 
 \newtoks\Realtoks
 
  %
 \newdimen\Wd@ \newdimen\Ht@
 \newdimen\Wd@@ \newdimen\Ht@@
 \newdimen\TT@
 \newdimen\LT@
 \newdimen\BT@
 \newdimen\RT@
 \newdimen\XSlide@ \newdimen\YSlide@ 
 \newdimen\TheScale  
 \newdimen\FigScale  
 \newdimen\ForcedDim@@

 \newtoks\EPSFDirectorytoks@
 \newtoks\EPSFNametoks@
 \newtoks\BdBoxtoks@
 \newtoks\LLXtoks@  
 \newtoks\LLYtoks@

 \newif\ifNotIn@
 \newif\ifForcedDim@
 \newif\ifForceOn@
 \newif\ifForcedHeight@
 \newif\ifPSOrigin

 \newread\EPSFile@ 
 
  \def\ms@g{\immediate\write16}

 \newif\ifIN@\def\IN@{\expandafter\INN@\expandafter}
  \long\def\INN@0#1@#2@{\long\def\NI@##1#1##2##3\ENDNI@
    {\ifx\m@rker##2\IN@false\else\IN@true\fi}%
     \expandafter\NI@#2@@#1\m@rker\ENDNI@}
  \def\m@rker{\m@@rker}

  \newtoks\Initialtoks@  \newtoks\Terminaltoks@
  \def\SPLIT@{\expandafter\SPLITT@\expandafter}
  \def\SPLITT@0#1@#2@{\def\TTILPS@##1#1##2@{%
     \Initialtoks@{##1}\Terminaltoks@{##2}}\expandafter\TTILPS@#2@}


  \newtoks\Trimtoks@

 \def\ForeTrim@{\expandafter\ForeTrim@@\expandafter}
 \def\ForePrim@0 #1@{\Trimtoks@{#1}}
 \def\ForeTrim@@0#1@{\IN@0\m@rker. @\m@rker.#1@%
     \ifIN@\ForePrim@0#1@%
     \else\Trimtoks@\expandafter{#1}\fi}

  \def\Trim@0#1@{%
      \ForeTrim@0#1@%
      \IN@0 @\the\Trimtoks@ @%
        \ifIN@ 
             \SPLIT@0 @\the\Trimtoks@ @\Trimtoks@\Initialtoks@
             \IN@0\the\Terminaltoks@ @ @%
                 \ifIN@
                 \else \Trimtoks@ {FigNameWithSpace}%
                 \fi
        \fi
      }


   \newtoks\pt@ks
   \def \getpt@ks 0.0#1@{\pt@ks{#1}}
   \dimen0=0pt\relax\expandafter\getpt@ks\the\dimen0@

  \newtoks\Realtoks
  \def\Real#1{%
    \dimen2=#1%
      \SPLIT@0\the\pt@ks @\the\dimen2@
       \Realtoks=\Initialtoks@
            }

   \newdimen\Product
   \def\Mult#1#2{%
     \dimen4=#1\relax
     \dimen6=#2%
     \Real{\dimen4}%
     \Product=\the\Realtoks\dimen6%
        }

 \newdimen\Inverse
 \newdimen\hmxdim@ \hmxdim@=8192pt
 \def\Invert#1{%
  \Inverse=\hmxdim@
  \dimen0=#1%
  \divide\Inverse \dimen0%
  \multiply\Inverse 8}

   \def\Rescale#1#2#3{
              \divide #1 by 100\relax
              \dimen2=#3\divide\dimen2 by 100 \Invert{\dimen2}%
              \Mult{#1}{#2}%
              \Mult\Product\Inverse 
              #1=\Product}

  \def\Scale#1{\dimen0=\TheScale %
      \divide #1 by  1280 
      \divide \dimen0 by 5120 %
      \multiply#1 by \dimen0 
      \divide#1 by 10   
     }
 

 \newbox\scrunchbox

 \def\Scrunched#1{{\setbox\scrunchbox\hbox{#1}%
   \wd\scrunchbox=0pt
   \ht\scrunchbox=0pt
   \dp\scrunchbox=0pt
   \box\scrunchbox}}

 \def\Shifted@#1{%
   \vbox {\kern-\YShift@
       \hbox {\kern\XShift@\hbox{#1}\kern-\XShift@}%
           \kern\YShift@}}


 \def\cBoxedEPSF#1{{\leavevmode 
   \ReadNameAndScale@{#1}%
   \SetEPSFSpec@
   \ReadEPSFile@ \ReadBdB@x  
     \TrimFigDims@ 
     \CalculateFigScale@  
     \ScaleFigDims@
     \SetInkShift@
   \hbox{$\mathsurround=0pt\relax
         \vcenter{\hbox{%
             \FrameSpider{\hskip-.4pt\vrule}%
             \vbox to \Ht@{\offinterlineskip\parindent=\z@%
                \FrameSpider{\vskip-.4pt\hrule}\vfil 
                \hbox to \Wd@{\hfil}%
                \vfil
                \InkShift@{\EPSFSpecial{\EPSFSpec@}{\FigSc@leReal}}%
             \FrameSpider{\hrule\vskip-.4pt}}%
         \FrameSpider{\vrule\hskip-.4pt}}}%
     $}%
    \CleanRegisters@ 
    \ms@g{ *** Box composed for the %
         EPSF file \the\EPSFNametoks@}%
    }}
 
 \def\tBoxedEPSF#1{\setbox4\hbox{\cBoxedEPSF{#1}}%
     \setbox4\hbox{\raise -\ht4 \hbox{\box4}}%
     \box4
      }

 \def\bBoxedEPSF#1{\setbox4\hbox{\cBoxedEPSF{#1}}%
     \setbox4\hbox{\raise \dp4 \hbox{\box4}}%
     \box4
      }

  \let\BoxedEPSF\cBoxedEPSF

   %

   %
  \def\gLinefigure[#1scaled#2]_#3{%
        \BoxedEPSF{#3 scaled #2}}
    
   %

  \def\EPSFxsize{\afterassignment\ForceW@\ForcedDim@@}
      \def\ForceW@{\ForcedDim@true\ForcedHeight@false}
  
  \def\EPSFysize{\afterassignment\ForceH@\ForcedDim@@}
      \def\ForceH@{\ForcedDim@true\ForcedHeight@true}

  \def\EmulateRokicki{%
       \let\epsfbox\bBoxedEPSF \let\epsffile\bBoxedEPSF
       \let\epsfxsize\EPSFxsize \let\epsfysize\EPSFysize} 
 
  %
 \def\ReadNameAndScale@#1{\IN@0 scaled@#1@
   \ifIN@\ReadNameAndScale@@0#1@%
   \else \ReadNameAndScale@@0#1 scaled\DefaultMilScale @%
   \fi}
  
 \def\ReadNameAndScale@@0#1scaled#2@{
    \let\OldBackslash@\\%
    \def\\{\OtherB@ckslash}%
    \edef\temp@{#1}%
    \Trim@0\temp@ @%
    \EPSFNametoks@\expandafter{\the\Trimtoks@ }%
    \FigScale=#2 pt%
    \let\\\OldBackslash@
    }
 
 \def\SetDefaultEPSFScale#1{%
      \global\def\DefaultMilScale{#1}}

 \SetDefaultEPSFScale{1000}

  %
 \def \SetBogusBbox@{%
     \global\BdBoxtoks@{ BoundingBox:0 0 100 100 }%
     \global\def\BdBoxLine@{ BoundingBox:0 0 100 100 }%
     \ms@g{ !!! Will use placeholder !!!}%
     }

 {\catcode`\%=12\gdef\P@S@{

 \def\ReadEPSFile@{
     \openin\EPSFile@\EPSFSpec@
     \relax  
  \ifeof\EPSFile@
     \ms@g{}%
     \ms@g{ !!! EPS FILE \the\EPSFDirectorytoks@
       \the\EPSFNametoks@\space WAS NOT FOUND !!!}%
     \SetBogusBbox@
  \else
   \begingroup
   \catcode`\%=12\catcode`\:=12\catcode`\!=12
   \catcode`\G=14\catcode`\\=14\relax
   \global\read\EPSFile@ to \BdBoxLine@
   \IN@0\P@S@ @\BdBoxLine@ @%
   \ifIN@ 
     \NotIn@true
     \loop   
       \ifeof\EPSFile@\NotIn@false 
         \ms@g{}%
         \ms@g{ !!! BoundingBox NOT FOUND IN %
            \the\EPSFDirectorytoks@\the\EPSFNametoks@\space!!! }%
         \SetBogusBbox@
       \else\global\read\EPSFile@ to \BdBoxLine@
       \fi
       \global\BdBoxtoks@\expandafter{\BdBoxLine@}%
       \IN@0BoundingBox:@\the\BdBoxtoks@ @%
       \ifIN@\NotIn@false\fi%
     \ifNotIn@\repeat
   \else
         \ms@g{}%
         \ms@g{ !!! \the\EPSFNametoks@\space not PS!\space !!!}%
         \SetBogusBbox@
   \fi
  \endgroup\relax
  \fi
  \closein\EPSFile@ 
   }

  \def\ReadBdB@x{
   \expandafter\ReadBdB@x@\the\BdBoxtoks@ @}
  
  \def\ReadBdB@x@#1BoundingBox:#2@{
    \ForeTrim@0#2@%
    \IN@0atend@\the\Trimtoks@ @%
       \ifIN@\Trimtoks@={0 0 100 100 }%
         \ms@g{}%
         \ms@g{ !!! BoundingBox not found in %
         \the\EPSFDirectorytoks@\the\EPSFNametoks@\space !!!}%
         \ms@g{ !!! It must not be at end of EPSF !!!}%
         \ms@g{ !!! Will use placeholder !!!}%
       \fi
    \expandafter\ReadBdB@x@@\the\Trimtoks@ @%
   }
    
  \def\ReadBdB@x@@#1 #2 #3 #4@{
      \Wd@=#3bp\advance\Wd@ by -#1bp%
      \Ht@=#4bp\advance\Ht@ by-#2bp%
       \Wd@@=\Wd@ \Ht@@=\Ht@ 
       \LLXtoks@={#1}\LLYtoks@={#2}
      \ifPSOrigin\XShift@=-#1bp\YShift@=-#2bp\fi 
     }

   %
   \def\G@bbl@#1{}
   \bgroup
     \global\edef\OtherB@ckslash{\expandafter\G@bbl@\string\\}
   \egroup

  \def\SetEPSFDirectory{
           \bgroup\PunctOther@\relax
           \let\\\OtherB@ckslash
           \SetEPSFDirectory@}

 \def\SetEPSFDirectory@#1{
    \edef\temp@{#1}%
    \Trim@0\temp@ @
    \global\toks1\expandafter{\the\Trimtoks@ }\relax
    \egroup
    \EPSFDirectorytoks@=\toks1
    }

 \def\SetEPSFSpec@{%
     \bgroup
     \let\\=\OtherB@ckslash
     \global\edef\EPSFSpec@{%
        \the\EPSFDirectorytoks@\the\EPSFNametoks@}%
     \global\edef\EPSFSpec@{\EPSFSpec@}%
     \egroup}

  %
 \def\TrimTop#1{\advance\TT@ by #1}
 \def\TrimLeft#1{\advance\LT@ by #1}
 \def\TrimBottom#1{\advance\BT@ by #1}
 \def\TrimRight#1{\advance\RT@ by #1}

 \def\TrimBoundingBox#1{%
   \TrimTop{#1}%
   \TrimLeft{#1}%
   \TrimBottom{#1}%
   \TrimRight{#1}%
       }

 \def\TrimFigDims@{%
    \advance\Wd@ by -\LT@ 
    \advance\Wd@ by -\RT@ \RT@=\z@
    \advance\Ht@ by -\TT@ \TT@=\z@
    \advance\Ht@ by -\BT@ 
    }

  %
  \def\ForceWidth#1{\ForcedDim@true
       \ForcedDim@@#1\ForcedHeight@false}
  
  \def\ForceHeight#1{\ForcedDim@true
       \ForcedDim@@=#1\ForcedHeight@true}

  \def\ForceOn{\ForceOn@true}
  \def\ForceOff{\ForceOn@false\ForcedDim@false}
  
  \def\CalculateFigScale@{%
     \ifForcedDim@\FigScale=1000pt
           \ifForcedHeight@
                \Rescale\FigScale\ForcedDim@@\Ht@
           \else
                \Rescale\FigScale\ForcedDim@@\Wd@
           \fi
     \fi
     \Real{\FigScale}%
     \edef\FigSc@leReal{\the\Realtoks}%
     }
   
  \def\ScaleFigDims@{\TheScale=\FigScale
      \ifForcedDim@
           \ifForcedHeight@ \Ht@=\ForcedDim@@  \Scale\Wd@
           \else \Wd@=\ForcedDim@@ \Scale\Ht@
           \fi
      \else \Scale\Wd@\Scale\Ht@        
      \fi
      \ifForceOn@\relax\else\global\ForcedDim@false\fi
      \Scale\LT@\Scale\BT@  
      \Scale\XShift@\Scale\YShift@
      }
      
 \def\HideReservedBoxes{\global\def\FrameSpider##1{\null}}
 \def\ShowReservedBoxes{\global\def\FrameSpider##1{##1}}
 \let\HideDisplacementBoxes\HideReservedBoxes  
 \let\ShowDisplacementBoxes\ShowReservedBoxes
 \let\HideFigureFrames\HideReservedBoxes
 \let\ShowFigureFrames\ShowReservedBoxes
  \ShowDisplacementBoxes
 
 \def\hSlide#1{\advance\XSlide@ by #1}
 \def\vSlide#1{\advance\YSlide@ by #1}
 
  \def\SetInkShift@{%
            \advance\XShift@ by -\LT@
            \advance\XShift@ by \XSlide@
            \advance\YShift@ by -\BT@
            \advance\YShift@ by -\YSlide@
             }
  \def\InkShift@#1{\Shifted@{\Scrunched{#1}}}
 
   %
  \def\CleanRegisters@{%
      \globaldefs=1\relax
        \XShift@=\z@\YShift@=\z@\XSlide@=\z@\YSlide@=\z@
        \TT@=\z@\LT@=\z@\BT@=\z@\RT@=\z@
      \globaldefs=0\relax}

 
 \def\SetTexturesEPSFSpecial{\PSOriginfalse
  \gdef\EPSFSpecial##1##2{\relax
    \edef\specialthis{##2}%
    \SPLIT@0.@\specialthis.@\relax
    \special{illustration ##1 scaled
                        \the\Initialtoks@}}}
 
  \def\SetUnixCoopEPSFSpecial{\PSOrigintrue 
   \gdef\EPSFSpecial##1##2{%
      \dimen4=##2pt
      \divide\dimen4 by 1000\relax
      \Real{\dimen4}
      \edef\Aux@{\the\Realtoks}%
      \includegraphics{##1\space}}}

  \def\SetBechtolsheimEPSFSpecial@{
   \PSOrigintrue
   \special{\DriverTag@ Include0 "psfig.pro"}%
   \gdef\EPSFSpecial##1##2{%
      \dimen4=##2pt 
      \divide\dimen4 by 1000\relax
      \Real{\dimen4} 
      \edef\Aux@{\the\Realtoks}
      \special{\DriverTag@ Literal "10 10 0 0 10 10 startTexFig
           \the\mag\space 1000 div 3.25 neg mul 
           \the\mag\space 1000 div .25 neg mul translate 
           \the\mag\space 1000 div \Aux@\space mul 
           \the\mag\space 1000 div \Aux@\space mul scale "}%
      \special{\DriverTag@ Include1 "##1"}%
      \special{\DriverTag@ Literal "endTexFig "}%
        }}

  \def\SetBechtolsheimEPSFSpecial@{
   \PSOrigintrue
   \special{\DriverTag@ Include0 "psfig.pro"}%
   \gdef\EPSFSpecial##1##2{%
      \dimen4=##2pt 
      \divide\dimen4 by 1000\relax
      \Real{\dimen4} 
      \edef\Aux@{\the\Realtoks}
      \special{\DriverTag@ Literal "10 10 0 0 10 10 startTexFig
           \the\mag\space 1000 div 
           dup 3.25 neg mul 2 index .25 neg mul translate 
           \Aux@\space mul dup scale "}%
      \special{\DriverTag@ Include1 "##1"}%
      \special{\DriverTag@ Literal "endTexFig "}%
        }}

  \def\SetBechtolsheimDVITPSEPSFSpecial{\def\DriverTag@{dvitps: }%
      \SetBechtolsheimEPSFSpecial@}

  \def\SetBechtolsheimDVI2PSEPSFSSpecial{\def\DriverTag@{DVI2PS: }%
      \SetBechtolsheimEPSFSpecial@}

  \def\SetLisEPSFSpecial{\PSOrigintrue 
   \gdef\EPSFSpecial##1##2{%
      \dimen4=##2pt
      \divide\dimen4 by 1000\relax
      \Real{\dimen4}
      \edef\Aux@{\the\Realtoks}%
      \special{pstext="10 10 0 0 10 10 startTexFig\space
           \the\mag\space 1000 div \Aux@\space mul 
           \the\mag\space 1000 div \Aux@\space mul scale"}%
      \includegraphics{##1}%
      \special{pstext=endTexFig}%
        }}

  \def\SetRokickiEPSFSpecial{\PSOrigintrue 
   \gdef\EPSFSpecial##1##2{%
      \dimen4=##2pt
      \divide\dimen4 by 10\relax
      \Real{\dimen4}
      \edef\Aux@{\the\Realtoks}%
      \includegraphics{##1}}}

  \def\SetInlineRokickiEPSFSpecial{\PSOrigintrue 
   \gdef\EPSFSpecial##1##2{%
      \dimen4=##2pt
      \divide\dimen4 by 1000\relax
      \Real{\dimen4}
      \edef\Aux@{\the\Realtoks}%
      \special{ps::[begin] 10 10 0 0 10 10 startTexFig\space
           \the\mag\space 1000 div \Aux@\space mul 
           \the\mag\space 1000 div \Aux@\space mul scale}%
      \special{ps: plotfile ##1}%
      \special{ps::[end] endTexFig}%
        }}

 \def\SetOzTeXEPSFSpecial{\PSOrigintrue
 \gdef\EPSFSpecial##1##2{%
 \dimen4=##2pt
 \divide\dimen4 by 1000\relax
 \Real{\dimen4}
 \edef\Aux@{\the\Realtoks}
 \special{epsf=\string"##1\string"\space scale=\Aux@}%
 }} 

  \def\SetPSprintEPSFSpecial{\PSOriginFALSE 
   \gdef\EPSFSpecial##1##2{
     \special{##1\space 
       ##2 1000 div \the\mag\space 1000 div mul
       ##2 1000 div \the\mag\space 1000 div mul scale
       \the\LLXtoks@\space neg \the\LLYtoks@\space neg translate
       }}}

 \def\SetArborEPSFSpecial{\PSOriginfalse 
   \gdef\EPSFSpecial##1##2{%
     \edef\specialthis{##2}%
     \SPLIT@0.@\specialthis.@\relax 
     \special{ps: epsfile ##1\space \the\Initialtoks@}}}

 \def\SetClarkEPSFSpecial{\PSOriginfalse 
   \gdef\EPSFSpecial##1##2{%
     \Rescale {\Wd@@}{##2pt}{1000pt}%
     \Rescale {\Ht@@}{##2pt}{1000pt}%
     \special{dvitops: import 
           ##1\space\the\Wd@@\space\the\Ht@@}}}

  \let\SetDVIPSONEEPSFSpecial\SetUnixCoopEPSFSpecial
  \let\SetDVIPSoneEPSFSpecial\SetUnixCoopEPSFSpecial

  \def\SetBeebeEPSFSpecial{
   \PSOriginfalse%
   \gdef\EPSFSpecial##1##2{\relax
    \special{language "PS",
      literal "##2 1000 div ##2 1000 div scale",
      position = "bottom left",
      include "##1"}}}
  \let\SetDVIALWEPSFSpecial\SetBeebeEPSFSpecial

  \def\SetNorthlakeEPSFSpecial{\PSOrigintrue
   \gdef\EPSFSpecial##1##2{%
     \edef\specialthis{##2}%
     \SPLIT@0.@\specialthis.@\relax 
     \special{insert ##1,magnification=\the\Initialtoks@}}}

 \def\SetStandardEPSFSpecial{%
   \gdef\EPSFSpecial##1##2{%
     \ms@g{}
     \ms@g{%
       !!! Sorry! There is still no standard for \string%
       \special\space EPSF integration !!!}%
     \ms@g{%
      --- So you will have to identify your driver using a command}%
     \ms@g{%
      --- of the form \string\Set...EPSFSpecial, in order to get}%
     \ms@g{%
      --- your graphics to print.  See BoxedEPS.doc.}%
     \ms@g{}
     \gdef\EPSFSpecial####1####2{}
     }}

  \SetStandardEPSFSpecial 
 
 \let\wlog\wlog@ld 

 \catcode`\:=\C@tColon
 \catcode`\;=\C@tSemicolon
 \catcode`\?=\C@tQmark
 \catcode`\!=\C@tEmark
 \catcode`\"=\C@tDqt

 \catcode`\@=\EPSFCatAt

\def\G{\Gamma}
\def\sg{\partial}
\def\ul{\underline}
\def\ignore#1{}
\def\remark#1{\smskip\pn{\bf Remark #1}\quad} 

\newcount\sectnum
\newcount\subsectnum
\newcount\eqnumber

\global\eqnumber=1\sectnum=0


\def\lab{(\the\sectnum.\the\eqnumber)}



\def\show#1{#1}



\def\magnify{\magnification=1200}

\def\smskip{\vskip 5 pt}
\def\medskip{\vskip 10 pt}
\def\bigskip{\vskip 15 pt}
\def\pn{\par\noindent}
\def\br{\break}
\def\un{\underline}
\def\ov{\overline}
\def\Bl{\Bigl}
\def\Br{\Bigr}
\def\bl{\bigl} 
\def\br{\bigr} 
\def\lf{\left}
\def\ri{\right}

\def\argmin{\mathop{\arg \min}}
\def\tendsd{\downarrow}
\def\mod{\rm mod}
\def\tends{\rightarrow}
\def\thereis{\exists}
\def\implies{\Rightarrow}
\def\implied{\Leftarrow}
\def\ubar{\underline}
\def\ol#1{\overline{#1}}
\def\kth{$k^{\rm th}$ }
\def\ith{$i^{\rm th}$ }
\def\jth{$j^{\rm th}$ }
\def\lth{$\ell^{\rm th}$ }
\def\grad{\nabla}
\def\tendsd{\downarrow}
\def\tr{ ^{\prime}}
\def\half{{\scriptstyle {1\over 2}}}
\def\Ascr{{\cal A}}
\def\Fscr{{\cal F}}
\def\Pscr{{\cal P}}
\def\Zscr{{\cal Z}}
\def\Nscr{{\cal N}}
\def\Cscr{{\cal C}}

\def\a{\alpha}
\def\be{\beta}
\def\b{\beta}
\def\l{\lambda}
\def\g{\gamma}
\def\m{\mu}
\def\p{\pi}
\def\r{\rho}
\def\e{\epsilon}
\def\t{\tau}
\def\d{\delta}
\def\s{\sigma}
\def\f{\phi}
\def\o{\omega}
\def\D{\Delta}
\def\P{\Pi}
\def\G{\Gamma}

\def\re{\Re}
\def\rn{\Re^n}

\def\thereis{\exists}
\def\gr{\nabla}
\def\lt{\lim_{t\tends\infty}}
\def\lty{\lim_{t\tends\infty}}
\def\noteq{\ne}
\def\mapr{:\re\mapsto\re}
\def\mapn{:\rn\mapsto\re}

\def\bn{\hfil\break}
\def\bnt{\hfil\break\indent} 
\def\tl{\tilde}
\def\ab{\allowbreak}
\def\old#1{}
\def\leaderfill{\leaders\hbox to 1em{\hss.\hss}\hfill}

\def\ie{i.e.}
\def\eg{e.g.}

\parindent=2pc
\baselineskip=15pt
\vsize=8.7 true in
\voffset=0.125 true in
\parskip=3pt
\def\singlespace{\baselineskip 12 pt}
\def\onehalfspace{\baselineskip 18 pt}
\def\doublespace{\baselineskip 24 pt}
\def\normalspace{\baselineskip 15 pt}


\def\colvect#1#2{\lf(\matrix{#1\cr#2\cr}\ri)}
\def\rowvect#1#2{\lf(\matrix{#1&#2\cr}\ri)}
\def\twomat#1#2#3#4{\lf(\matrix{#1&#2\cr #3&#4\cr}\ri)}
\def\ncolvect#1#2{\lf(\matrix{#1\cr\vdots\cr#2\cr}\ri)}
\def\nrowvect#1#2{\lf(\matrix{#1&\ldots&#2\cr}\ri)}

\def\minprob#1#2#3{$$\eqalign{&\hbox{minimize\ \ }#1\cr &\hbox{subject to\ \
}#2\cr}\ifnum 0=#3{}\else\eqno(#3)\fi$$}        
\def\minprobn#1#2#3{$$\eqalign{&\hbox{minimize\ \ }#1\cr &\hbox{subject to\ \
}#2\cr}\eqno(#3)$$}     
\def\maxprob#1#2#3{$$\eqalign{&\hbox{maximize\ \ }#1\cr &\hbox{subject to\ \
}#2\cr}\ifnum 0=#3{}\else\eqno(#3)\fi$$}
\def\maxprobn#1#2#3{$$\eqalign{&\hbox{maximize\ \ }#1\cr &\hbox{subject to\ \
}#2\cr}\eqno(#3)$$}     
\def\aligntwo#1#2#3#4#5{$$\eqalign{#1&#2\cr #3&#4\cr}
\ifnum 0=#5{}\else\eqno(#5)\fi$$}
\def\alignthree#1#2#3#4#5#6#7{$$\eqalign{#1&#2\cr #3&#4\cr #5&#6\cr}
\ifnum 0=#7{}\else\eqno(#7)\fi$$}


\def\eqnum{\eqno{\hbox{(\the\sectnum.\the\eqnumber)}\global\advance\eqnumber
by1}}

\def\eqnu{\eqno{\hbox{(\the\sectnum.\the\eqnumber)}\global\advance\eqnumber
by1}}

\newcount\examplnumber
\def\examplnum{\global\advance\examplnumber by1}

\newcount\figrnumber
\def\figrnum{\global\advance\figrnumber by1}

\newcount\propnumber
\def\propnum{\global\advance\propnumber by1}

\newcount\defnumber
\def\defnum{\global\advance\defnumber by1}

\newcount\lemmanumber
\def\lemmanum{\global\advance\lemmanumber by1}

\newcount\assumptionnumber
\def\assumptionnum{\global\advance\assumptionnumber by1}

\def\exampl{\the\sectnum.\the\examplnumber}
\def\figr{\the\sectnum.\the\figrnumber}
\def\propn{\the\sectnum.\the\propnumber}
\def\defn{\the\sectnum.\the\defnumber}
\def\lemman{\the\sectnum.\the\lemmanumber}
\def\assumptionn{\the\sectnum.\the\assumptionnumber}

\def\section#1{\goodbreak\vskip 3pc plus 6pt minus 3pt\leftskip=-2pc
   \global\advance\sectnum by 1\eqnumber=1
\global\examplnumber=1\figrnumber=1\propnumber=1\defnumber=1\lemmanumber=1\assumptionnumber=1%
   \line{\hfuzz=1pc{\hbox to 3pc{\bf 
   \vtop{\hfuzz=1pc\hsize=38pc\hyphenpenalty=10000\noindent\uppercase{\the\sectnum.\quad #1}}\hss}}
			\hfill}
			\leftskip=0pc\nobreak\tenf
			\vskip 1pc plus 4pt minus 2pt\noindent\ignorespaces}



\def\sect#1{\noindent\leftskip=-2pc\tenf
   \goodbreak\vskip 1pc plus 4pt minus 2pt
                \global\advance\subsectnum by 1\eqnumber=1
   \line{\hfuzz=1pc{\hbox to 3pc{\bf 
   \vtop{\hfuzz=1pc\hsize=38pc\hyphenpenalty=10000\noindent\uppercase{{\bf #1}}}\hss}}
                        \hfill}
   \leftskip=0pc\nobreak\tenf
                        \vskip 1pc plus 4pt minus 2pt\nobreak\noindent\ignorespaces}

\def\subsection#1{\noindent\leftskip=0pc\tenf
   \goodbreak\vskip 1pc plus 4pt minus 2pt
   \line{\hfuzz=1pc{\hbox to 3pc{\bf 
   \vtop{\hfuzz=1pc\hsize=38pc\hyphenpenalty=10000\noindent{\bf #1}}\hss}}
                        \hfill}
   \leftskip=0pc\nobreak\tenf
                        \vskip 1pc plus 4pt minus 2pt\nobreak\noindent\ignorespaces}
\def\subsubsection#1{\goodbreak\vskip 1pc plus 4pt minus 2pt
   \hfuzz=3pc\leftskip=0pc\noindent\tenit #1 \nobreak\tenf\vskip 6pt plus 1pt
                                minus 1pt\nobreak\ignorespaces\leftskip=0pc}
%
\def\textlist#1{\par{\bf #1}\ }

\def\beginexample#1{\noindent\goodbreak\vskip 6pt plus 1pt minus 1pt
\noindent
  \hbox {\bf Example #1\hss}
  \nobreak\vskip 4pt plus 1pt minus 1pt \nobreak\noindent\ninef
  \global\advance
                \leftskip by\parindent\pn}
\def\endexample{\vskip 12pt\tenf\par
  \global\advance\leftskip by -\parindent
  }

\def\beginexercise#1{\noindent\goodbreak\vskip 6pt plus 1pt minus 1pt \noindent\global\normalbaselineskip=12pt
  \hbox {\bf Exercise #1\hss}
  \nobreak\vskip 4pt plus 1pt minus 1pt 
  \nobreak\noindent\ninef\global\advance\leftskip
                        by\parindent\pn}
\def\endexercise{\vskip 12pt\tenf\par
  \global\advance\leftskip by -\parindent
  }

\def\beginsection#1{\noindent\goodbreak\vskip 6pt plus 1pt minus 1pt \noindent\global\normalbaselineskip=12pt
  \hbox {\it #1\hss}
  \vskip 0.1pt plus 1pt minus 1pt \nobreak\noindent\ninef\global\advance
                \leftskip by\parindent\noindent\pn}
\def\endsection{\vskip 12pt\tenf\par
  \global\advance\leftskip by -\parindent
}

\def\enddisplaylist{\vskip 12pt\par}


\def\section#1{\goodbreak\vskip 3pc plus 6pt minus 3pt\leftskip=-2pc
   \global\advance\sectnum by 1\eqnumber=1
\global\examplnumber=1\figrnumber=1\propnumber=1\defnumber=1\lemmanumber=1\assumptionnumber=1\subsectnum=0%
   \line{\hfuzz=1pc{\hbox to 3pc{\bf 
   \vtop{\hfuzz=1pc\hsize=38pc\hyphenpenalty=10000\noindent\uppercase{\the\sectnum.\quad #1}}\hss}}
			\hfill}
			\leftskip=0pc\nobreak\tenf
			\vskip 1pc plus 4pt minus 2pt\noindent\ignorespaces}

\def\subsection#1{\noindent\leftskip=0pc\tenf
   \goodbreak\vskip 1pc plus 4pt minus 2pt
               \global\advance\subsectnum by 1
   \line{\hfuzz=1pc{\hbox to 3pc{\bf \the\sectnum.\the\subsectnum\ \ \
   \vtop{\hfuzz=1pc\hsize=38pc\hyphenpenalty=10000\noindent{\bf #1}}\hss}}
                        \hfill}
   \leftskip=0pc\nobreak\tenf
                        \vskip 1pc plus 4pt minus 2pt\nobreak\noindent\ignorespaces}

\def\subsubsection#1{\goodbreak\vskip 1pc plus 4pt minus 2pt
   \hfuzz=3pc\leftskip=0pc\noindent{\bf #1} \nobreak\vskip 6pt plus 1pt
                                minus 1pt\nobreak\ignorespaces\leftskip=0pc}



\def\lemma#1{\smskip\pn{\bf Lemma #1}\quad}
\def\theorem#1{\smskip\pn{\bf Theorem #1}\quad}
\def\proposition#1{\smskip\pn{\bf Proposition #1}\quad}
\def\proof{\smskip\pn{\bf Proof:}\quad} 
\def\definition#1{\smskip\pn{\bf
Definition #1}\quad} \def\assumption#1{\smskip\pn{\bf Assumption #1}\quad}
\def\corollary#1{\smskip\pn{\bf Corollary #1}\quad}
\def\exercise#1{\smskip\pn{\bf Exercise #1}\quad}
\def\figure#1{\smskip\pn{\bf Figure #1}\quad}
\def\QED{\quad{\bf Q.E.D.} \par\bigskip} \def\qed{\quad{\bf
Q.E.D.} \par\bigskip}
\def\ref{\smskip\pn}

\def\chapter#1#2{{\bf \centerline{\helbigbig
{#1}}}\bigskip\bigskip{\bf \centerline{\helbigbig
{#2}}}\bigskip\bigskip} 

\def\longchapter#1#2#3{{\bf \centerline{\helbigbig
{#1}}}\bigskip\bigskip{\bf \centerline{\helbigbig
{#2}}}\bigskip{\bf \centerline{\helbigbig
{#3}}}\bigskip\bigskip} 

\def\papertitle#1#2{{\bf \centerline{\helbigb
{#1}}}\bigskip\bigskip{\centerline{
by}}\bigskip{\bf \centerline{
{#2}}}\bigskip\bigskip} 

\def\longpapertitle#1#2#3{{\bf \centerline{\helbigb
{#1}}}\bigskip{\bf \centerline{\helbigb
{#2}}}\bigskip\bigskip{\centerline{
by}}\bigskip{\bf \centerline{
{#3}}}\bigskip\bigskip} 


\def\nitem#1{\smskip\item{#1}}
\def\nitemitem#1{\smskip\itemitem{#1}}
\def\endlist{\smskip}

\newcount\alphanum
\newcount\romnum

\def\alphaenumerate{\ifcase\alphanum \or (a)\or (b)\or (c)\or (d)\or (e)\or
(f)\or (g)\or (h)\or (i)\or (j)\or (k)\fi}
\def\romenumerate{\ifcase\romnum \or (i)\or (ii)\or (iii)\or (iv)\or (v)\or
(vi)\or (vii)\or (viii)\or (ix)\or (x)\or (xi)\fi}

\def\alist{\begingroup\vskip10pt\alphanum=1
\parskip=2pt\parindent=0pt \leftskip=3pc
\everypar{\llap{{\rm\alphaenumerate\hskip1em}}\advance\alphanum by1}}
\def\endalist{\vskip1pc\endgroup\parskip=0pt
  \parindent=\bigskipamount \leftskip=0pc\tenf
                        \noindent\ignorespaces}

\def\nolist{\begingroup\vskip10pt\alphanum=0
\parskip=2pt\parindent=0pt \leftskip=3pc
\everypar{\llap{\global\advance\alphanum by1(\the\alphanum)\hskip1em}}}
\def\endnolist{\vskip1pc\endgroup\parskip=0pt\leftskip=0pc\tenf
                        \noindent\ignorespaces}

\def\romlist{\begingroup\vskip10pt\romnum=1
\parskip=2pt\parindent=0pt \leftskip=5pc
\everypar{\llap{{\rm\romenumerate\hskip1em}}\advance\romnum by1}}
\def\endromlist{\vskip1pc\endgroup\parskip=0pt\leftskip=0pc\tenf
                        \noindent\ignorespaces}


\long\def\fig#1#2#3{\vbox{\vskip1pc\vskip#1
\prevdepth=12pt \baselineskip=12pt
\vskip1pc
\hbox to\hsize{\hfill\vtop{\hsize=25pc\noindent{\eightbf Figure #2\ }
{\eightpoint#3}}\hfill}}}

\long\def\widefig#1#2#3{\vbox{\vskip1pc\vskip#1
\prevdepth=12pt \baselineskip=12pt
\vskip1pc
\hbox to\hsize{\hfill\vtop{\hsize=28pc\noindent{\eightbf Figure #2\ }
{\eightpoint#3}}\hfill}}}

\long\def\table#1#2{\vbox{\vskip0.5pc
\prevdepth=12pt \baselineskip=12pt
\hbox to\hsize{\hfill\vtop{\hsize=25pc\noindent{\eightbf Table #1\ }
{\eightpoint#2}}\hfill}}}

\def\rightleftheadline#1#2{ifodd\pageno\rightheadline{#1}
\else\leftheadline{#2}\fi} 
\def\rightheadline#1{\headline{\tenrm\hfil #1}}
\def\leftheadline#1{\headline{\tenrm#1\hfil}}


\long\def\leftfig#1#2{\vbox{\smskip\hsize=220pt
\vtop{{\noindent {\bf #1}}}
\smskip
\noindent
\vbox{{\noindent #2}}
}}

\long\def\rightfig#1#2#3{\vbox{\smskip\vskip#1
\prevdepth=12pt \baselineskip=12pt
\hsize=210pt
\smskip
\vbox{\noindent{\eightbold #2}
\hskip1em{\eightpoint#3}}
}}

\long\def\concept#1#2#3#4#5{\bigskip\hrule
\vbox{\hbox{\leftfig{#1}{#2} \hskip3em
\rightfig{#3}{#4}{#5}} \smskip}
\hrule\bigskip}


\long\def\bconcept#1#2#3#4#5#6#7{
\vbox{
\hbox to \hsize{\vtop{\par #1}}
\concept{#2}{#3}{#4}{#5}{#6}
\hbox to \hsize{\vtop{\par #7}}
\smskip}
}


\long\def\boxconcept#1#2#3#4#5{
\vbox{\hbox{\leftfig{#1}{#2} \hskip3em
\rightfig{#3}{#4}{#5}} \smskip}
}


\def\boxit#1{\vbox{\hrule\hbox{\vrule\kern3pt
                                \vbox{\kern3pt#1\kern3pt}\kern3pt\vrule}\hrule}}
\def\centerboxit#1{$$\vbox{\hrule\hbox{\vrule\kern3pt
                                \vbox{\kern3pt#1\kern3pt}\kern3pt\vrule}\hrule}$$}

\long\def\boxtext#1#2{$$\boxit{\vbox{\hsize #1\noindent\strut #2\strut}}$$}

%
%
%

\def\picture #1 by #2 (#3){
  \vbox to #2{
    \hrule width #1 height 0pt depth 0pt
    \vfill
    \special{picture #3} 
    }
  }

\def\scaledpicture #1 by #2 (#3 scaled #4){{
  \dimen0=#1 \dimen1=#2
  \divide\dimen0 by 1000 \multiply\dimen0 by #4
  \divide\dimen1 by 1000 \multiply\dimen1 by #4
  \picture \dimen0 by \dimen1 (#3 scaled #4)}
  }

%
%

\long\def\captfig#1#2#3#4#5{\vbox{\vskip1pc
\hbox to\hsize{\hfill{\picture #1 by #2 (#3)}\hfill}
\prevdepth=9pt \baselineskip=9pt
\vskip1pc
\hbox to\hsize{\hfill\vtop{\hsize=24pc\noindent{\eightbold Figure #4}
\hskip1em{\eightpoint#5}}\hfill}}}

%
%
%

\def\illustration #1 by #2 (#3){
  \vskip#2\hskip#1\special{illustration #3} 
    }

\def\scaledillustration #1 by #2 (#3 scaled #4){{
  \dimen0=#1 \dimen1=#2
  \divide\dimen0 by 1000 \multiply\dimen0 by #4
  \divide\dimen1 by 1000 \multiply\dimen1 by #4
  \illustration \dimen0 by \dimen1 (#3 scaled #4)}
  }


\font\hel=cmr10%
\font\helb=cmbx10%
\font\heli=cmti10%
\font\helbi=cmsl10%
\font\ninehel=cmr9%
\font\nineheli=cmti9%
\font\ninehelb=cmbx9%
\font\helbig=cmr10 scaled 1500%
\font\helbigbig=cmr10 scaled 2500%
\font\helbigb=cmbx10 scaled 1500%
\font\helbigbigb=cmbx10 scaled 2500%
\font\bigi=cmti10 scaled \magstep5%
\font\eightbold=cmbx8%
\font\ninebold=cmbx9%
\font\boldten=cmbx10%

\def\tenf{\hel}%
\def\tenit{\heli}%
\def\ninef{\ninehel}%
\def\nineb{\ninehelb}%
\def\nineit{\nineheli}%
\def\smit{\nineheli}%
\def\smbf{\ninehelb}%


\font\tenrm=cmr10%
\font\teni=cmmi10%
\font\tensy=cmsy10%
\font\tenbf=cmbx10%
\font\tentt=cmtt10%
\font\tenit=cmti10%
\font\tensl=cmsl10%

\def\tenpoint{\def\rm{\fam0\tenrm}%
\textfont0=\tenrm%
\textfont1=\teni%
\textfont2=\tensy%
\textfont\itfam=\tenit%
\textfont\slfam=\tensl%
\textfont\ttfam=\tentt%
\textfont\bffam=\tenbf%
\scriptfont0=\sevenrm%
\scriptfont1=\seveni%
\scriptfont2=\sevensy%
\scriptscriptfont0=\sixrm%
\scriptscriptfont1=\sixi%
\scriptscriptfont2=\sixsy%
\def\it{\fam\itfam\tenit}%
\def\tt{\fam\ttfam\tentt}%
\def\sl{\fam\slfam\tensl}%
\scriptfont\bffam=\sevenbf%
\scriptscriptfont\bffam=\sixbf%
\def\bf{\fam\bffam\tenbf}%
\normalbaselineskip=18pt%
\normalbaselines\rm}%

\font\ninerm=cmr9%
\font\ninebf=cmbx9%
\font\nineit=cmti9%
\font\ninesy=cmsy9%
\font\ninei=cmmi9%
\font\ninett=cmtt9%
\font\ninesl=cmsl9%

\def\ninepoint{\def\rm{\fam0\ninerm}%
\textfont0=\ninerm%
\textfont1=\ninei%
\textfont2=\ninesy%
\textfont\itfam=\nineit%
\textfont\slfam=\ninesl%
\textfont\ttfam=\ninett%
\textfont\bffam=\ninebf%
\scriptfont0=\sixrm%
\scriptfont1=\sixi%
\scriptfont2=\sixsy%
\def\it{\fam\itfam\nineit}%
\def\tt{\fam\ttfam\ninett}%
\def\sl{\fam\slfam\ninesl}%
\scriptfont\bffam=\sixbf%
\scriptscriptfont\bffam=\fivebf%
\def\bf{\fam\bffam\ninebf}%
\normalbaselineskip=16pt%
\normalbaselines\rm}%

\font\eightrm=cmr8%
\font\eighti=cmmi8%
\font\eightsy=cmsy8%
\font\eightbf=cmbx8%
\font\eighttt=cmtt8%
\font\eightit=cmti8%
\font\eightsl=cmsl8%

\def\eightpoint{\def\rm{\fam0\eightrm}%
\textfont0=\eightrm%
\textfont1=\eighti%
\textfont2=\eightsy%
\textfont\itfam=\eightit%
\textfont\slfam=\eightsl%
\textfont\ttfam=\eighttt%
\textfont\bffam=\eightbf%
\scriptfont0=\sixrm%
\scriptfont1=\sixi%
\scriptfont2=\sixsy%
\scriptscriptfont0=\fiverm%
\scriptscriptfont1=\fivei%
\scriptscriptfont2=\fivesy%
\def\it{\fam\itfam\eightit}%
\def\tt{\fam\ttfam\eighttt}%
\def\sl{\fam\slfam\eightsl}%
\scriptscriptfont\bffam=\fivebf%
\def\bf{\fam\bffam\eightbf}%
\normalbaselineskip=14pt%
\normalbaselines\rm}%

\font\sevenrm=cmr7%
\font\seveni=cmmi7%
\font\sevensy=cmsy7%
\font\sevenbf=cmbx7%
\font\seventt=cmtt8 at 7pt%
\font\sevenit=cmti8 at 7pt%
\font\sevensl=cmsl8 at 7pt%

\def\sevenpoint{%
   \def\rm{\sevenrm}\def\bf{\sevenbf}%
   \def\smc{\sevensmc}\baselineskip=12pt\rm}%

\font\sixrm=cmr6%
\font\sixi=cmmi6%
\font\sixsy=cmsy6%
\font\sixbf=cmbx6%
\font\sixtt=cmtt8 at 6pt%
\font\sixit=cmti8 at 6pt%
\font\sixsl=cmsl8 at 6pt%
\font\sixsmc=cmr8 at 6pt%

\def\sixpoint{%
   \def\rm{\sixrm}\def\bf{\sixbf}%
   \def\smc{\sixsmc}\baselineskip=12pt\rm}%

\fontdimen13\tensy=2.6pt%
\fontdimen14\tensy=2.6pt%
\fontdimen15\tensy=2.6pt%
\fontdimen16\tensy=1.2pt%
\fontdimen17\tensy=1.2pt%
\fontdimen18\tensy=1.2pt%

\def\tenf{\tenpoint}%
\def\ninef{\ninepoint}%
\def\eightf{\eightpoint}%


\def\texshopbox#1{\boxtext{462pt}{\vskip-1.5pc\pshade{\vskip-1.0pc#1\vskip-2.0pc}}}
\def\texshopboxnt#1{\boxtextnt{462pt}{\vskip-1.5pc\pshade{\vskip-1.0pc#1\vskip-2.0pc}}}
\def\texshopboxnb#1{\boxtextnb{462pt}{\vskip-1.5pc\pshade{\vskip-1.0pc#1\vskip-2.0pc}}}


\input miniltx

\ifx\pdfoutput\undefined
  \def\Gin@driver{dvips.def} 
\else
  \def\Gin@driver{pdftex.def} 
\fi

\input graphicx.sty
\resetatcatcode

\long\def\fig#1#2#3{\vbox{\vskip1pc\vskip#1
\prevdepth=12pt \baselineskip=12pt
\vskip1pc
\hbox to\hsize{\hskip3pc\hfill\vtop{\hsize=35pc\noindent{\eightbf Figure #2\ }
{\eightpoint#3}}\hfill}}}

\def\show#1{}
\def\dpbshow#1{#1}
\def\frac#1#2{{#1\over #2}}

\rightheadline{\botmark}

\pageno=1

\def\longpapertitle#1#2#3{{\bf \centerline{\helbigb
{#1}}}\medskip{\bf \centerline{\helbigb
{#2}}}\bigskip{\bf \centerline{
{#3}}}\bigskip}

\vskip-3pc

\def\xstar{X^{\raise0.04pt\hbox{\sevenpoint *}} }

\def\jstar{J^{\raise0.04pt\hbox{\sevenpoint *}} }
\def\qstar{Q^{\raise0.04pt\hbox{\sevenpoint *}} }
\def\ebar{ \bar{E}(X )}
\def\bbar{ \bar{B}(X )}
\def\ehat{ \hat{E}(X )}
\def\mbar{ \overline{\cal M}}

\def\rstar{\re^*}

\rightheadline{\botmark}

\pageno=1

\rightheadline{\botmark}

\pn {\bf April 2018 (revised August 2018)}\hfill{\bf MIT/LIDS Report}%
\bigskip \bigskip\bigskip

\bigskip

\def\longpapertitle#1#2#3{{\bf \centerline{\helbigb
{#1}}}\medskip{\bf \centerline{\helbigb
{#2}}}\bigskip{\bf \centerline{
{#3}}}\bigskip}

\vskip-2pc

\longpapertitle{Feature-Based Aggregation and Deep Reinforcement Learning:}{A Survey and Some New Implementations}{ {Dimitri P.\ Bertsekas\footnote{\dag}{\ninepoint Dimitri Bertsekas is with the Dept.\ of Electr.\ Engineering and
Comp.\ Science, and the Laboratory for Information and Decision Systems, M.I.T., Cambridge, Mass., 02139.  A version of this paper will appear in IEEE/CAA Journal of Automatica Sinica.}  }}

\centerline{\bf Abstract}

In this paper we discuss policy iteration methods for approximate solution of a finite-state discounted Markov decision problem, with a focus on feature-based aggregation methods and their connection with deep reinforcement learning schemes. We introduce features of the states of the original problem, and we formulate a smaller ``aggregate" Markov decision problem, whose states relate to the features. We discuss properties and possible implementations of this type of aggregation, including a new approach to approximate policy iteration. In this approach the policy improvement operation combines feature-based aggregation with  feature construction using deep neural networks or other calculations. We argue that the cost function of a policy may be approximated much more accurately by the nonlinear function of the features provided by aggregation, than by the linear function of the features provided by neural network-based reinforcement learning, thereby potentially leading to more effective policy improvement. 


\def\jstar{J^{\raise0.04pt\hbox{\sevenpoint *}} }

\vfill\eject

\centerline{\bf Contents}
\bigskip

\nitem{1.} Introduction
\nitemitem{1.1} Alternative Approximate Policy Iteration Methods
\nitemitem{1.2} Terminology
\nitem{2.} Approximate Policy Iteration: An Overview
\nitemitem{2.1} Direct and Indirect Approximation Approaches for Policy Evaluation
\nitemitem{2.2} Indirect Methods Based on Projected Equations
\nitemitem{2.3} Indirect Methods Based on Aggregation
\nitemitem{2.4} Implementation Issues
\nitem{3.} Approximate Policy Evaluation Based on Neural Networks
\nitem{4.} Feature-Based Aggregation Framework
\nitemitem{4.1} The Aggregate Problem
\nitemitem{4.2} Solving the Aggregate Problem with Simulation-Based Methods
\nitemitem{4.3} Feature Formation by Using Scoring Functions
\nitemitem{4.4} Using Heuristics to Generate Features - Deterministic Optimization and Rollout
\nitemitem{4.5} Stochastic Shortest Path Problems - Illustrative Examples
\nitemitem{4.6} Multistep Aggregation
\nitem{5.} Policy Iteration with Feature-Based Aggregation and a Neural Network
\nitem{6.} Concluding Remarks
\nitem{7.} References

\vfill\eject

\section{Introduction}

\vskip-0.5pc

\pn We consider a discounted infinite horizon dynamic programming (DP) problem with $n$ states,  which we denote by $i=1,\ldots,n$. State transitions $(i,j)$ under control $u$ occur at discrete times according to transition probabilities $p_{ij}(u)$, and generate a cost $\a^k g(i,u,j)$ at time $k$, where $\a\in(0,1)$ is the discount factor. We consider deterministic stationary policies $\m$ such that for each $i$, $\m(i)$ is a control that belongs to a constraint set $U(i)$. We denote by $J_\m(i)$ the total discounted expected cost  of $\m$ over an infinite number of stages starting from state $i$, and by $\jstar(i)$ the minimal value of $J_\m(i)$ over all $\m$. We denote by $J_\m$ and $\jstar $ the $n$-dimensional vectors that have components $J_\m(i)$ and $\jstar(i)$, $i=1,\ldots,n$, respectively. As is well known, $J_\m$ is the unique solution of the Bellman equation for policy $\m$:
 $$J_\m(i) = \sum_{j=1}^n
p_{ij}\bl(\mu(i)\br)\Big(g\bl(i,\mu(i),j\br) + \a J_\m(j)\Big),\qquad  i=1,\ldots,n,\xdef\onefi{\lab}\eqnum\show{oneo}$$
while $\jstar $ is the unique solution of the Bellman equation
$$\jstar (i) = \min_{u\in U(i)}\sum_{j=1}^n p_{ij}(u)\big(g(i,u,j) + \a \jstar (j)\big),
\qquad i=1,\ldots,n.\xdef\onef{\lab}\eqnum\show{oneo}$$

\xdef\figapi{\figr}\figrnum\show{myfigure}

In this  paper, we survey several ideas from aggregation-based approximate DP and deep reinforcement learning, all of which have been essentially known for some time, but are combined here in a new way. We will focus on methods of approximate policy iteration (PI for short), whereby we  evaluate approximately the cost vector $J_\m$ of each generated policy $\m$. Our cost approximations use a feature vector
$F(i)$ of each state $i$, and replace $J_\m(i)$ with a function that depends on $i$ through $F(i)$, i.e., a function of the form
$$\hat J_\m\big(F(i)\big)\approx J_\m(i),\qquad i=1,\ldots,n.$$
We refer to such $\hat J_\m$ as a {\it feature-based approximation architecture\/}. 

At the typical iteration of our approximate PI methodology, the cost vector  $J_\m$ of the current policy $\m$ is approximated using a feature-based architecture $\hat J_\m$, and a new policy $\hat \m$ is then generated using a policy ``improvement" procedure; see Fig.\ \figapi. The salient characteristics of our approach are two:

\nitem{(a)} The feature vector $F(i)$ may be obtained using a neural network or other calculation that automatically constructs features.

\nitem{(b)} The policy improvement, which generates  $\hat \m$ is based on a DP problem that involves feature-based aggregation. 

\pn By contrast, the standard policy improvement method is based on the one-step lookahead minimization
$$\hat \m(i)\in\arg\min_{u\in U(i)}\sum_{j=1}^n p_{ij}(u)\Big(g(i,u,j) + \a \hat J_\m\big(F(j)\big)\Big),
\qquad i=1,\ldots,n,\xdef\polimprove{\lab}\eqnum\show{oneo}$$
or alternatively, on multistep lookahead, possibly combined with Monte-Carlo tree search. We will argue that our feature-based aggregation approach has the potential to generate far better policies at the expense of a more computation-intensive policy improvement phase. 

\topinsert
\centerline{\hskip0pc\includegraphics[width=5in]{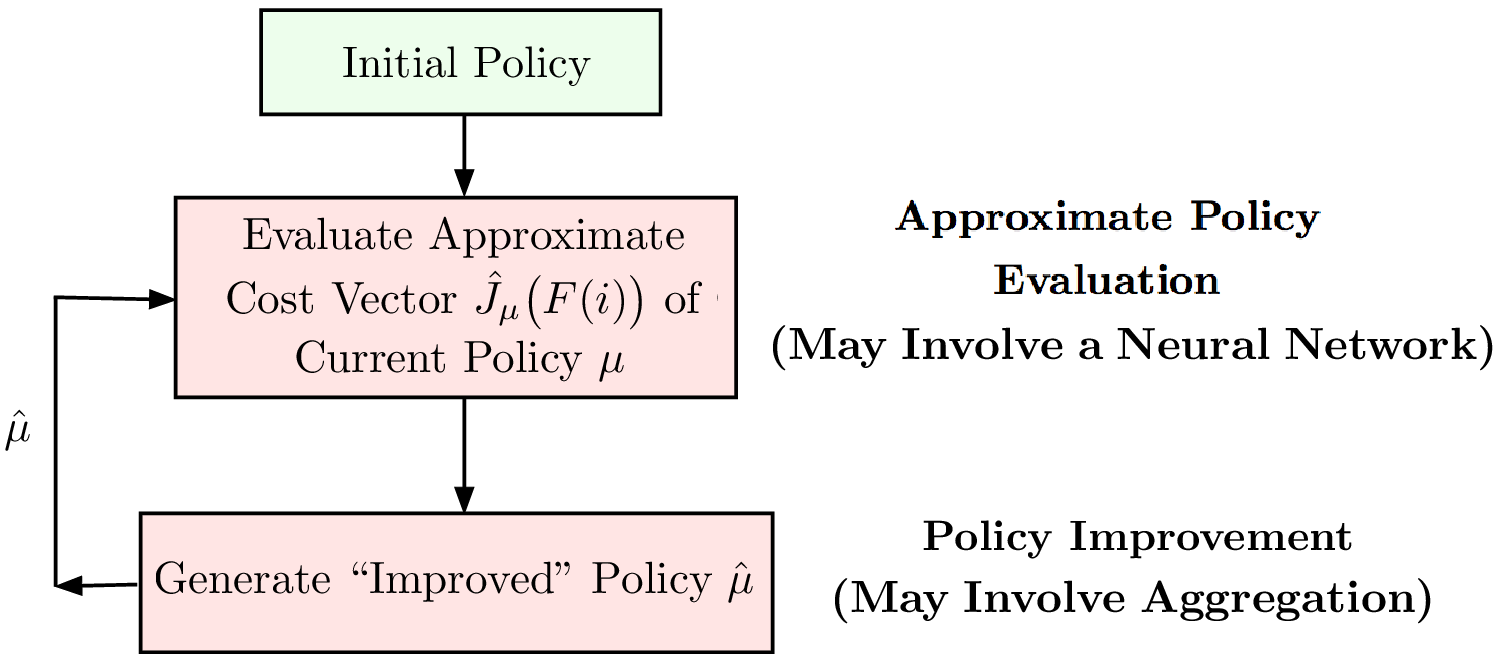}}\vskip0pc
\hskip-3pc\fig{0pc}{\figapi} {Schematic view of feature-based approximate PI. The cost  $J_\m(i)$ of the current policy $\m$ starting from state $i$ is replaced by an approximation $\skew5\hat J_\m\big(F(i)\big)$ that depends on $i$ through its feature vector $F(i)$. The feature vector is assumed independent of the current policy $\m$ in this figure, but in general could depend on $\m$.}
\endinsert

\subsection{Alternative Approximate Policy Iteration Methods}

\pn A survey of approximate PI methods was given in 2011 by the author [Ber11a], and focused on {\it linear feature-based architectures\/}. These are architectures where $F(i)$ is an $s$-dimensional vector 
$$F(i)=\big(F_1(i),\ldots,F_s(i)\big),$$
and $\hat J_\m$ depends linearly on $F$, i.e.,
$$\hat J_\m\big(F(i)\big)=\sum_{\ell=1}^s F_\ell(i)r_\ell,\qquad i=1,\ldots,n,$$
for some scalar weights $r_1,\ldots,r_s$.
We considered in  [Ber11a] two types of methods:
\nitem{(a)} Projected equation methods, including temporal difference methods, where policy evaluation is based on simulation-based matrix inversion methods such as LSTD($\l$), or stochastic iterative methods such as TD($\l$), or variants of $\l$-policy iteration such as LSPE($\l$)]. 

\nitem{(b)} General aggregation methods (not just the  feature-based type considered here). 
\smskip

\pn These methods will be briefly discussed in Section 2.
The present paper is complementary to the survey [Ber11a], and deals with approximate PI with nonlinear  feature-based architectures, including some where features are generated with the aid of neural networks or some other heuristic calculations.

An important advantage of linear feature-based architectures is that given the form of the feature vector $F(\cdot)$, they can be trained with linear least squares-type methods. However, determining good features may be a challenge in general. Neural networks resolve this challenge through training that constructs automatically features and simultaneously combines the components of the features linearly with weights. This is commonly done by cost fitting/nonlinear regression, using a large number of state-cost sample pairs, which are processed through a sequence of alternately linear and nonlinear layers (see Section 3). The outputs of the final nonlinear layer are the features, which are then processed by a final linear layer that provides a linear combination of the features as a cost function approximation. 

The idea of representing cost functions in terms of features of the state in a context that we may now call ``approximation in value space" or ``approximate  DP" goes back to the work of Shannon on chess [Sha50]. The work of Samuel [Sam59], [Sam67] on checkers extended some of Shannon's algorithmic schemes and introduced  {\it temporal difference} ideas that motivated much subsequent research. The use of neural networks to simultaneously extract features of the optimal or the policy cost functions, and construct an approximation to these cost functions was also investigated in the early days of reinforcement learning; some of the original contributions that served as motivation for much subsequent work are Werbos [Wer77], Barto, Sutton, and Anderson [BSA83], Christensen and Korf [ChK86], Holland [Hol86], and Sutton [Sut88]. The use of a neural network as a cost function approximator for a challenging DP problem was first demonstrated impressively in the context of the game of backgammon by Tesauro [Tes92], [Tes94], [Tes95], [Tes02]. In Tesauro's work the parameters of the network were trained by using a form of temporal differences (TD) learning, and the features constructed by the neural network were supplemented by some handcrafted features.\footnote{\dag}{\ninepoint Tesauro also constructed a different backgammon player, trained by a neural network, but with a supervised learning approach, which used examples from human expert play [Tes89a], [Tes89b] (he called this approach ``comparison learning"). However, his TD-based algorithm performed substantially better, and its success has been replicated by others, in both research and commercial programs. Tesauro and Galperin [TeG96] proposed still another approach to backgammon, based on a rollout strategy, which resulted in an even better playing program (see [Ber17] for an extensive discussion of rollout as a general approximate DP approach). At present, rollout-based backgammon programs are viewed as the most powerful in terms of performance, but are too time-consuming for real-time play. They have been used in a limited diagnostic way to assess the quality of neural network-based programs. A list of articles on computer backgammon may be found at http://www.bkgm.com/articles/page07.html.}

Following Tesauro's work, the synergistic potential of approximations using neural network or other architectures, and DP techniques had become apparent, and it was laid out in an influential survey paper by Barto, Bradtke, and Singh [BBS95]. It was then systematically developed in the neuro-dynamic programming book by Bertsekas and Tsitsiklis [BeT96], and the reinforcement learning book by Sutton and Barto [SuB98]. Subsequent books on approximate DP and reinforcement learning, which discuss approximate PI, among other techniques, include Cao [Cao07], Busoniu et.\ al.\ [BBD10], Szepesvari [Sze10], Powell [Pow11],  Chang, Fu, Hu, and Marcus [CFH13], Vrabie, Vamvouda\-kis, and Lewis [VVL13], and Gosavi [Gos15]. To these, we may add the edited collections by Si, Barto, Powell, and Wunsch [SBP04], Lewis, Liu, and Lendaris [LLL08], and Lewis and Liu [LeL12], which contain several survey papers.

The original ideas on approximate PI were enriched by further research ideas such as {\it rollout} (Abramson [Abr90], Tesauro and Galperin [TeG96],  Bertsekas,
Tsitsiklis, and Wu [BTW97], Bertsekas and Castanon [BeC99]; see the surveys in [Ber13], [Ber17]), {\it adaptive simulation and Monte Carlo tree search} (Chang, Hu, Fu, and Marcus [CFH05], [CFH13], Coulom [Cou06]; see the survey by Browne et al.\ [BPW12]), and {\it deep neural networks} (which are neural networks with many and suitably specialized layers; see for the example the book by Goodfellow, Bengio, and Courville [GBC16], the textbook discussion in [Ber17], Ch.\ 6, and the recent surveys by Schmidhuber [Sch15], Arulkumaran et al.\ [ADB17], Liu et al.\ [LWL17], and Li [Li17]). 

\old{
The original ideas on approximate PI were enriched by further methodological research, including:
\nitem{(a)} {\it Rollout} (Abramson [Abr90], Tesauro and Galperin [TeG96],  Bertsekas,
Tsitsiklis, and Wu [BTW97], Bertsekas and Castanon [BeC99]); see the surveys in [Ber13], [Ber17].
\nitem{(b)} {\it  Adaptive simulation and Monte Carlo tree search} (Chang, Hu, Fu, and Marcus [CFH05], [CFH13], Coulom [Cou06]); see the survey by Browne et al.\ [BPW12].
\nitem{(c)} {\it Deep neural networks\/}, which are neural networks with many and suitably specialized layers; see for the example the book by Goodfellow, Bengio, and Courville [GBC16], the textbook discussion in [Ber17], Ch.\ 6, and the recent surveys by Schmidhuber [Sch15], Arulkumaran et al.\ [ADB17], Liu et al.\ [LWL17], and Li [Li17]. 
\smskip
}

A recent impressive success of the deep neural network-based approximate PI methodology is the AlphaZero program, which attained a superhuman level of play for the games of chess, Go, and others (see Silver et al.\ [SHS17]). A noteworthy characteristic of this program is that it does not use domain-specific knowledge (i.e., handcrafted features), but rather relies entirely on the deep neural network to construct features for cost function approximation (at least as reported in [SHS17]). Whether it is advisable to rely exclusively on the neural network to provide features is an open question, as other investigations, including the ones by Tesauro noted earlier, suggest that using additional problem-specific hand-crafted features can be very helpful in the context of approximate DP. Except for the use of deep rather than shallow neural networks (which are used in backgammon), the AlphaZero algorithm is similar to several other algorithms that have been proposed in the literature and/or have been developed in the past. It can be viewed as a conceptually straightforward implementation of approximate PI, using Monte Carlo tree search and a single neural network to construct a cost and policy approximation, and does not rely on any fundamentally new ideas or insightful theoretical analysis. Conceptually, it bears considerable similarity to Tesauro's TD-Gammon program. Its spectacular success may be attributed to the skillful implementation of an effective mix of known ideas, coupled with great computational power.

We note that the ability to simultaneously extract features and optimize their linear combination is not unique to neural networks. Other approaches that use a multilayer architecture have been proposed (see the survey by Schmidhuber [Sch15]), including the Group Method for Data Handling (GMDH), which is principally based on the use of polynomial (rather than sigmoidal) nonlinearities. The GMDH method was  investigated extensively in the Soviet Union starting with the work of Ivakhnenko in the late 60s; see e.g., [Iva68]. It has been used in a large variety of applications, and its similarities with the neural network methodology have been noted (see the survey by Ivakhnenko [Iva71], and the large literature summary at the web site http://www.gmdh.net). Most of the GMDH research relates to inference-type problems. We are unaware of any application of GMDH in the context of approximate DP, but we believe this to be a fruitful area of investigation. In any case, the feature-based PI ideas of the present paper apply equally well in conjunction with GMDH networks as with the neural networks described in Section 3.

While automatic feature extraction  is a critically important aspect of neural network architectures, the linearity of the combination of the feature components at the final layer may be a limitation. A nonlinear alternative is based on aggregation, a dimensionality reduction approach to address large-scale problems. This approach has a long history in scientific computation and operations research (see for example Bean, Birge, and Smith [BBS87], Chatelin and Miranker [ChM82], Douglas and Douglas [DoD93], Mendelssohn [Men82], and Rogers et.\ al.\ [RPW91]). It was introduced in the simulation-ba\-sed approximate DP context, mostly in the form of value iteration; see Singh, Jaakkola, and Jordan [SJJ95], Gordon [Gor95], Tsitsiklis and Van Roy [TsV96] (see also the book [BeT96], Sections 3.1.2 and 6.7).  More recently, aggregation was discussed in a reinforcement learning context involving the notion of ``options" by Ciosek and Silver [CiS15], and the notion of ``bottleneck simulator" by Serban et.\ al.\ [SSP18]; in both cases encouraging computational results were presented. Aggregation architectures based on features were discussed in Section 3.1.2 of the neuro-dynamic programming book [BeT96], and in Section 6.5 of the author's DP book [Ber12] (and earlier editions), including the feature-based architecture that is the focus of the present paper. They have the capability to produce policy cost function approximations that are nonlinear functions of the feature components, thus yielding potentially more accurate approximations. Basically, in feature-based aggregation the original problem is approximated by a problem that involves a relatively small number of ``feature states."

Feature-based aggregation assumes a given form of feature vector, so for problems where good features are not apparent, it needs to be modified or to be supplemented by a method that can construct features from training data. Motivated by the reported successes of deep reinforcement learning with neural networks, we propose a two-stage process: first use a neural network or other scheme to construct good features for cost approximation, and then use there features to construct a nonlinear feature-based aggregation architecture. In effect we are proposing a new way to implement approximate PI: {\it retain the policy evaluation phase which uses a neural network or alternative scheme, but replace the policy improvement phase with the solution of an aggregate DP problem\/}. This DP problem involves the features that are generated by a neural network or other scheme (possibly together with other handcrafted features). Its dimension may be reduced to a manageable level by sampling, while its cost function values are generalized to the entire feature space by linear interpolation. In summary, our suggested policy improvement phase may be more complicated, but may be far more powerful as it relies on the potentially more accurate function approximation provided by a nonlinear combination of features.

Aside from the power brought to bear by nonlinearly combining features, let us also note some other advantages that are generic to aggregation. In particular:

\nitem{(a)} Aggregation aims to solve an ``aggregate" DP problem, itself an approximation of the original DP problem, in the spirit of coarse-grid discretization of large state space problems. As a result, aggregation methods enjoy the stability and policy convergence guarantee of exact PI. By contrast, temporal difference-based and other PI methods can suffer from  convergence difficulties such as policy oscillations and chattering (see e.g., [BeT96], [Ber11a], [Ber12]). A corollary to this is that when an aggregation scheme performs poorly, it is easy to identify the cause: it is the quantization error due to approximating a large state space with a smaller ``aggregate" space. The possible directions for improvement (at a computational cost of course) are then clear: introduce additional aggregate states, and increase/improve these features. 

\nitem{(b)} Aggregation methods are characterized by error bounds, which are generic to PI methods that guarantee the convergence of the generated policies. These error bounds are better by a factor $(1-\a)$ compared to the corresponding error bounds for methods where policies need not converge, such as generic temporal difference methods with linear cost function approximation [see Eqs.\ (2.2) and (2.3) in the next section].
\smskip

Let us finally note that the idea of using a deep neural network to extract features for use in another approximation architecture has been used earlier. In particular, it is central in the Deepchess program by  David, Netanyahu, and Wolf [DNW16], which was estimated to perform at the level of a strong grandmaster, and at the level of some of the strongest computer chess programs. In this work the features were used, in conjunction with supervised learning and human grandmaster play selections, to train a deep neural network to compare any pair of legal moves in a given chess position, in the spirit of Tesauro's comparison training approach [Tes89b]. By contrast in our proposal the features are used to formulate an aggregate DP problem, which can be solved by exact methods, including some that are based on simulation. 

The paper is organized as follows. In Section 2, we provide context for the subsequent developments, and summarize some of the implementation issues in approximate PI methods. In Section 3, we review some of the central ideas of approximate PI based on neural networks. In Section 4, we discuss PI ideas based on feature-based aggregation, assuming good features are known. In this section, we also discuss how features may be constructed on one or more ``scoring functions," which are estimates of the cost function of a policy, provided by a neural network or a heuristic. We also pay special attention to deterministic discrete optimization problems. Finally, in Section 5, we describe some of the ways to combine the feature extraction capability of deep neural networks with the nonlinear approximation possibilities offered by aggregation.

\vskip-0.5pc

\subsection{Terminology}
\vskip-0.5pc

\pn The success of approximate DP in addressing challenging large-scale applications owes much to an enormously beneficial cross-fertilization of ideas from decision and control, and from artificial intelligence. The boundaries between these fields are now diminished thanks to a deeper understanding of the foundational issues, and the associated methods and core applications. Unfortunately, however, there have been substantial discrepancies of notation and terminology between the artificial intelligence and the optimization/decision/control fields, including the typical use of maximization/value function/reward in the former field and the use of minimization/cost function/cost per stage in the latter field. The notation and terminology used in this paper is standard in DP and optimal control, and in an effort to forestall confusion of  readers that are accustomed to either the reinforcement learning or the optimal control terminology, we provide a list of selected terms commonly used in reinforcement learning (for example in the popular book by Sutton and Barto [SuB98], and its 2018 on-line 2nd edition), and their optimal control counterparts.

\nitem{(a)} Agent = Controller or decision maker.

\nitem{(b)} Action = Control.

\nitem{(c)} Environment = System.

\nitem{(d)} Reward of a stage = (Opposite of) Cost of a stage.

\nitem{(e)} State value = (Opposite of) Cost of a state.

\nitem{(f)} Value (or state-value) function = (Opposite of) Cost function.

\nitem{(g)} Maximizing the value function = Minimizing the cost function.

\nitem{(h)} Action (or state-action) value = $Q$-factor of a state-control pair.

\nitem{(i)} Planning  = Solving a DP problem with a known mathematical model.

\nitem{(j)} Learning  = Solving a DP problem in model-free fashion.

\nitem{(k)} Self-learning  (or self-play in the context of games) = Solving a DP problem using policy iteration.

\nitem{(l)} Deep reinforcement learning = Approximate DP using value and/or policy approximation with deep neural networks.

\nitem{(m)} Prediction = Policy evaluation. 

\nitem{(n)} Generalized policy iteration = Optimistic policy iteration.

\nitem{(o)} State abstraction = Aggregation.

\nitem{(p)} Episodic task or episode = Finite-step system trajectory.

\nitem{(q)} Continuing task = Infinite-step system trajectory.

\nitem{(r)} Afterstate = Post-decision state.


\section{Approximate Policy Iteration: An Overview}


\pn Many approximate DP algorithms are based on the principles of PI: the policy evaluation/policy improvement structure of PI is maintained, but the policy evaluation is done approximately, using simulation and some approximation architecture.
In the standard form of the method, at each iteration, we compute an approximation $\tl J_\m(\cdot,r)$ to the cost function
$J_\m$ of the current policy $\m$, and we  generate an ``improved" policy $\hat \m$ using\footnote{\dag}{\ninepoint The minimization in the policy improvement phase may alternatively involve multistep lookahead, possibly combined with Monte-Carlo tree search. It may also be done approximately through $Q$-factor approximations. Our discussion extends straightfowardly to schemes that include multistep lookahead or approximate policy improvement.} 
$$\hat \m(i)\in\arg\min_{u\in U(i)}\sum_{j=1}^n p_{ij}(u)\bl(g(i,u,j)+\a\tl J_\m(j,r)\br),\qquad i=1,\ldots,n.\xdef\thrtene{\lab}\eqnum\show{oneo}$$
Here $\tl J_\m$ is a function of some chosen form (the {\it approximation architecture\/}), which depends on the state and on a parameter vector 
$r=(r_1,\ldots,r_s)$ of relatively small dimension $s$.
 
The theoretical basis for the method was discussed in the neuro-dynamic programming book [BeT96], Prop.\ 6.2 (see also [Ber12], Section 2.5.6, or [Ber18a], Sections 2.4.1 and 2.4.2). It was shown there that if
the policy evaluation is accurate to within $\d$ (in the sup-norm sense), then for an $\a$-discounted problem, the method, while not convergent, is stable in the sense that it
will yield in the limit (after infinitely many policy evaluations) stationary policies that are optimal to within
$${2\a \d\over (1-\a)^2},\xdef\bound{\lab}\eqnum\show{oneo}$$
where $\a$ is the discount factor. Moreover, if the generated sequence of policies actually converges to some $\bar\m$, then $\bar\m$ is optimal to within 
$${2\a \d\over 1-\a}\xdef\betterbound{\lab}\eqnum\show{oneo}$$ (see [BeT96], Section 6.4.1); this is a significantly improved error bound. In general, policy convergence may not be guaranteed, although it is guaranteed for the aggregation methods of this paper. Experimental evidence indicates that these bounds are often
conservative, with just a few policy iterations needed before most of the eventual cost improvement is achieved. 

\def\cov{{\rm cov}} 
\def\var{{\rm var}}
\def\O{\Omega}

\subsection{Direct and Indirect Approximation Approaches for Policy Evaluation}


\xdef\figsofi{\figr}\figrnum\show{myfigure}

\pn Given a class of functions 
${\cal J}$ 
that defines an approximation architecture, there are two general approaches for approximating the cost function $J_\m$ of a fixed  policy $\m$ within ${\cal J}$.  The most straightforward approach, referred to as {\it direct} (or cost fitting),
is to find a $\tl J_\m\in {\cal J}$ that matches $J_\m$ in some least squares error
sense, i.e.,\footnote{\dag}{\ninepoint  Nonquadratic optimization criteria may also be used, although in practice the simple quadratic cost function has been adopted most frequently.}
$$\tl J_\m\in \arg\min_{\tl J\in {\cal J}}\|\tl J-J_\m\|^2.\xdef\costfit{\lab}\eqnum\show{oneo}$$
Typically $\|\cdot\|$ is some weighted Euclidean norm with positive weights $\xi_i$, $i=1,\ldots,n$, while ${\cal J}$ consists of a parametrized class of functions $\tl J(i,r)$ where $r=(r_1,\ldots,r_s)\in\re^s$ is the parameter vector, i.e.,\footnote{\ddag}{\ninepoint We use standard vector notation. In particular, $\re^s$ denotes the Euclidean space of $s$-dimensional real vectors, and $\re$ denotes the real line.} 
$${\cal J}=\big\{\tl J(\cdot,r)\mid r\in\re^s\big\}.$$
Then the minimization problem in Eq.\ \costfit\ is written as
$$\min_{r\in\re^s}\sum_{i=1}^n\xi_i\big(\tl J(i,r)-J_\m(i)\big)^2,\xdef\squarenormmin{\lab}\eqnum\show{oneo}$$
and can be viewed as an instance of nonlinear regression.

In simulation-based methods, the preceding minimization is usually approximated by a least squares minimization of the form
$$\min_{r\in\re^s}\sum_{m=1}^M\big(\tl J(i_m,r)-\b_m\big)^2,\xdef\leastsquaresfit{\lab}\eqnum\show{oneo}$$
where $(i_m,\b_m)$, $m=1,\ldots,M$, are a large number of state-cost sample pairs, i.e., for each $m$, $i_m$ is a sample state and $\b_m$ is equal to $J_\m(i_m)$ plus some simulation noise. Under mild statistical assumptions on the sample collection process, the sample-based minimization \leastsquaresfit\ is equivalent in the limit to the exact minimization \squarenormmin. Neural network-based approximation, as described in Section 3,  is an important example of direct approximation that uses state-cost training pairs. 

A common choice is to take ${\cal J}$ to be the subspace $\{\Phi r\mid r\in \re^s\}$ that is spanned by the columns of an $n\times s$ matrix $\Phi$, which can be viewed as basis functions 
(see the left side of Fig.\ \figsofi). Then the approximation problem \leastsquaresfit\ becomes the linear least
squares problem
$$\min_{(r_1,\ldots,r_s)\in\re^s}\sum_{m=1}^M\lf(\sum_{\ell=1}^s\phi_{i_m\ell}r_\ell-\b_m\ri)^2,\xdef\leastsquaresfit{\lab}\eqnum\show{oneo}$$
where $\phi_{i\ell}$ is the $i\ell$th entry of the matrix $\Phi$ and $r_\ell$ is the $\ell$th component of $r$. The solution of this problem can be obtained analytically and can be written in closed form (see e.g., [BeT96], Section 3.2.2). Note that the $i$th row of $\Phi$ may be viewed as a feature vector of state $i$, and $\Phi r$ may be viewed as a linear feature-based architecture.

In Section 3, we will see that neural network-based policy evaluation combines elements of both a linear and a nonlinear architecture. The nonlinearity is embodied in the features that the neural network constructs through training, but once the features are given, the neural network can be viewed as a linear feature-based architecture. 

\topinsert
\centerline{\hskip2pc\includegraphics[width=5.5in]{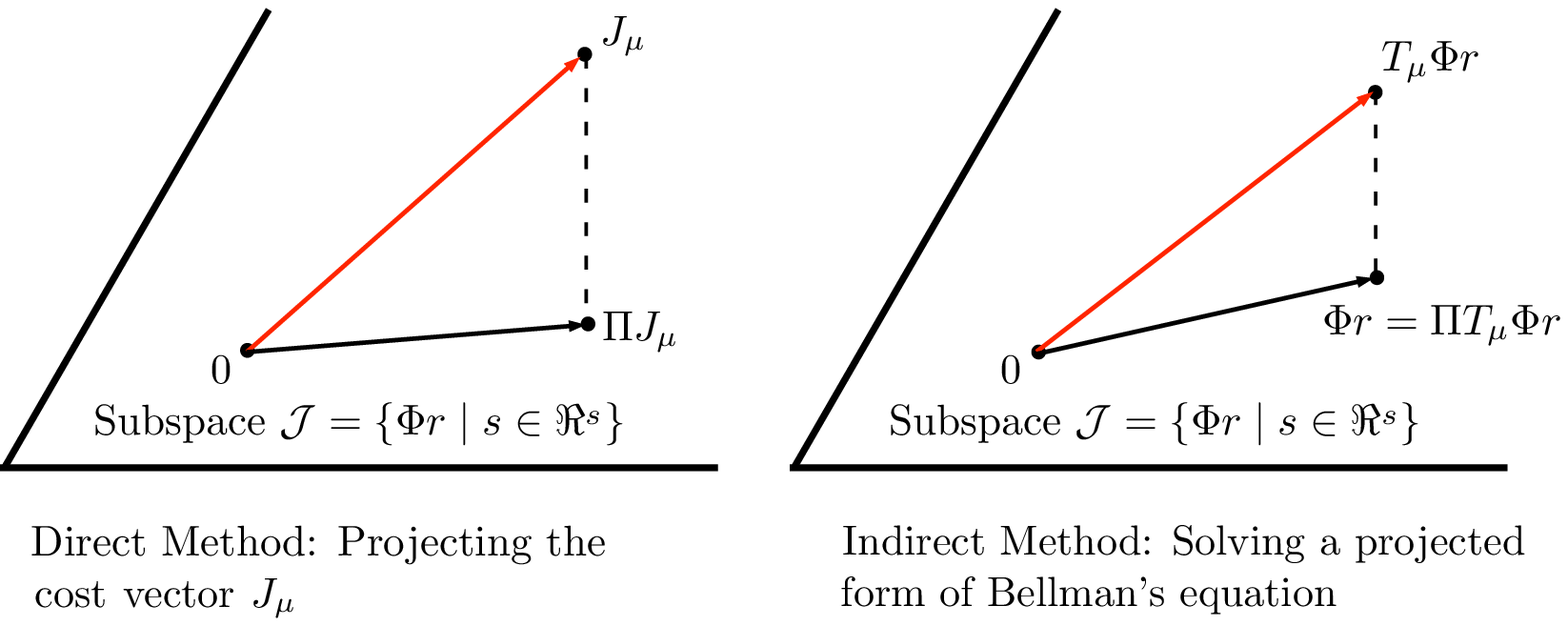}}
\vskip-1pc
\hskip-3pc\fig{0.0pc}{\figsofi} {Two methods for approximating the cost function $J_\m$ as a linear combination of basis
functions. The approximation architecture is the subspace
${\cal J}=\{\Phi r\mid r\in \re^s\}$, where $\Phi$ is matrix whose columns are the basis functions.
In the direct method (see the figure on the left), $J_\m$ is projected on ${\cal J}$. In an example of the indirect method, the approximation is obtained by solving the projected form of Bellman's equation $\Phi r=\Pi T_\m\Phi r$, where $T_\m \Phi r$ is the vector with components
$$(T_\m \Phi r)(i)=\sum_{j=1}^n
p_{ij}\bl(\mu(i)\br)\Big(g\bl(i,\mu(i),j\br) + \a (\Phi r)(j)\Big),\qquad  i=1,\ldots,n,$$
and $(\Phi r)(j)$ is the $j$th component of the vector $\Phi r$
(see the figure on the
right).}\endinsert

An often cited  weakness of simulation-based direct approximation is excessive simulation noise in the cost samples $\b_m$ that are used in the least squares minimization \leastsquaresfit\ or \leastsquaresfit. This has motivated alternative approaches for policy evaluation that inherently involve less noise. A major approach of this type, referred to as {\it indirect} (or equation fitting), is to approximate Bellman's equation for the policy $\m$,
$$J(i) = \sum_{j=1}^n
p_{ij}\bl(\mu(i)\br)\Big(g\bl(i,\mu(i),j\br) + \a J(j)\Big),\qquad  i=1,\ldots,n,\xdef\belequmu{\lab}\eqnum\show{oneo}$$
with another equation that is defined on the set ${\cal J}$.
The solution of the approximate equation is then used as an approximation of the solution of the original.  The most common indirect methods assume a linear approximation architecture, i.e., ${\cal J}$ is the subspace
${\cal J}=\{\Phi r\mid r\in \re^s\},$
and approximate Bellman's equation with another equation with fewer variables, the $s$ parameters $r_1,\ldots,r_s$. Two major examples of this approach are {\it projected equation} methods and {\it aggregation} methods, which we proceed to discuss.

\vskip-0.5pc
\subsection{Indirect Methods Based on Projected Equations}

\pn Approximation using projected equations has a long history in numerical computation (e.g., partial differential equations) where it is known as {\it Galerkin approximation} [see e.g., [KVZ72], [Fle84],  [Saa03], [Kir11]]. 
The projected equation approach is a special case of the so called Bubnov-Galerkin method, as noted in the papers  [Ber11a], [Ber11b], and [YuB10]. In the context of approximate DP it is connected with {\it temporal difference methods\/}, and it is discussed in detail in many sources (see e.g., [BeT96], [BBD10], [Ber12], [Gos15]). 

To state the projected equation, let us introduce the transformation $T_\m$, which is defined by the right-hand side of the Bellman equation \belequmu; i.e., for any $J\in\rn$, $T_\m J$ is the vector of $\rn$ with components
$$(T_\m J)(i)=\sum_{j=1}^n
p_{ij}\bl(\mu(i)\br)\Big(g\bl(i,\mu(i),j\br) + \a J(j)\Big),\qquad  i=1,\ldots,n.\xdef\tmumap{\lab}\eqnum\show{oneo}$$
Note that $T_\m$ is a linear transformation from $\rn$ to $\rn$, and in fact in compact vector-matrix notation, it is written as
$$T_\m J=g_\m+\a P_\m J,\qquad J\in\rn,\xdef\compactbelmu{\lab}\eqnum\show{oneo}$$
 where $P_\m$ is the transition probability matrix of $\m$, and $g_\m$ is the expected cost vector of $\m$, i.e., the vector with components 
$$\sum_{j=1}^n p_{ij}\bl(\mu(i)\br)g\bl(i,\mu(i),j\br),\qquad i=1,\ldots,n.$$
Moreover the Bellman equation \belequmu\ is written as the fixed point equation
$$J=T_\m J.$$

Let us denote by $\Pi J$ the projection of a vector $J\in\rn$ onto ${\cal J}$ with respect to some weighted Euclidean norm, and consider $\Pi T_\m\Phi r$, the projection of $T_\m\Phi r$ (here  $T_\m\Phi r$ is viewed as a vector in $\rn$, and $\P$ is viewed as an $n\times n$ matrix multiplying this vector). The projected equation takes the form
$$\Phi r=\Pi T_\m\Phi r;\xdef\projfix{\lab}\eqnum\show{oneo}$$
see the right-hand side of Fig.\ \figsofi. With this equation we want to find a vector $\Phi r$ of ${\cal J}$, which when transformed by $T_\m$ and then projected back onto ${\cal J}$, yields itself. This is an overdetermined system of linear equations ($n$ equations in the $s$ unknowns $r_1,\ldots,r_s$), which is equivalently written as
$$\sum_{\ell=1}^s \phi_{i\ell}r_\ell=\sum_{m=1}^n\p_{im}\sum_{j=1}^n p_{mj}\big(\m(m)\big)\left(g\big(m,\m(m),j\big)+\a \sum_{\ell=1}^s\phi_{j\ell} r_\ell\right),\qquad i=1,\ldots,n;\xdef\longprojfix{\lab}\eqnum\show{oneo}$$
here $\phi_{i\ell}$ is the $i\ell$th component of the matrix $\Phi$ and $\p_{im}$ is the ${im}$th component of the projection matrix $\Pi$.
The system can be shown to have a unique solution under conditions that can be somewhat restrictive, e.g., assuming that the Markov chain corresponding to the policy $\m$ has a unique steady-state distribution with positive components, that the projection norm involves this distribution, and that $\Phi$ has linearly independent columns (see e.g., [Ber12], Section 6.3).

An important extension is to replace the projected equation \projfix\ with the equation
$$\Phi r=\Pi T_\m^{(\l)}\Phi r,\xdef\projfixlambda{\lab}\eqnum\show{oneo}$$
where $\l$ is a scalar with $0\le \l<1$, and the transformation $T_\m^{(\l)}$ is defined by
$$\big(T_\m^{(\l)}J\big)(i)=(1-\l)\sum_{\ell=0}^\infty \l^{\ell }(T_\m^{\ell+1} J)(i),\qquad i=1,\ldots,n,\ J\in\rn,\xdef\tdlambdamap{\lab}\eqnum\show{oneo}$$
and $T_\m^\ell J$ is the $\ell $-fold composition of $T_\m$ applied to the vector $J$.  This approach to the approximate solution of Bellman's equation is supported by extensive theory and practical experience (see the textbooks noted earlier). In particular, the TD($\l$) algorithm, and other related temporal difference methods, such as LSTD($\l$) and LSPE($\l$), aim to solve by simulation the projected equation \projfixlambda. The choice of $\l$ embodies the important bias-variance tradeoff: larger values of $\l$ lead to better approximation of $J_\m$, but require a larger number of simulation samples because of increased simulation noise (see the discussion in Section 6.3.6 of [Ber12]). An important insight is that the operator $T_\m^{(\l)}$ is closely related to the proximal operator of convex analysis (with $\l$ corresponding to the penalty parameter of the proximal operator), as shown in the author's paper [Ber16a] (see also the monograph [Ber18a], Section 1.2.5, and the paper [Ber18b]). In particular, TD($\l$) can be viewed as a stochastic simulation-based version of the proximal algorithm.

A major issue in projected equation methods is whether the linear transformation $\Pi T_\m$ [or $\Pi T_\m^{(\l)}$] is a contraction mapping, in which case Eq.\ \projfix\ [or Eq.\ \projfixlambda, respectively] has a unique solution, which may be obtained by iterative fixed point algorithms. This depends on the projection norm, and it turns out that there are special
norms for which $\Pi T_\m^{(\l)}$ is a contraction (these are related to the steady-state distribution of the system's Markov chain under $\m$; see the discussion of [Ber11a] or Section
6.3 of [Ber12]). An important fact is that for any projection norm, $\Pi T_\m^{(\l)}$ is a contraction provided $\l$ is sufficiently close to 1. Still the contraction issue regarding $\P T_\m^{(\l)}$ is significant and affects materially the implementation of the corresponding approximate PI methods. 

Another important concern is that the projection matrix $\Pi$ may have some negative entries [i.e., some of the components $\p_{im}$ in Eq.\ \longprojfix\ may be negative], and as a result the linear transformations $\Pi T_\m$ and $\P T_\m^{(\l)}$ may lack the monotonicity property that is essential for the convergence of the corresponding approximate PI method. Indeed the lack of monotonicity (the possibility that we may not have $\P T_\m J\ge \P T_\m J'$ for two vectors $J,J'$ with $J\ge J'$) is the fundamental mathematical reason for policy oscillations in PI methods that are based on temporal differences (see [Ber11a], [Ber12]). We refer to the literature for further details and analysis regarding the projected equations \projfix\ and \projfixlambda, as our focus will be on aggregation methods, which we discuss next. 

\vskip-0.5pc
\subsection{Indirect Methods Based on Aggregation}

\pn Aggregation is another major indirect 
approach, which has originated in numerical linear algebra. Simple examples of aggregation involve finite-dimensional approximations of infinite dimensional equations, coarse grid approximations of linear systems of equations defined over a dense grid, and other related methods for dimensionality reduction of high-dimensional systems. In the context of DP, the aggregation idea is implemented by replacing the Bellman equation $J=T_\m J$ [cf.\ Eq.\ \belequmu] with a lower-dimensional  ``aggregate" equation, which is defined on an approximation subspace ${\cal J}=\{\Phi r\mid r\in \re^s\}$. The aggregation counterpart of the projected equation  $\Phi r=\Pi T_\m\Phi r$ is  
$$\Phi r=\Phi DT_\m\Phi r,\xdef\aggreleq{\lab}\eqnum\show{oneo}$$
 where $\Phi$ and $D$ are some matrices, and $T_\m$ is the linear transformation given by Eq.\ \tmumap.\footnote{\dag}{\ninepoint  It turns out that under some widely applicable conditions, including the assumptions of Section 4, the projected and aggregation equations are closely related. In particular, it can be proved under these conditions that the matrix $\Phi D$ that appears in the aggregation equation \aggreleq\ is a  projection with respect to a suitable weighted Euclidean seminorm (see [YuB12], Section 4, or the book [Ber12]; it is a norm projection in the case of hard aggregation). Aside from establishing the relation between the two major indirect approximation methods, projected equation and aggregation, this result provides the basis for transferring the rich methodology of temporal differences methods such as TD($\l$) to the aggregation context.}
 This is a vector-matrix notation for the linear system of $n$ equations in the $s$ variables $r_1,\ldots,r_s$
 $$\sum_{k=1}^s \phi_{ik}r_k=\sum_{k=1}^s \phi_{ik} \sum_{m=1}^n d_{k m}\sum_{j=1}^n p_{mj}\big(\m(m)\big)\left(g\big(m,\m(m),j\big)+\a \sum_{\ell=1}^s\phi_{j\ell} r_\ell\right),\qquad i=1,\ldots,n,$$
where $\phi_{i\ell}$ is the $i\ell$th component of the matrix $\Phi$ and $d_{k m}$ is the ${k m}$th component of the matrix $D$.
 
A key restriction for aggregation methods as applied to DP is that {\it the rows of $D$ and $\Phi$ should be probability distributions\/}. These distributions usually have intuitive interpretations in the context of specific aggregation schemes; see [Ber12], Section 6.5 for a discussion. 
Assuming that $\Phi$ has linearly independent columns, which is true for the most common types of aggregation schemes, Eq.\ \aggreleq\ can be seen to be equivalent to
$$r=DT_\m\Phi r,\xdef\aggreleqs{\lab}\eqnum\show{oneo}$$ 
or 
 $$r_k=\sum_{m=1}^n d_{k m}\sum_{j=1}^n p_{mj}\big(\m(m)\big)\left(g\big(m,\m(m),j\big)+\a \sum_{\ell=1}^s\phi_{j\ell} r_\ell\right),\qquad k=1,\ldots,s.\xdef\longaggreleqs{\lab}\eqnum\show{oneo}$$
In most of the important aggregation methods, including the one of Section 4, $D$ and $\Phi$ are chosen so that the product $D\Phi$ is the identity:
 $$D\Phi=I.$$
Assuming that this is true, the operator $I-DT_\m \Phi$ of the aggregation equation \aggreleqs\ is obtained by pre-multiplying and post-multiplying the operator $I-T_\m$ of the Bellman equation 
with $D$ and $\Phi$, respectively. Mathematically, this can be interpreted as follows:

\nitem{(a)} {\it Post-multiplying with $\Phi$\/}: We replace the $n$ variables $J(j)$ of the Bellman equation $J=T_\m J$ with convex combinations of the $s$ variables $r_\ell$ of the system \aggreleq,  using the rows $(\phi_{j1},\ldots,\phi_{js})$ of $\Phi$: 
$$J(j)\approx \sum_{\ell=1}^s\phi_{j\ell}\,r_\ell.$$

\nitem{(b)} {\it Pre-multiplying with $D$\/}:  We form the $s$ equations of the aggregate system by taking convex combinations of the $n$ components of the $n\times n$ Bellman equation using the rows of $D$.
\smskip

We will now describe how the aggregate system of Eq.\ \longaggreleqs\ can be associated with a discounted DP problem that  has $s$ states, called the {\it aggregate states} in what follows. At an abstract level, the aggregate states may be viewed as entities associated with the $s$ rows of $D$ or the $s$ columns of $\Phi$. 
Indeed, since $T_\m$ has the form $T_\m J=g_\m+\a P_\m J$ [cf.\ Eq.\ \compactbelmu], the aggregate system  \longaggreleqs\ becomes
\ $$r=\hat g_\m+\a \hat P_\m r,\xdef\aggreleqt{\lab}\eqnum\show{oneo}$$
where 
$$\hat g_\m=Dg_\m,\qquad \hat P_\m=DP_\m\Phi.\xdef\aggrdata{\lab}\eqnum\show{oneo}$$
 It is straightforward to verify that $\hat P_\m$ is a transition probability matrix, since the rows of $D$ and $\Phi$ are probability distributions.
This means that the aggregation equation \aggreleqt\ [or equivalently Eq.\ \longaggreleqs] represents a policy evaluation/Bellman equation for the discounted problem with transition matrix $\hat P_\m$ and cost vector $\hat g_\m$. This problem will be called the {\it aggregate DP problem} associated with policy $\m$ in what follows. The corresponding aggregate state costs are $r_1,\ldots,r_s$. 
Some important consequences of this are:
\nitem{(a)} The aggregation equation \aggreleqt-\aggrdata\ inherits the favorable characteristics of the Bellman equation $J=T_\m J$, namely its monotonicity and contraction properties, and its uniqueness of solution.
\nitem{(b)} Exact DP methods may be used to solve the aggregate DP problem. These methods often have more regular behavior than their counterparts based on  projected equations.
\nitem{(c)} Approximate DP methods, such as variants of simulation-based PI, may also be used to solve approximately the aggregate DP problem. 
\smskip
\pn The preceding characteristics of the aggregation approach may be turned to significant advantage, and may counterbalance the restriction on the structure of $D$ and $\Phi$ (their rows must be probability distributions, as stated earlier).

\old{
\subsection{Relation Between Aggregation and Projected Equations}
\pn To show this, let us consider a fixed policy $\m$ of the original problem, and the corresponding aggregate Bellman equation, which takes the form
$$\Phi r=\Phi DT_\m\Phi r,\xdef\aggrbelmu{\lab}\eqnum\show{qfortyss}$$ 
[cf.\ Eq.\ \aggreleq]. It turns out 
that this equation can also be viewed as a projected equation, 
$$\Phi r=\Pi T_\m\Phi r,\xdef\projbelmu{\lab}\eqnum\show{qfortyss}$$ 
because $\Pi=\Phi D$ is a projection  over $S$ with respect to a {\it weighted seminorm\/}. 
In particular, for a vector $\xi=\{\xi_1,\dots,\xi_n\}$ with $\xi_i\ge0$ for all $i$, let $\Xi$ be the diagonal matrix with the components $\xi_i$ along the diagonal, and let $\|\cdot\|_\xi$ be the seminorm defined by
$\|x\|_\xi=\sqrt{x'\Xi x}$. (Note that $\|\cdot\|_\xi$ need not be a norm, since some of the components $\xi_i$ may be 0.) 
We let the vector $\xi$ be defined by the probability distribution defined by $D$:
$$\xi_i={d_{\ell i}\over \sum_{k=1}^s\sum_{j=1}^nd_{kj}}={d_{\ell i}\over s},\qquad\hbox{if }i\in I_\ell,\ \ell=1,\ldots,s,\xdef\xiicoeff{\lab}\eqnum\show{qfortyss}$$
and $\xi_i=0$ for all $i\notin \cup_{\ell=1}^sI_\ell$. Then it is straightforward to show that $\Phi D$ is the projection operator defined by this seminorm and that the aggregation and projection equations \aggrbelmu\ and \projbelmu\ become identical.\footnote{\dag}{\ninepoint  The projection operator can be written explicitly. To see this note that our assumptions imply that the matrix $\Phi'\Xi\Phi$ is invertible. Next we argue that for each $x\in\rn$ there exists a unique vector, denoted $\Pi x$, that minimizes $\|x-y\|_\xi$ over $y\in S$. Moreover,
$$\Pi x=\Phi(\Phi'\Xi\Phi)^{-1}\Phi'\Xi x,\qquad\forall\ x\in\rn.$$
This follows from the necessary and sufficient optimality condition for the problem of minimizing $\|x-\Phi r\|_\xi^2$ over $r\in\re^s$:
$$\Phi'\Xi(\Phi r-x)=0.$$
Since $\Phi'\Xi \Phi$ is invertible, the solution of this linear system is unique and given by $(\Phi'\Xi \Phi)^{-1}\Phi'\Xi x$. Now by working the formula for $\Pi$ using Eq.\ \xiicoeff, we can verify that $\Pi=\Phi D$.}
}


\subsection{Implementation Issues}

\pn The implementation of approximate PI methods involves several delicate issues, which have been extensively investigated but have not been fully resolved, and are the subject of continuing research. We will discuss briefly some of these issues in what follows in this section. We preface this discussion by noting that all of these issues are addressed more easily and effectively within the direct approximation and the aggregation frameworks, than within the temporal difference/projected equation framework, because of the deficiencies relating to the lack of monotonicity and contraction of the operator $\Pi T_\m$, which we noted in Section 2.2.

\subsubsection{The Issue of Exploration}

\pn 
An important generic difficulty with simulation-based PI is that in order to evaluate a
policy
$\m$, we may need to generate cost samples using that policy, but this may bias the simulation
by underrepresenting states that are unlikely to occur under  $\m$. As a result, the
cost-to-go estimates of these underrepresented states may be highly inaccurate, causing potentially serious
errors in the calculation of the improved control policy $\hat \m$ via the policy improvement equation \thrtene. 

The situation just described is known as {\it inadequate exploration} 
of the system's dynamics. It is a particularly acute
difficulty when the system is deterministic [i.e., $p_{ij}(u)$ is equal to 1 for a single successor state $j$], or when the randomness embodied in the transition
probabilities of the current policy is ``relatively small," since then few states may be reached from a given initial state when the current policy is simulated. 

One possibility to guarantee adequate exploration of the
state space is to break down the simulation to multiple short trajectories (see [Ber11c], [Ber12], [YuB12])
and to ensure that the initial states employed
form a rich and representative subset. This is naturally done within the direct approximation and the aggregation frameworks, but less so in the temporal difference framework, where the theoretical convergence analysis relies on the generation of a single long trajectory.
 
Another possibility for exploration is to artificially introduce some extra randomization in the
simulation of the current policy,  by occasionally generating random transitions using some policy other than $\m$ (this is called an {\it off-policy approach} and its implementation has been the subject of considerable discussion; see the books [SuB98], [Ber12]). A Monte Carlo tree search implementation may naturally provide some degree of such randomization, and has worked well in game playing contexts, such as the AlphaZero architecture for playing chess, Go, and other games (Silver et al., [SHS17]).
Other related approaches to improve exploration based on generating multiple short trajectories are discussed in Sections 6.4.1 and 6.4.2 of [Ber12]. 

\vskip-0.5pc
\subsubsection{Limited Sampling/Optimistic Policy Iteration}
\vskip-0.3pc

\pn In the approximate PI approach discussed so far, the evaluation of the current policy $\m$ must be fully
carried out. An alternative is {\it optimistic PI\/}, where relatively few simulation samples are processed between successive policy changes and corresponding parameter updates. 

Optimistic PI with cost function approximation is frequently used in practical applications. In particular, extreme optimistic schemes, including nonlinear architecture versions, and involving a single or very few $Q$-factor updates between parameter updates have been widely recommended; see e.g., the books [BeT96], [SuB98], [BBD10] (where they are referred to as SARSA, a shorthand for State-Action-Reward-State-Action). The behavior of such schemes is very complex, and their theoretical convergence properties are unclear. In particular, they can exhibit  fascinating and counterintuitive behavior, including a natural tendency for policy oscillations. This tendency is common to both optimistic and nonoptimistic PI, as we will discuss shortly, but in extreme optimistic PI schemes, oscillations tend to manifest themselves in an unusual form whereby we may have convergence in parameter space and oscillation in policy space (see [BeT96], Section 6.4.2, or [Ber12], Section 6.4.3).

On the other hand optimistic PI may in some cases deal better with the problem of 
exploration discussed earlier. The reason is that with rapid changes of policy, there may be less tendency to bias the simulation towards particular states that are
favored by any single policy. 

\vskip-0.5pc
\subsubsection{Policy Oscillations and Chattering}
\vskip-0.3pc

\pn  Contrary to exact PI, which converges to an optimal policy in a fairly regular manner, approximate PI may oscillate. By this we mean that after a few iterations, policies tend to repeat in cycles. The  parameter vectors $r$ that correspond to the oscillating policies may also tend to oscillate, although it is possible, in optimistic approximate PI methods, that there is convergence in parameter space and oscillation in policy space, a peculiar phenomenon known as {\it chattering\/}.

Oscillations and chattering have been explained with the use of the so-called ``greedy partition" of the parameter space into subsets that correspond to the same improved policy (see [BeT96], Section 6.4.2, or [Ber12], Section 6.4.3).  Policy oscillations occur when the generated parameter sequence straddles the boundaries that separate sets of the partition.
 Oscillations can be potentially very damaging, because there is no guarantee that the policies involved in the oscillation are ``good" policies, and there is often no way to verify how well they compare to the optimal.

We note that oscillations are avoided and approximate PI can be shown to converge to a single policy under special conditions that arise in particular when aggregation is used for policy evaluation. These conditions involve certain monotonicity assumptions  [e.g., the nonnegativity of the components $\p_{im}$ of the projection matrix in Eq.\ \longprojfix], which are fulfilled in the case of aggregation (see [Ber11a]). However, for temporal difference methods, policy oscillations tend to occur generically, and often for very simple problems, involving few states (a two-state example is given in [Ber11a], and in [Ber12], Section 6.4.3). This is a potentially important advantage of the aggregation approach.

\vskip-0.5pc
\subsubsection{Model-Free Implementations}

\pn In many problems a mathematical model [the transition probabilities $p_{ij}(u)$ and the cost vector $g$] is unavailable or hard to construct, but instead the system and cost structure
can be simulated far more easily. In particular, let us assume that there is a computer program that for any given state $i$ and control
$u$, simulates sample  transitions to a successor state $j$ according to
$p_{ij}(u)$, and  generates the transition cost $g(i,u,j)$.  

As noted earlier, the direct and indirect approaches to approximate evaluation of a single policy may be implemented in model-free fashion, simply by generating the needed cost samples for the current policy by simulation. However, given the result $\tl J_\m(\cdot)$ of the approximate policy evaluation, the policy improvement minimization 
$$\hat \m(i)\in\arg\min_{u\in U(i)}\sum_{j=1}^n p_{ij}(u)\bl(g(i,u,j)+\a\tl J_\m(j)\br),\qquad i=1,\ldots,n,\xdef\thrtenea{\lab}\eqnum\show{oneo}$$
still requires the transition probabilities $p_{ij}(u)$, so it is not model-free. To provide a model-free version we may use a parametric regression approach. In particular, suppose that for any state $i$ and control $u$, state transitions $(i,j)$, and corresponding transition costs $g(i,u,j)$ and values of $\tl J_\m(j)$ can be generated in a model-free fashion when needed, by using a simulator of the true system. Then we can introduce a parametric family/approximation architecture of $Q$-factor functions, $\tl Q_\m(i,u,\theta)$, where $\theta$ is the parameter vector, and use a regularized least squares fit/regression to approximate the expected value that is minimized in Eq.\ \thrtenea. The steps are as follows:

\nitem{(a)} Use the simulator to collect a large number of ``representative" sample state-control pairs $(i_m,u_m)$, and successor states $j_m$, $m=1,\ldots,M$, and corresponding sample $Q$-factors
$$\b_m=g(i_m,u_m,j_m)+\a \tl J_\m(j_m),\qquad m=1,\ldots,M.\xdef\qsamples{\lab}\eqnum\show{oneo}
$$
\nitem{(b)} Determine the parameter vector $\tl \theta$ with the least-squares minimization
$$\tl \theta \in\arg\min_\theta\sum_{m=1}^M\big(\tl Q_\m(i_m,u_m,\theta)-\b_m\big)^2\xdef\qregression{\lab}\eqnum\show{oneo}$$
(or a regularized minimization whereby a quadratic regularization term is added to the above quadratic objective).
\nitem{(c)} Use the policy
$$\hat \m(i)\in\arg\min_{u\in U(i)}\tl Q_\m(i,u,\tl \theta),\qquad i=1,\ldots,n.\xdef\qpolicy{\lab}\eqnum\show{oneo}$$
This policy may be generated on-line when the control constraint set $U(i)$ contains a reasonably small number of elements. Otherwise an approximation in policy space is needed to represent the policy $\hat \m$ using a policy approximation architecture. Such an architecture could be based on a neural network, in which case it is commonly called an ``action network" or ``actor network" to distinguish from its cost function approximation counterpart, which is called a ``value network"  or ``critic network."
\smskip 

Note some important points about the preceding approximation procedure:
\nitem{(1)} It does not need the transition probabilities $p_{ij}(u)$ to  generate  the policy $\hat \m$ through the minimization \qpolicy. The simulator to collect the samples \qsamples\ suffices.

\nitem{(2)} The policy $\hat \m$ obtained through the minimization \qpolicy\ is not the same as the one obtained through the minimization \thrtenea. There are two reasons for this. 
One is the approximation error introduced by the $Q$-factor architecture $\tl Q_\m$, and the other is the simulation error introduced by the finite-sample regression \qregression.
 We  have to accept these sources of error as the price to pay for the convenience of not requiring a mathematical model.

\nitem{(3)} Two approximations are potentially required: One to compute $\tl J_\m$, which is needed for the samples $\b_m$ [cf.\ Eq.\ \qsamples], and another to compute $\tl Q_\m$ through the least squares minimization \qregression, and the subsequent policy generation formula \qpolicy. The approximation methods to obtain $\tl J_\m$ and $\tl Q_\m$ may not be the same and in fact may be unrelated (for example $\tl J_\m$ need not involve a parametric approximation, e.g., it may be obtained by some type of problem approximation approach). 

\smskip

An alternative to first computing $\tl J_\m(\cdot)$ and then computing subsequently $\tl Q_\m(\cdot,\cdot,\theta)$ via the procedure \qsamples-\qpolicy\ is to forgo the computation of $\tl J_\m(\cdot)$, and use just the parametric approximation architecture for the policy $Q$-factor, $\tl Q_\m(i,u,\theta)$. We may then train this $Q$-factor architecture, using state-control $Q$-factor samples, and either the direct or the indirect approach.  Generally, algorithms for approximating policy cost functions can be adapted to approximating policy $Q$-factor functions. 

As an example, a direct model-free approximate PI scheme can be defined by Eqs.\ \qregression-\qpolicy, using $M$ state-control samples $(i_m,u_m)$, corresponding successor states  $j_m$ generated according to the probabilities $p_{i_mj}(u_m)$, and sample costs $\b_m$ equal to the sum of:
\nitem{(a)} The first stage cost $g(i_m,u_m,j_m)$.
\nitem{(b)} A $\a$-discounted simulated sample of the infinite horizon cost of starting at $j_m$ and using $\m$ [in place of the term $\a \tl J_\m(j_m)$ in  Eq.\ \qsamples].
\smskip
\pn A PI scheme of this type was suggested by Fern, Yoon, and Givan [FYG06], and has  been discussed by several other authors; see [Ber17], Section 6.3.4. In particular, a variant of the method was used to train a tetris playing computer program that performs impressively better than programs that are based on other variants of approximate PI, and various other methods; see Scherrer [Sch13], Scherrer et al.\ [SGG15], and Gabillon, Ghavamzadeh, and Scherrer [GGS13], who also provide an analysis.

\vskip-1.5pc

\section{Approximate Policy Evaluation Based on Neural Networks}

\pn In this section we will describe some of the basic ideas of the neural network methodology as it applies to the approximation of the cost vector $J_\m$ of a fixed policy $\m$. Since $\m$ is fixed throughout this section, we drop the subscript $\m$ is what follows. A neural network provides an  architecture  of the form
$$\tl J(i,v,r)=\sum_{\ell=1}^sF_{\ell}(i,v)r_\ell\xdef\nnform{\lab}\eqnum\show{oneo}$$
that depends on a parameter vector $v$ and a parameter vector $r=(r_1,\ldots,r_s)$. Here for each state $i$, $\tl J(i,v,r)$ approximates $J_\m(i)$, while the vector 
$$F(i,v)=\big(F_{1}(i,v),\ldots,F_{s}(i,v)\big)$$
may be viewed  as a feature vector of the state $i$.
Notice the different roles of the two parameter vectors: $v$ parametrizes  $F(i,v)$, and $r$ is a vector of weights that combine linearly the components of $F(i,v)$.
The idea is to use training to obtain simultaneously both the features and the linear weights. 

Consistent with the direct approximation framework of Section 2.1, to train a neural network, we generate a training set that consists of a large number of state-cost pairs $(i_m,\b_m)$, $m=1,\ldots,M$, and we find $(v,r)$ that minimizes
 $$\sum_{m=1}^M\big(\tl J(i_m,v,r)-\b_m\big)^2.\xdef\leastsquares{\lab}\eqnum\show{oneo}$$
The training pairs $(i_m,\b_m)$ are generated by some kind of calculation or simulation, and they may contain noise, i.e., $\b_m$ is the cost of the policy starting from state $i_m$ plus some error.\footnote{\dag}{\ninepoint There are also neural network implementations of the indirect/projected equation approximation approach, which make use of temporal differences, such as for example nonlinear versions of TD($\l$). We refer to the textbook literature on the subject, e.g., [SuB98]. In this paper, we will focus on neural network training that is based on minimization of the quadratic cost function \leastsquares.}

\xdef\fignnsinglelayer{\figr}\figrnum\show{myfigure}
\xdef\fignnmultilayer{\figr}\figrnum\show{myfigure}

The simplest type of neural network is the {\it single layer perceptron\/}; see  Fig.\ \fignnsinglelayer. Here the state
$i$ is encoded as a vector of numerical values 
$y(i)$ with components $y_1(i),\ldots,y_k(i)$, which is then transformed
linearly as 
$$Ay(i)+b,$$
where $A$ is an $m\times k$ matrix and $b$ is a vector in $\re^m$. Some of the components of $y(i)$ may be known interesting features of $i$ that can be designed based on  problem-specific knowledge or prior training experience.   This transformation will be referred to as the {\it linear layer} of the neural network. We view the components of $A$ and $b$ as parameters to be determined, and we group them together into the parameter vector $v=(A,b)$.

 Each of the $s$ scalar output components of the linear layer, 
$$\big(Ay(i)+b\big)_\ell,\qquad \ell=1,\ldots,s,$$
becomes the input to a nonlinear differentiable function $\sigma$ that maps scalars to scalars.  Typically $\sigma$ is monotonically increasing. A simple and popular possibility is the {\it rectified linear unit\/}, which is simply the function $\max\{0,\xi\}$, ``rectified" to a differentiable function by some form of smoothing operation; for example 
$$\s(\xi)=\ln(1+e^\xi).$$
 Other functions, used since the early days of neural networks, have the property
$$-\infty<\lim_{\xi\to-\infty}\s(\xi)<\lim_{\xi\to \infty}\s(\xi)<\infty.$$ 
Such functions are referred to as {\it sigmoids\/},  and some
common choices are the {\it hyperbolic tangent} function
$$\sigma(\xi)=\tanh(\xi)={e^\xi-e^{-\xi}\over e^\xi+e^{-\xi}},$$
and the {\it logistic} function
$$\sigma(\xi)={1\over 1+e^{-\xi}}.$$
In what follows, we will ignore the character of the function $\s$ (except for the differentiability requirement), and simply refer to it as a ``nonlinear unit" and to the corresponding layer as a ``nonlinear layer."

\topinsert
\centerline{\hskip0pc\includegraphics[width=6in]{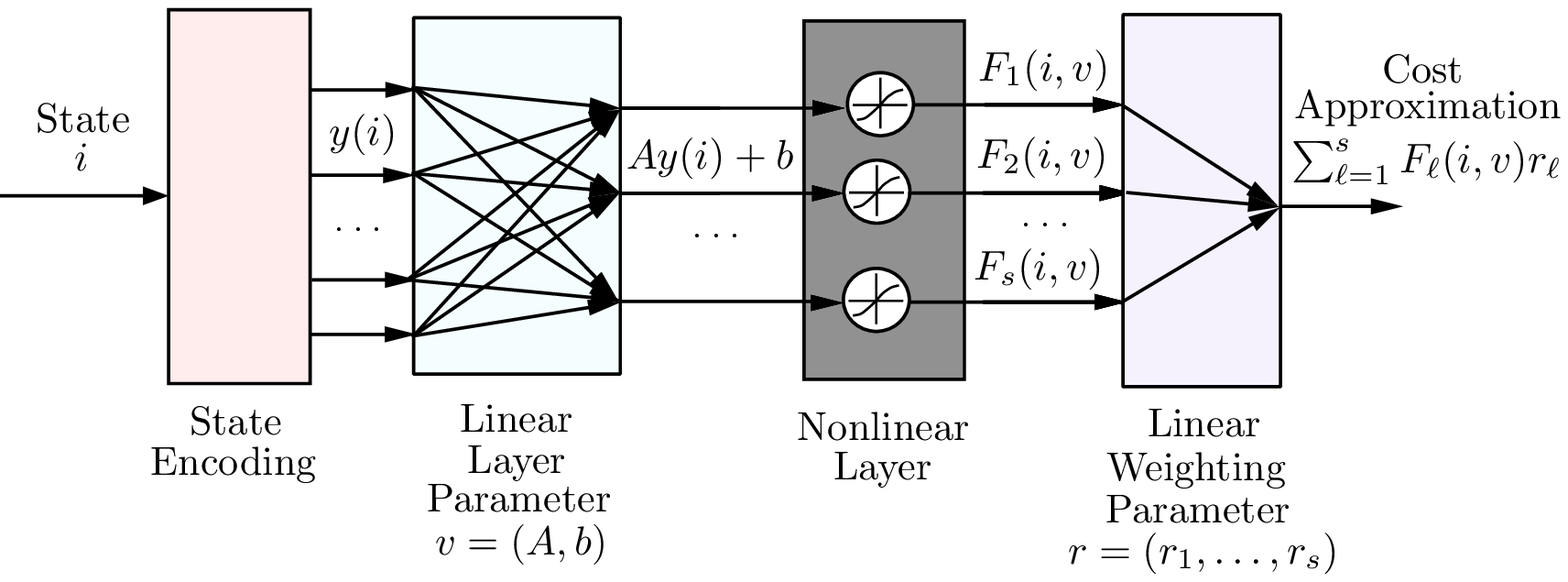}}
\vskip-1pc
\hskip-3pc\fig{0pc}{\fignnsinglelayer}{A perceptron consisting of a linear layer and a nonlinear layer. It provides a way to compute features of the state, which can be used for approximation of the cost function of a given policy. The state
$i$ is encoded as a vector of numerical values 
$y(i)$, which is then transformed
linearly as 
$Ay(i)+b$ in the linear layer. The scalar output components of the linear layer, 
become the inputs to single input-single output nonlinear functions that produce the $s$ scalars 
$F_{\ell}(i,v)=\sigma\big((Ay(i)+b)_\ell\big),$
 which can be viewed as feature components that are in turn linearly weighted with parameters $r_\ell$.
}
\endinsert
 
At the outputs of the nonlinear units, we obtain the scalars
$$F_{\ell}(i,v)=\sigma\big((Ay(i)+b)_\ell\big),\qquad \ell=1,\ldots,s.$$
One possible interpretation is to view these scalars as features of state $i$, which are linearly combined using weights $r_\ell$, $\ell=1,\ldots,s$,  to produce the final output
$$\sum_{\ell=1}^s
F_{\ell}(i,v)r_\ell=\sum_{\ell=1}^s
\sigma\Big(\big(Ay(i)+b\big)_\ell\Big)\,r_\ell.\xdef\nnap{\lab}\eqnum\show{nnap}$$
Note that each value $F_{\ell}(i,v)$ depends on just the $\ell$th row of $A$ and the $\ell$th component of $b$, not on the entire vector $v$. In some cases this motivates placing some constraints on individual components of $A$ and $b$ to achieve special problem-dependent ``handcrafted" effects. 

Given a set of state-cost training pairs $(i_m,\b_m)$, $m=1,\ldots,M$, the parameters of the neural network $A$, $b$, and $r$ are obtained by solving the training problem \leastsquares, i.e., 
$$\min_{A,b,r}\sum_{m=1}^M\lf(\sum_{\ell=1}^s
\sigma\Big(\big(Ay(i_m)+b\big)_\ell\Big)\,r_\ell-\b_m\ri)^2.\xdef\trainingpr{\lab}\eqnum\show{nnap}$$
The cost function of this problem is generally nonconvex, so there may exist multiple local minima.

It is common to augment the cost function of this problem with a {\it regularization} function, such as a quadratic in the parameters $A$, $b$, and $r$. This is customary in least squares problems in order to make the problem easier to solve algorithmically. However, in the context of neural network training, regularization is primarily important for a different reason: it helps to avoid {\it overfitting\/}, which refers to a situation where a neural network model matches the training data very well but 
does not do as well on new data. This is a well known difficulty in machine learning, which may occur when the number of parameters of the neural network is relatively large (roughly comparable to the size of the training set). 
We refer to machine learning and neural network textbooks for a discussion of algorithmic questions regarding  regularization and other issues that relate to the practical implementation of the training process. In any case, the training problem \trainingpr\ is an unconstrained nonconvex differentiable optimization problem that can in principle be addressed with standard gradient-type methods. 

Let us now discuss briefly two issues regarding the neural network formulation and training process just described:

\nitem{(a)} A major question is how to solve the training problem \trainingpr. The salient characteristic of the cost function of this problem is its form as the sum of a potentially very large number $M$ of component functions. This structure can be exploited with a variant of the gradient method, called {\it incremental\/},\footnote{\dag}{\ninepoint Sometimes the more recent name ``stochastic gradient descent" is used in reference to this method. However, once the training set has been generated, possibly by some deterministic process, the method need not have a stochastic character, and it also does not guarantee cost function descent at each iteration.} which computes just the gradient of a {\it single} squared error component 
$$\lf(\sum_{\ell=1}^s
\sigma\Big(\big(Ay(i_m)+b\big)_\ell\Big)\,r_\ell-\b_m\ri)^2$$
 of the sum in Eq.\ \trainingpr\ at each iteration, and then changes the current iterate in the opposite direction of this gradient using some stepsize; the books [Ber15], [Ber16b] provide extensive accounts, and theoretical analyses including the connection with stochastic gradient methods are given in the  book [BeT96] and the paper [BeT00].  
Experience has shown that the incremental gradient method can be vastly superior to the ordinary (nonincremental) gradient method in the context of neural network training, and in fact the methods most commonly used in practice are incremental.

\nitem{(b)} Another important question is how well  we can approximate the cost function of the policy with a neural network architecture, assuming we can choose the number of the nonlinear units $s$ to be as large as we want. The answer to this question is quite favorable and is provided by the so-called {\it universal approximation theorem\/}.  Roughly, the theorem says that assuming that $i$ is an element of a Euclidean space $X$ and $y(i)\equiv i$, a neural network of the form described can approximate arbitrarily closely (in an appropriate mathematical sense), over a closed and bounded subset $S\subset X$, any piecewise continuous function $J:S\mapsto \re$, provided the number $s$ of nonlinear units is sufficiently large. For proofs of the theorem at different levels of generality, we refer to Cybenko [Cyb89], Funahashi [Fun89], Hornik, Stinchcombe, and White
[HSW89], and Leshno et al.\ [LLP93]. For intuitive explanations we refer to Bishop ([Bis95],
pp.\ 129-130) and Jones [Jon90].
\smskip

While the universal approximation theorem provides some assurance about the adequacy of the neural network structure, it does not predict the number of nonlinear units that we may need for ``good" performance in a given problem. Unfortunately, this is a difficult question to even pose precisely, let alone to answer adequately. In practice, one is reduced to trying increasingly larger numbers of units until one is convinced that satisfactory performance has been obtained for the task at hand. Experience has shown that in many cases the number of required nonlinear units and corresponding dimension of $A$ can be very large, adding significantly to the difficulty of solving the training problem. This has motivated various suggestions for modifications of the neural network structure. One possibility is to concatenate multiple single layer perceptrons so that the output of the nonlinear layer of one perceptron becomes the input to the linear layer of the next, as we will now discuss.

\vskip-0.5pc
\subsubsection{Multilayer and Deep Neural Networks} 

\pn An important generalization of the single  layer perceptron architecture is deep neural networks, which
involve multiple layers of linear and nonlinear functions. The number of layers can be quite large, hence the ``deep" characterization. The outputs of each nonlinear layer become the inputs of the next linear layer; see 
Fig.\ \fignnmultilayer. In some cases it may make sense to add as additional inputs some of the components of the state $i$ or the state encoding $y(i)$. 

The training problem for multilayer networks  has the form
$$\min_{v,r}\,\sum_{m=1}^M\lf(\sum_{\ell=1}^s
F_{\ell}(i,v)r_\ell-\b_m\ri)^2,$$
where $v$ represents the collection of all the parameters of the linear layers, and $F_{\ell}(i,v)$ is the $\ell$th feature component produced at the output of the final nonlinear layer. Various types of incremental gradient methods can also be applied here, specially adapted to the multi-layer structure and they are the methods most commonly used in practice, in combination with techniques for finding good starting points, etc. An important fact is that the gradient with respect to $v$ of each feature component $F_{\ell}(i,v)$ can be efficiently calculated using a special procedure known as {\it
backpropagation\/}, which is just a computationally efficient way to
apply the chain rule of differentiation. We refer to the specialized literature for various accounts (see e.g., [Bis95], [BeT96], [HOT06], [Hay08], [LZX17]).
  
\topinsert
\centerline{\includegraphics[width=6in]{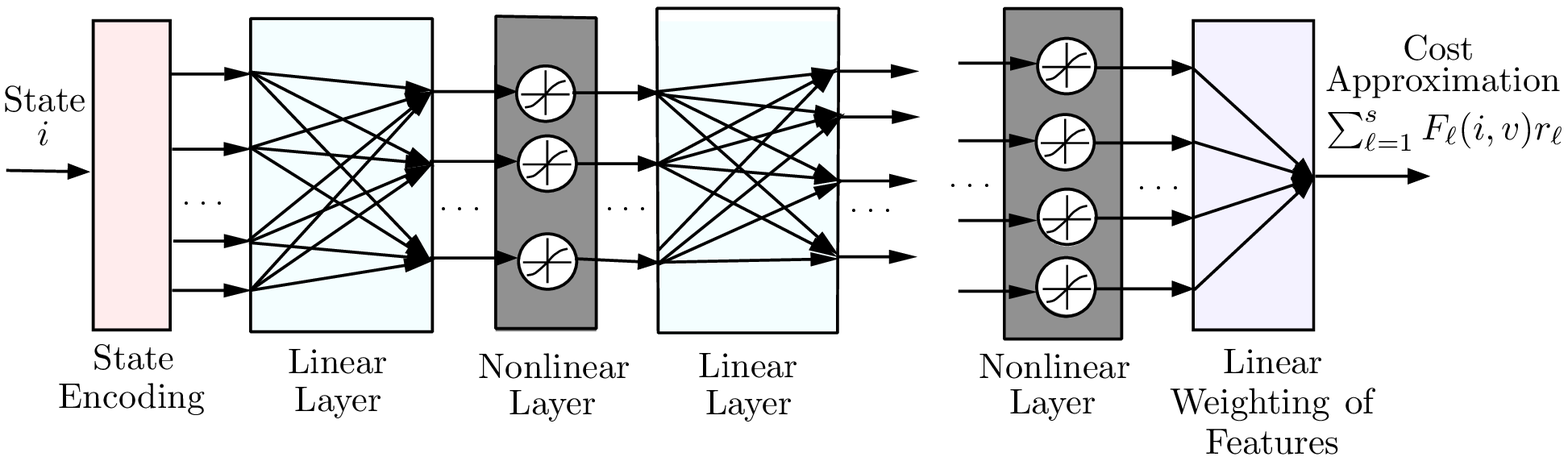}}
\vskip-1pc
\hskip-3pc\fig{0pt}{\fignnmultilayer}{A 
neural network with multiple  layers. Each nonlinear layer constructs a set of features as inputs of the next linear layer. The features are obtained at the output of the final nonlinear layer are linearly combined to yield a cost function approximation.}
\endinsert

In view of the universal approximation property, the reason for having multiple nonlinear layers is not immediately apparent. A commonly given explanation is that a multilayer network provides a hierarchical sequence of features, where each set of features in the sequence is a function of the preceding set of features in the sequence. In the context of specific applications, this hierarchical structure can be exploited in order to specialize the role of some of the layers and to enhance particular characteristics of the state. 
Another reason commonly given is that with multiple linear layers, one may consider the possibility of using matrices $A$ with a particular sparsity pattern, or other structure that embodies special linear operations such as convolution. When such structures are used, the  training problem often becomes easier, because the number of parameters in the linear layers may be drastically decreased.

Deep neural networks also have another advantage, which is important for our aggregation-related purposes in this paper: {\it the final features obtained as output of the last nonlinear layer tend to be more complex, so their number can be made smaller as the number of nonlinear  layers increases\/}. This tends to facilitate the implementation of the feature-based aggregation schemes that we will discuss in what follows.

\vskip-1.5pc

\section{Feature-Based Aggregation Framework}
\vskip-0.5pc
\pn In this section, we will specialize the general aggregation framework of Section 2.3 by introducing features in the definition of the matrices $D$ and $\Phi$. The starting point is a given {\it feature mapping\/}, i.e., a function $F$ that maps a state $i$ into its feature vector $F(i)$. We assume that $F$ is constructed in some way (including hand-crafted, or neural network-based), but we leave its construction unspecified for the moment. 

\xdef\figpiececonst{\figr}\figrnum\show{myfigure}

We will form a lower-dimensional DP approximation of the original problem, and to this end we introduce disjoint subsets $S_1,\ldots,S_q$ of state-feature pairs $\big(i,F(i)\big)$, which we call {\it aggregate states\/}. The subset of original system states $I_\ell$ that corresponds to  $S_\ell$,
$$I_\ell=\big\{i\mid (i,F(i))\in S_\ell\big\},\qquad \ell=1,\ldots,q,\xdef\disaggrset{\lab}\eqnum\show{nnap}$$
is called the {\it disaggregation set} of $S_\ell$. An alternative and equivalent definition, given $F$, is to start with disjoint subsets of states $I_\ell$, $\ell=1,\ldots,q$, and define the aggregates states $S_\ell$ by
$$S_\ell=\big\{(i,F(i))\mid i\in I_\ell\big\},\qquad \ell=1,\ldots,q.\xdef\disaggrset{\lab}\eqnum\show{nnap}$$
Mathematically, the aggregate states are the restrictions of the feature mapping on the disaggregation sets $I_\ell$. In simple terms, we may view the aggregate states  $S_\ell$ as some ``pieces" of the graph of the feature mapping $F$; see Fig.\ \figpiececonst.

To preview our framework, we will aim to construct an aggregate DP problem whose states will be the aggregate states $S_1,\ldots,S_q$, and whose optimal costs, denoted $r^*_1,\ldots,r^*_q$, will be used to construct a function approximation $\tl J$ to the optimal cost function $\jstar$. This approximation will be constant over each disaggregation set; see Fig.\ \figpiececonst.\old{\footnote{\dag}{\ninepoint Strictly speaking, in the general version of our scheme, the approximate cost of a state $i$ will depend on $F$ nonlinearly, but may depend on values $F(j)$ of some other states $j\ne i$. However, dependence on just $F(i)$ holds in the most commonly used feature-based aggregation schemes, including the hard aggregation scheme to be introduced later in this section, while in other schemes it can be made to hold after a redefinition of $F$.}}  Our ultimate objective is that $\tl J$ approximates closely $\jstar$, which suggests as a general guideline that {\it the aggregate states should be selected so that $\jstar$ is  nearly constant over each of the disaggregation sets $I_1,\ldots,I_q$\/}. This will also be brought out by our subsequent analysis. 

\topinsert
\centerline{\includegraphics[width=5.3in]{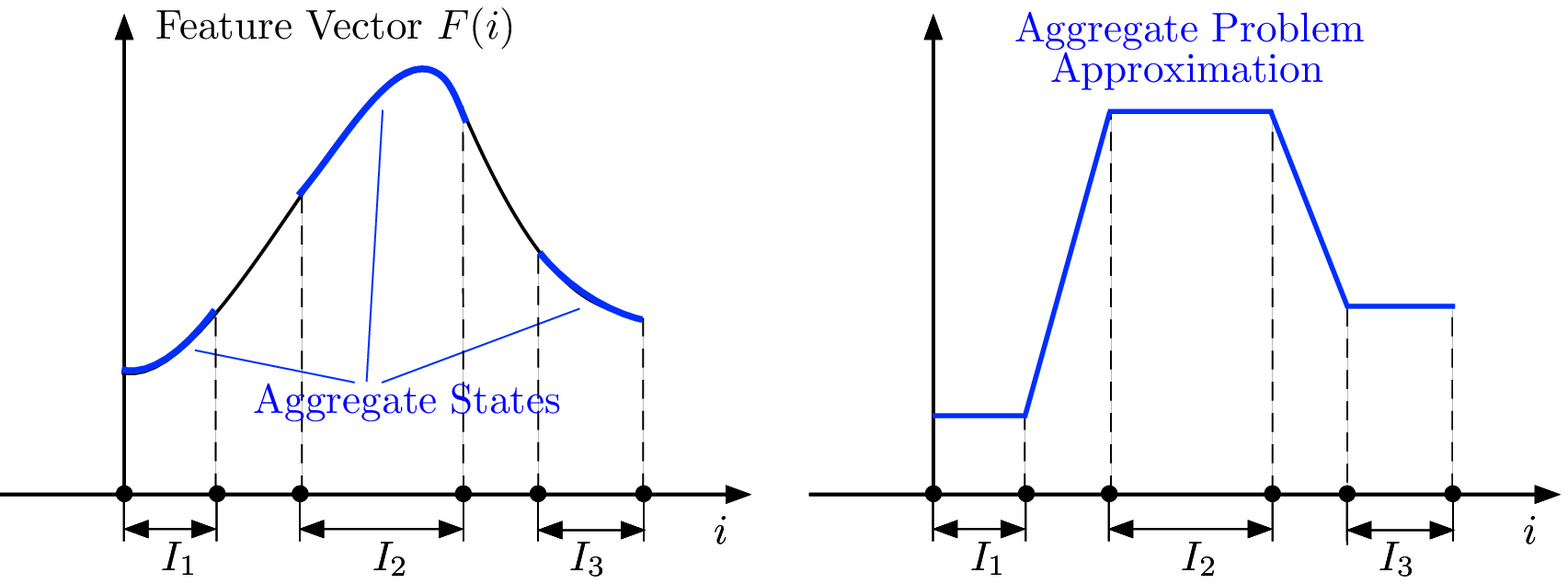}}
\vskip-0.5pc
\hskip-3pc\fig{0pt}{\figpiececonst}{Illustration of aggregate states and a corresponding cost approximation, which is constant over each disaggregation set. Here there are three aggregate states, with disaggregation sets denoted $I_1,I_2,I_3$.}
\endinsert

To formulate an aggregation model that falls within the framework of Section 2.3, we need to specify the matrices $\Phi$ and $D$. We refer to the row of $D$ that corresponds to aggregate state $S_\ell$ as the {\it disaggregation distribution of $S_\ell$} and to its elements $d_{\ell 1},\ldots,d_{\ell n}$ as the {\it disaggregation probabilities of $S_\ell$\/}. Similarly, we refer to the row of $\Phi$ that corresponds to state $j$,  $\{\phi_{j\ell}\mid \ell=1,\ldots,q\}$, as the {\it aggregation distribution of $j$\/}, and to its elements as the {\it aggregation probabilities of $j$\/}. We impose some restrictions on the components of $D$ and $\Phi$, which we describe next.

\texshopbox{
\pn{\bf Definition of a Feature-Based Aggregation Architecture}:
\pn Given the collection of aggregate states $S_1,\ldots,S_q$ and the corresponding disaggregation sets $I_1,\ldots,I_q$, the aggregation and disaggregation probabilities  satisfy the following:
\nitem{(a)} The disaggregation probabilities map each aggregate state onto its disaggregation set. By this we mean that the row of the matrix $D$ that corresponds to an aggregate state $S_\ell$ is a probability distribution $(d_{\ell 1},\ldots,d_{\ell n})$ over the original system states that assigns zero probabilities  to states that are outside the disaggregation set $I_\ell$:
$$d_{\ell i}=0,\qquad \forall\ i\notin I_\ell,\quad \ell=1,\ldots,q.\xdef\disaggrconstr{\lab}\eqnum\show{qfortyss}$$
(For example, in the absence of special problem-specific considerations, a reasonable and convenient choice would be to assign equal probability to all states in $I_\ell$, and zero probability to all other states.)
\nitem{(b)} The aggregation probabilities map each original system state that belongs to a disaggregation set onto the aggregate state of that set. By this we mean that the row  $\{\phi_{j\ell}\mid \ell=1,\ldots,q\}$ of the matrix $\Phi$ that corresponds to an original system state $j$  is specified as follows:
\nitemitem{(i)} If  $j$ belongs to some disaggregation set, say $I_\ell$, then 
$$\phi_{j\ell}=1,\xdef\aggrconstr{\lab}\eqnum\show{qfortyss}$$
\itemitem{} and $\phi_{j\ell{'}}=0$ for all $\ell{'}\ne \ell$.
\nitemitem{(ii)} If  $j$ does not belong to any disaggregation set, the row $\{\phi_{j\ell}\mid \ell=1,\ldots,q\}$ is an arbitrary probability distribution.} 

There are several  possible methods to choose the aggregate states. Generally, as noted earlier, the idea will be to form disaggregation sets over which the cost function values [$\jstar(i)$ or $J_\m(i)$, depending on the situation] vary as little as possible. We list three general approaches below, and we illustrate these approaches later with examples:

\nitem{(a)}  {\it State and feature-based approach\/}: Sample in some way the set of original system states $i$, compute the corresponding feature vectors  $F(i)$, and divide the pairs $\big(i,F(i)\big)$ thus obtained into subsets $S_1,\ldots,S_q$. Some problem-specific knowledge may be used to organize the state sampling, with proper consideration given to issues of sufficient exploration and adequate representation of what is viewed as important parts of the state space. This scheme is suitable for problems where states with similar feature vectors have similar cost function values, and is ordinarily the type of scheme that we would use in conjunction with neural network-constructed features (see Section 5). 

\nitem{(b)} {\it Feature-based approach\/}: Start  with a collection of disjoint subsets $F_\ell$, $\ell=1,\ldots,q$, of the set of all possible feature values
$${\cal F}=\big\{F(i)\mid i=1,\ldots,n\big\},$$
compute in some way disjoint state subsets $I_1,\ldots,I_q$ such that 
$$F(i)\in F_\ell,\qquad\forall\ i\in I_\ell,\ \ell=1,\ldots,q,$$
and obtain the aggregate states 
$$S_\ell=\big\{(i,F(i))\mid i\in I_\ell\big\},\qquad \ell=1,\ldots,q,$$
with corresponding disaggregation sets $I_1,\ldots,I_q$.
This scheme is appropriate for problems where it can be implemented so that each disaggregation set $I_\ell$ consists of states with similar cost function values; examples will be given in Section 4.3. 
  
\nitem{(c)} {\it State-based approach\/}: Start  with a collection of disjoint subsets of states $I_1,\ldots,I_q$, and introduce an artificial feature vector $F(i)$ that is equal to the index $\ell$ for the states $i\in I_\ell$, $\ell=1,\ldots,q$, and to some default index, say 0, for the states that do not belong to $\cup_{\ell=1}^q I_\ell$. Then use as aggregate states the subsets
$$S_\ell=\big\{(i,\ell)\mid i\in I_\ell\big\},\qquad \ell=1,\ldots,q,$$
with $I_1,\ldots,I_q$ as the corresponding disaggregation sets. In this scheme, the feature vector plays a subsidiary role, but the idea of using disaggregation subsets with similar cost function values is still central, as we will discuss shortly. (The scheme where the aggregate states are identified with subsets  $I_1,\ldots,I_q$ of original system states has been called ``aggregation with representative features" in [Ber12], Section 6.5, where its connection with feature-based aggregation has been discussed.)

\smskip

The approaches of forming aggregate states just described cover most of the aggregation schemes that have been used in practice.  Two classical examples of the state-based approach are the following:

\vskip-1pc
\texshopbox{\pn {\bf Hard Aggregation\/}: 
\pn The starting point here is a partition of the state space that consists of disjoint subsets $I_1,\ldots,I_q$  of states with $I_1\cup\cdots\cup I_q=\{1,\ldots,n\}$. The  feature vector $F(i)$ of a state $i$ identifies the set of the partition that $i$ belongs to:
$$F(i)=\ell,\qquad \forall\ i\in I_\ell,\ \ell=1,\ldots,q.\xdef\hardaggrF{\lab}\eqnum\show{qfortyss}$$
The aggregate states are the subsets 
$$S_\ell=\big\{(i,\ell)\mid i\in I_\ell\big\},\qquad \ell=1,\ldots,q,$$
and their disaggregation sets are the subsets $I_1,\ldots,I_q$. The disaggregation probabilities $d_{i\ell}$ are positive only for states $i\in I_\ell$ [cf.\ Eq.\ \disaggrconstr]. The aggregation probabilities are equal to either 0 or 1, according to
$$\phi_{j\ell}=\cases{1&if $j\in I_\ell$,\cr
0&otherwise,\cr}\qquad j=1,\ldots,n,\ \ell=1,\ldots,q,\xdef\hardaggrphi{\lab}\eqnum\show{qfortyss}$$
[cf.\ Eq.\ \aggrconstr].}
\smskip

The following aggregation example is typical of a variety of schemes arising in discretization or coarse grid schemes, where a smaller problem is obtained by discarding some of the original system states. The essence of this  scheme is to solve a reduced DP problem, obtained by approximating the discarded state costs by interpolation using the nondiscarded state costs.

\texshopboxnb{\pn {\bf  Aggregation with Representative States\/}: 
\pn The starting point here is a collection of states $i_1,\ldots,i_q$ that we view as ``representative."  The costs of the nonrepresentative states are approximated by interpolation of the costs of the representative states, using the aggregation probabilities.
The  feature mapping is
$$F(i)=\cases{\ell&if $i=i_\ell,\ \ell=1,\ldots,q$,\cr
0&otherwise.\cr}\xdef\reffdef{\lab}\eqnum\show{qfortyss}$$
The aggregate states are $S_\ell=\big\{(i_\ell,\ell)\big\}$, $\ell=1,\ldots,q$, the disaggregation sets are $I_\ell=\{i_\ell\}$, $\ell=1,\ldots,q$, and the disaggregation probabilities are equal to either 0 or 1, according to
$$d_{\ell i}=\cases{1&if  $i=i_\ell$,\cr
0&otherwise,\cr}\qquad i=1,\ldots,n,\ \ell=1,\ldots,q,$$}\texshopboxnt{\pn   
[cf.\ Eq.\ \disaggrconstr]. The aggregation probabilities must satisfy the constraint $\phi_{j\ell}=1$ if $j=i_\ell$, $\ell=1,\ldots,q$ [cf.\ Eq.\ \aggrconstr], and can be arbitrary for states $j\notin\{i_1,\ldots,i_q\}$.} 
\smskip

An important class of aggregation frameworks with representative states  arises in partially observed Markovian decision problems (POMDP), where observations from a controlled  Markov chain become available sequentially over time. Here the states of the original high-dimensional DP problem are either information vectors (groups of past measurements) or ``belief states" (conditional probability distributions of the state of the Markov chain given the available information). Features may be state estimates (given the information) and possibly their variances, or a relatively small number of representative belief states (see e.g., Section 5.1 of [Ber12] or the paper by Yu and Bertsekas [YuB04] and the references quoted there).

\subsubsection{Choice of Disaggregation Probabilities}

\pn In both of the preceding aggregation schemes, the requirement $d_{\ell i}=0$ for all $i\notin I_\ell$, cf.\ Eq.\ \disaggrconstr, leaves a lot of room for choice  of the disaggregation probabilities. Simple examples show that the values of these probabilities can affect significantly the quality of aggregation-based approximations; the paper by Van Roy [Van06] provides a relevant discussion. Thus, finding a good set of disaggregation probabilities is an interesting issue. 

Generally, problem-specific knowledge and intuition can be helpful in designing aggregation schemes, but more systematic methods may be desirable, based on some kind of gradient or random search optimization. In particular, for a given set of aggregate states and matrix $\Phi$, we may introduce a parameter vector $\theta$ and a parametrized disaggregation matrix $D(\theta)$, which is differentiable with respect to $\theta$. Then for a given policy $\m$, we may try to find $\theta$ that minimizes some cost function $F\big(\Phi r(\theta)\big)$, where $r(\theta)$ is defined as the unique solution of the corresponding aggregation equation $r=D(\theta)T_\m\Phi r$. For example we may use as cost function $F$ the squared Bellman equation residual
$$F\big(\Phi r(\theta)\big)=\big\| \Phi r(\theta)-\Phi D(\theta)T_\m\Phi r(\theta)\big\|^2.$$
The key point here is that we can calculate the gradient of $r(\theta)$ with respect to each component of $\theta$ by using simulation and low-dimensional calculations based on aggregation. We can then use the chain rule to compute the gradient of $F$ with respect to $\theta$ for use in some gradient-based optimization method. This methodology has been developed for a related projected equation context by Menache, Mannor, and Shimkin [MMS06], Yu and Bertsekas YuB09], and Di Castro and Mannor [DiM10], but has never been tried in the context of aggregation. The paper [MMS06] also suggests the use of random search algorithms, such as the cross entropy method, in the context of basis function optimization. A further discussion of parametric optimization of the disaggregation probabilities or other structural elements of the aggregation framework is beyond the scope of the present paper, but may be an interesting subject for investigation.

\xdef\figsfo{\figr}\figrnum\show{myfigure}
\xdef\figsft{\figr}\figrnum\show{myfigure}

\subsection{The Aggregate Problem}

\pn Given a feature-based aggregation framework (i.e., the aggregate states $S_1,\ldots,S_q$, the corresponding disaggregation sets $I_1,\ldots,I_q$, and the aggregation and disaggregation distributions), we can consider an aggregate DP problem that involves transitions between aggregate states. In particular, the transition probabilities $p_{ij}(u)$, and the disaggregation and aggregation probabilities specify a controlled dynamic system involving both the original system states and the aggregate states (cf.\ Fig.\ \figsfo).\footnote{\dag}{\ninepoint We will consider the aggregate problem for the case where there are multiple possible controls at each state. However, it is also possible to consider the aggregate problem for the purpose of finding an approximation to the cost function $J_\m$ of a given policy $\m$; this is the special case where the control constraint set $U(i)$ consists of the single control $\m(i)$ for every state $i$.}

\topinsert
\centerline{\includegraphics[width=5.2in]{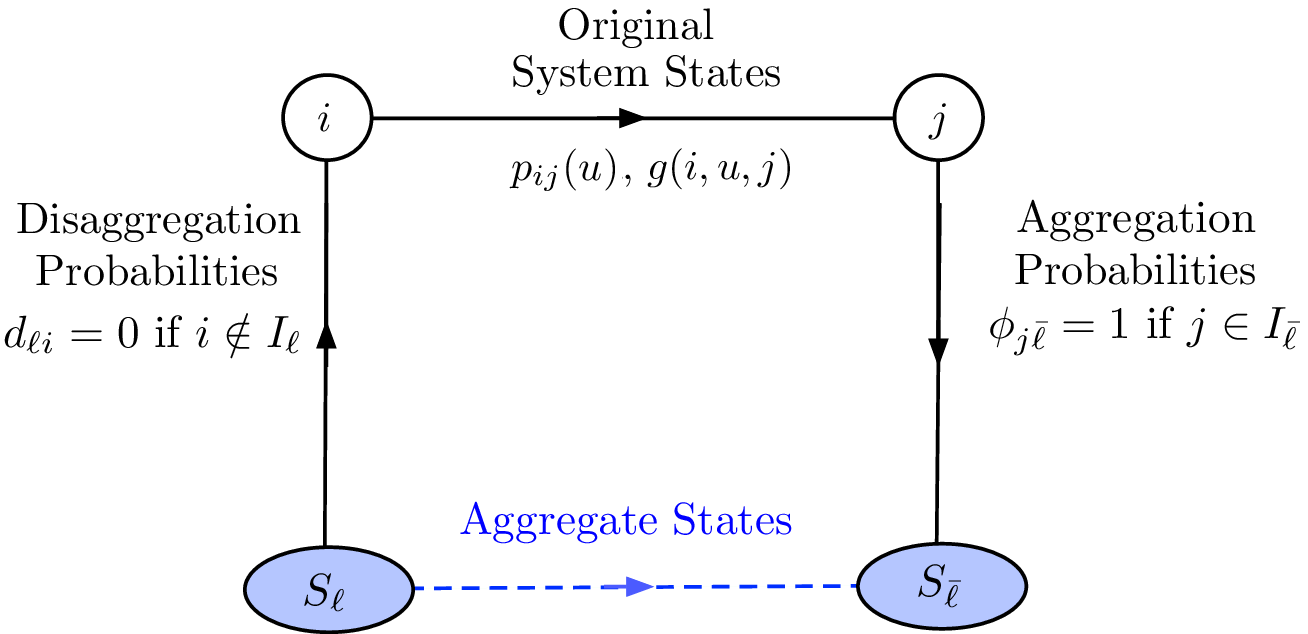}}
\vskip-0.0pc
\hskip-3pc\fig{0pc}{\figsfo.} {Illustration of the transition mechanism and the costs per stage of the aggregate problem.}\endinsert

\nitem{(i)} From aggregate state $S_\ell$, we generate a transition to original system state $i$ according to  $d_{\ell i}$ (note that $i$ must belong to the disaggregation set $I_\ell$, because of the requirement that  $d_{\ell i}>0$ only if $i\in I_\ell$).

\nitem{(ii)} From original system state $i$, we generate a transition to original system state $j$ according to $p_{ij}(u)$, with cost $g(i,u,j)$.

\nitem{(iii)}  From original system state $j$, we generate a transition to aggregate state $S_\ell$ according to  $\phi_{j\ell}$ [note here the requirement that $\phi_{j\ell}=1$ if $j\in I_\ell$; cf.\ Eq.\ \aggrconstr].
\smskip

This is a DP problem with an enlarged state space that consists of two copies of the original state space $\{1,\ldots,n\}$ plus the $q$ aggregate states. We introduce the corresponding optimal vectors $\tl J_0$, $\tl J_1$, and $r^*=\{r^*_1,\ldots,r^*_q\}$ where:

\nitem{} $r^*_\ell$ is the optimal cost-to-go from aggregate state $S_\ell$.
\nitem{} $\tl J_0(i)$ is the optimal cost-to-go from original system state $i$ that has just been generated from an  aggregate state (left side of Fig.\ \figsft).
\nitem{} $\tl J_1(j)$ is the optimal cost-to-go from original system state $j$ that has just been generated from an original system state (right side of Fig.\ \figsft).
\smskip
\pn Note that because of the intermediate transitions to aggregate states, $\tl J_0$ and $\tl J_1$ are different.

These three vectors satisfy the following three Bellman's equations:
$$r^*_\ell=\sum_{i=1}^nd_{\ell i}\tl J_0(i),\qquad \ell=1,\ldots,q,\xdef\belleqo{\lab}\eqnum\show{lsmin}$$
$$\tl J_0(i)=\min_{u\in U(i)}\sum_{j=1}^n p_{ij}(u)\big(g(i,u,j)+\a \tl J_1(j)\big),\qquad i=1,\ldots,n,\xdef\belleqt{\lab}\eqnum\show{lsmin}$$
$$\tl J_1(j)=\sum_{m=1}^q\phi_{j\ell}r^*_m,\qquad j=1,\dots,n.\xdef\belleqth{\lab}\eqnum\show{lsmin}$$
By combining these equations, we see that $r^*$ satisfies
$$r^*_\ell=\sum_{i=1}^nd_{\ell i}\min_{u\in U(i)}\sum_{j=1}^n p_{ij}(u)\lf(g(i,u,j)+\a \sum_{m=1}^q\phi_{jm}\,r^*_{m}\ri),\qquad \ell=1,\ldots,q,\xdef\rstfixed{\lab}\eqnum\show{lsmin}$$
or equivalently $r^*=Hr^*$, where  $H$ is the mapping that maps the vector $r$ to the vector $Hr$ with components
$$(Hr)(\ell)=\sum_{i=1}^nd_{\ell i}\min_{u\in U(i)}\sum_{j=1}^n p_{ij}(u)\lf(g(i,u,j)+\a \sum_{m=1}^q\phi_{jm}\,r_{m}\ri),\qquad \ell=1,\ldots,q.\xdef\aggrmap{\lab}\eqnum\show{lsmin}$$
It can be shown that $H$ is a contraction mapping with respect to the sup-norm and thus has $r^*$ as its unique fixed point. This follows from standard contraction arguments, and the fact that $d_{\ell i}$, $p_{ij}(u)$, and $\phi_{j\ell}$ are probabilities. Note the nature of $r^*_\ell$: it is the optimal cost of the aggregate state $S_\ell$, which  is the restriction of the feature mapping $F$ on the disaggregation set $I_\ell$. Thus, roughly, $r^*_\ell$ is an approximate optimal cost associated with states in $I_\ell$.

\topinsert
\centerline{\includegraphics[width=5.4in]{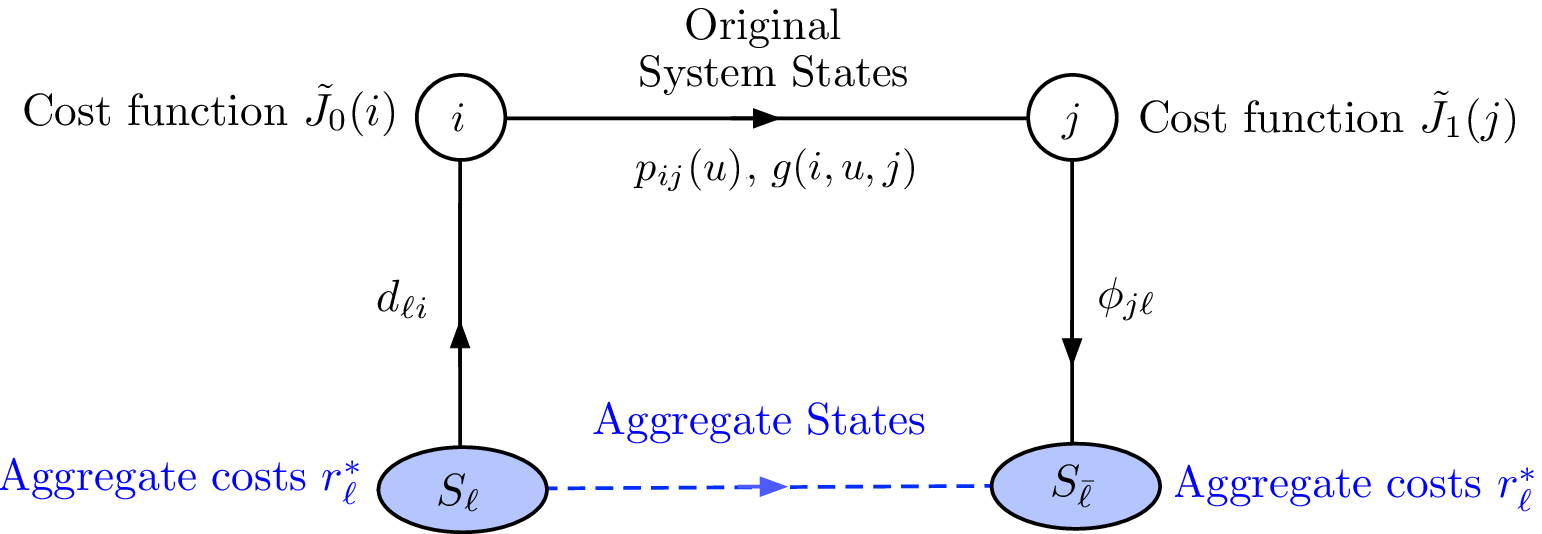}}
\vskip-0.5pc
\hskip-3pc\fig{0pc}{\figsft.} {The transition mechanism and the cost functions of the aggregate problem.}\endinsert

\subsubsection{Solution of the Aggregate Problem}
\vskip-0.5pc

\pn While the aggregate problem involves more states than the original DP problem, it is in fact easier in some important ways. The reason is that {\it it can be solved with algorithms that execute over the smaller space of aggregate states\/}. In particular, exact and approximate simulation-based algorithms, can be used to find the lower-dimensional vector $r^*$ without computing the higher-dimensional vectors $\tl J_0$ and $\tl J_1$. We describe some of these methods in Section 4.2, and we refer to Chapter 6 of [Ber12] for a more detailed discussion of simulation-based methods for computing the vector $r_\m$ of the costs of the aggregate states that correspond to a given policy $\m$. The simulator used for these methods is based on Figs.\ \figsfo\ and \figsft: transitions to and from the aggregate states are generated using the aggregation and disaggregation probabilities, respectively, while  transitions $(i,j)$ between original system states are generated using a simulator of the original system (which is assumed to be available).

\xdef\figcostinterpo{\figr}\figrnum\show{myfigure}

Once $r^*$ is found, the optimal-cost-to-go of the original problem may be approximated by the vector $\tl J_1$ of Eq.\ \belleqth. 
Note that  $\tl J_1$ is a ``piecewise linear" cost approximation of $\jstar$: it is constant over each of the disaggregation sets $I_\ell$, $\ell=1,\ldots,q$ [and equal to the optimal cost $r^*_\ell$ of the aggregate state $S_\ell$; cf.\ Eqs.\ \aggrconstr\ and \belleqth], and it is interpolated/linear outside the disaggregation sets [cf.\ Eq.\ \belleqth]. In the case where $\cup_{\ell=1}^q I_\ell=\{1,\ldots,n\}$ (e.g., in hard aggregation), the disaggregation sets $I_\ell$ form a partition of the original system state space, and $\tl J_1$ is piecewise constant. Figure \figcostinterpo\ illustrates a simple example of approximate cost function  $\tl J_1$.

\topinsert
\centerline{\includegraphics[width=3.7in]{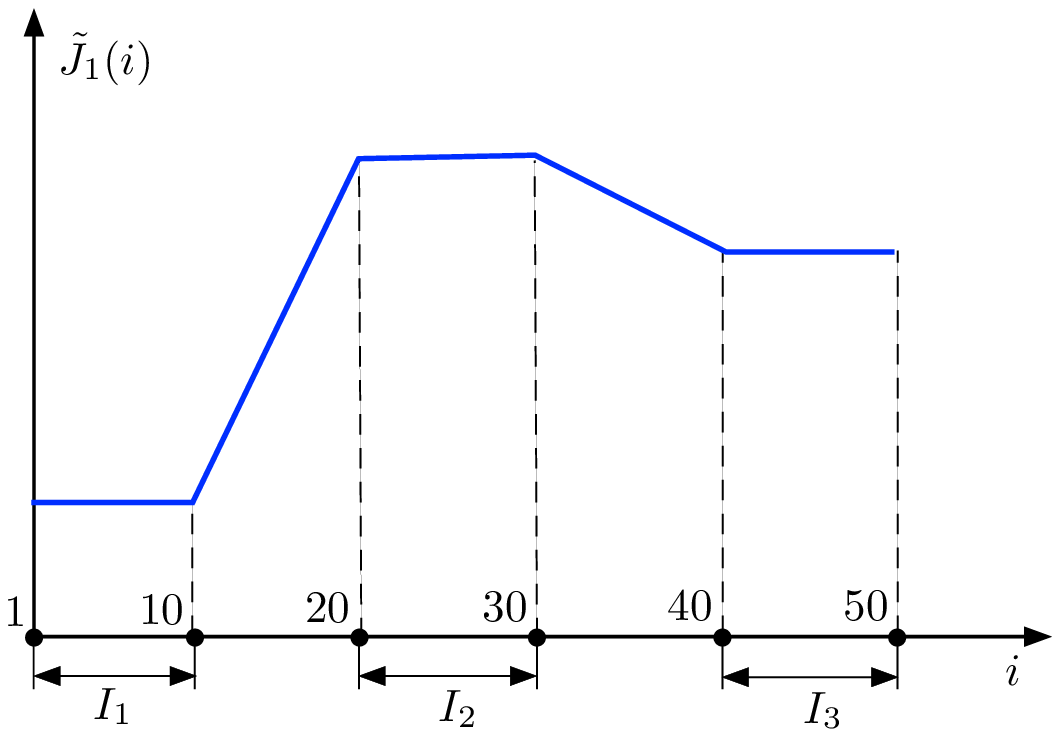}}
\vskip-2pc
\hskip-3pc\fig{0pc}{\figcostinterpo.} {Schematic illustration of the approximate cost function  $\skew5 \tl J_1$. Here the original states are the integers between 1 and  50. In this figure there are three aggregate states numbered $1, 2,3$. The corresponding disaggregation sets are $I_1=\{1,\ldots,10\}$, $I_2=\{20,\ldots,30\}$, $I_3=\{40,\ldots,50\}$ are shown in the figure. The values of the approximate cost function $\skew5 \tl J_1(i)$ are constant within each disaggregation set $I_\ell$, $\ell=1,2,3$, and are obtained by linear interpolation for states $i$ that do not belong to any one of the sets $I_\ell$. If the sets $I_\ell$, $\ell=1,2,3$, include all the states $1,\ldots,50$, we  have a case of hard aggregation. If each of the sets $I_\ell$, $\ell=1,2,3$, consist of a single state, we  have a case of aggregation with representative states. }\endinsert

Let us also note that for the purposes of using feature-based aggregation to improve a given policy $\m$, it is not essential to solve the aggregate problem to completion. Instead, we may perform  one or just a few PIs and adopt the final policy obtained as a new ``improved" policy. The quality of such a policy depends on how well the aggregate problem approximates the original DP problem. While it is not easy to quantify the relevant approximation error, generally a small error can be achieved if:

\nitem{(a)} The feature mapping $F$ ``conforms" to the optimal cost function $\jstar$ in the sense that $F$ varies little in regions of the state space where $\jstar$ also varies little. 
\nitem{(b)} The aggregate states are selected so that $F$ varies little over each of the disaggregation sets $I_1,\ldots,I_q$.
\smskip
\pn
This is intuitive and is supported by the subsequent discussion and analysis.

Given the optimal aggregate costs $r^*_\ell$, $\ell=1,\ldots,q$, the corresponding optimal policy is defined implicitly,  using the one-step lookahead minimization
$$\hat \m(i)\in\arg\min_{u\in U(i)}\sum_{j=1}^n p_{ij}(u)\left(g(i,u,j)+\a \sum_{\ell=1}^q\phi_{j\ell}\,r^*_{\ell}\right),\qquad i=1,\ldots,n,\xdef\optpolaggr{\lab}\eqnum\show{lsmin}$$
[cf.\ Eq.\ \belleqt] or a multistep lookahead variant. It is also possible to use a model-free implementation, as we describe next.

\subsubsection{Model-Free Implementation of the Optimal Policy of the Aggregate Problem}

\pn The computation of the optimal policy of the aggregate problem via Eq.\ \optpolaggr\ requires knowledge of the transition probabilities $p_{ij}(u)$ and the cost function $g$. Alternatively, this policy may be implemented in model-free fashion using a $Q$-factor architecture $\tl Q(i,u,\theta)$, as described in Section 2.4, i.e., compute sample approximate $Q$-factors
$$\b_m=g(i_m,u_m,j_m)+\a\sum_{\ell=1}^q\phi_{j_m\ell}\,r^*_{\ell},\qquad m=1,\ldots,M,\xdef\samplesaggr{\lab}\eqnum\show{lsmin}
$$
cf.\ Eq.\ \qsamples, compute $\tl \theta$ via a least squares regression
$$\tl \theta \in\arg\min_\theta\sum_{m=1}^M\big(\tl Q(i_m,u_m,\theta)-\b_m\big)^2\xdef\qregrlaggr{\lab}\eqnum\show{lsmin}$$
(or a regularized version thereof), cf.\ Eq.\ \qregression,  and approximate the optimal policy of the aggregate problem via 
$$\hat \m(i)\in\arg\min_{u\in U(i)}\tl Q(i,u,\tl \theta),\qquad i=1,\ldots,n,\xdef\qoptpolaggr{\lab}\eqnum\show{lsmin}$$
cf.\ Eq.\ \qpolicy.

\subsubsection{Error Bounds}

\pn Intuitively, if the disaggregation sets nearly cover the entire state space (in the sense that $\cup_{\ell=1,\ldots,q}I_\ell$ contains ``most" of the states $1,\ldots,n$) and $\jstar$ is nearly constant over each disaggregation set, then $\tl J_0$ and $\tl J_1$ should be close to $\jstar$. In particular, in the case of hard aggregation, we have the following error bound, due to Tsitsiklis and VanRoy [TsV96]. We adapt their proof to the notation and terminology of this paper.

\xdef\properrorbd{\propn}\propnum\show{myproposition}

\texshopbox{\proposition{\properrorbd:}In the case of hard aggregation, where $\cup_{\ell=1}^q I_\ell=\{1,\ldots,n\}$, and  Eqs.\ \hardaggrF, \hardaggrphi\ hold,  we have
$$\big|\jstar(i)-r^*_\ell\big|\le {\e\over 1-\a},\qquad  \forall\ i \hbox{ such that } i\in I_\ell,\ \ell=1,\ldots,q,\xdef\aggrbound{\lab}\eqnum\show{lsmin}$$
where
$$\e=\max_{\ell=1,\ldots,q}\,\max_{i,j\in I_\ell}\big|\jstar(i)-\jstar(j)\big|.\xdef\epsdef{\lab}\eqnum\show{lsmin}$$
}

\proof Consider the mapping 
$H$ defined by
Eq.\ \aggrmap, and consider the vector $\ol r$ with components defined by
$$\ol r_\ell=\min_{i\in I_\ell}\jstar (i)+{\e\over 1-\a},\qquad \ell\in 1,\ldots,q.$$
Denoting by $\ell(j)$ the index of the disaggregation set to which $j$ belongs, i.e., $j\in I_{\ell(j)}$, we have for all $\ell$,
$$\eqalignno{(H\ol r)(\ell)&=\sum_{i=1}^nd_{\ell i}\min_{u\in U(i)}\sum_{j=1}^n p_{ij}(u)\Big(g(i,u,j)+\a \ol r_{\ell(j)}\Big)\cr
&\le \sum_{i=1}^nd_{\ell i}\min_{u\in U(i)}\sum_{j=1}^n p_{ij}(u)\lf(g(i,u,j)+\a \jstar (j)+{\a\e\over 1-\a}\ri)\cr
&=\sum_{i=1}^nd_{\ell i}\lf(\jstar (i)+{\a\e\over 1-\a}\ri)\cr
&\le \min_{i\in I_\ell}\bl(\jstar(i)+\e\br)+{\a\e\over 1-\a}\cr
&=\min_{i\in I_\ell}\jstar(i)+{\e\over 1-\a}\cr
&=\ol r_\ell,\cr}$$
where for the second equality we used the Bellman equation for the original system, which is satisfied by $\jstar$, and for the second inequality we used Eq.\ \epsdef. Thus we have $H\ol r\le \ol r$, from which it follows that $r^*\le \ol r$ (since $H$ is monotone, which implies that the sequence $\{H^k\ol r\}$ is monotonically nonincreasing, and we have 
$$r^*=\lim_{k\to\infty}H^k\ol r$$
 since $H$ is a contraction). This proves one side of the desired error bound. The other side follows similarly.
 \qed

The scalar $\e$ of Eq.\ \epsdef\ is the maximum variation of optimal cost within the sets of the partition of the hard aggregation scheme. Thus the meaning of the preceding proposition is that if the optimal cost function $\jstar$ varies by at most $\e$ within each set of the partition, the hard aggregation scheme yields a piecewise constant approximation to the optimal cost function that is within 
${\e/(1-\a)}$ of the optimal. We know that for every  approximation $\tl J$ of $\jstar$ that is constant within each disaggregation set, the error 
$$\max_{i=1,\ldots,n} \big|\jstar(i)-\tl J(i)\big|$$
 is at least equal to $\e/2$. Based on the bound \aggrbound, the actual value of this error for the case where $\tl J$ is obtained by hard aggregation involves an additional multiplicative factor that is at most equal to $2/(1-\a)$, and depends on the disaggregation probabilities. In practice the bound  \aggrbound\ is typically conservative, and no examples are known where it is tight. Moreover, even for hard aggregation,  the manner in which the error  $\jstar-\tl J_1$ depends on the disaggregation distributions is complicated and is an interesting subject for research.

The following proposition extends the result of the preceding proposition to the case where  the aggregation probabilities are all either 0 or 1, in which case the cost function $\tl J_1$ obtained by aggregation is a piecewise constant function, but the disaggregation sets need not form a partition of the state space. Examples of this type of scheme include cases where the aggregation probabilities are generated by a ``nearest neighbor" scheme, and the cost $\tl J_1(j)$ of a state $j\notin \cup_{\ell=1}^q I_\ell$ is taken to be equal to the cost of the ``nearest" state within  $\cup_{\ell=1}^q I_\ell$.

\xdef\properrorbdt{\propn}\propnum\show{myproposition}

\texshopboxnb{\proposition{\properrorbdt:}Assume that each aggregation probability $\phi_{j\ell}$, $j=1,\ldots,n$, $\ell=1,\ldots,q$, is  equal to either 0 or 1, and consider the sets
$$\hat I_\ell=\{j\mid \phi_{j\ell}=1\},\qquad \ell=1,\ldots,q.$$}\texshopboxnt{\pn
Then we have
$$\big|\jstar(i)-r^*_\ell\big|\le {\e\over 1-\a},\qquad  \forall\ i\in \hat I_\ell,\ \ell=1,\ldots,q,$$
where
$$\e=\max_{\ell=1,\ldots,q}\,\max_{i,j\in \hat I_\ell}\big|\jstar(i)-\jstar(j)\big|.$$
}

\proof We first note that by the definition of a feature-based aggregation scheme, we have $I_\ell\subset \hat I_\ell$ for all $\ell=1,\ldots,q$, while the sets $\hat I_\ell$, $\ell=1,\ldots,q$, form a partition of the original state space, in view of our assumption on the aggregation probabilities. Let us replace the feature vector $F$ with another feature vector $\hat F$ of the form
$$\hat F(i)=\ell,\qquad \forall\ i\in\hat I_\ell,\ \ell=1,\ldots,q.$$
 Since the aggregation probabilities are all either 0 or 1, the resulting aggregation scheme with $I_\ell$ replaced by $\hat I_\ell$, and with the aggregation and disaggregation probabilities remaining unchanged, is a hard aggregation scheme. When the result of Prop.\ \properrorbd\ is applied to this hard aggregation scheme, the result of the present proposition follows.
\qed

The preceding propositions suggest the principal guideline for a feature-based aggregation scheme. It should be designed so that states that belong to the same disaggregation set have nearly equal optimal costs. In Section 4.3 we will elaborate on schemes that are based on this idea. In the next section we discuss the solution of the aggregate problem by simulation-based methods.

\subsection{Solving the Aggregate Problem with Simulation-Based Methods}

\pn  We will now focus on methods to compute the optimal cost vector $r^*$ of the aggregate problem that corresponds to the aggregate states. This is the unique solution of Eq.\ \rstfixed. We first note that since $r^*$,  together with the cost functions $\tl J_0$ and $\tl J_1$, form the solution of the Bellman equations \belleqo-\belleqth, they can all be computed with the classical (exact) methods of policy and value iteration (PI and VI for short, respectively). However, in this section, we will discuss specialized versions of PI and VI that compute just $r^*$ (which has relatively low dimension), but not $\tl J_0$ and $\tl J_1$ (which may have astronomical dimension). These methods are based on stochastic simulation
as they involve the aggregate problem, which is stochastic because of the disaggregation and aggregation probabilities, even if the original problem is deterministic. 

We start with simulation-based versions of PI, where policy evaluation is done with lookup table versions of classical methods such as LSTD(0), LSPE(0), and TD(0), applied to a reduced size DP problem whose states are just the aggregate states. 

\subsubsection{Simulation-Based Policy Iteration}
\vskip-6pt

\pn  
One possible way to compute $r^*$ is a PI-like algorithm, which generates sequences of policies $\{\m^k\}$  for the original problem and vectors $\{r^k\}$, which converge to an optimal policy and $r^*$, respectively. The algorithm does not compute any intermediate estimates of the high-dimensional vectors $\tl J_0$ and $\tl J_1$. It starts with a stationary policy $\m^0$ for the original problem, and given $\m^k$, it performs a  policy evaluation step by finding the unique fixed point of the contraction mapping $H_{\m^k}=DT_{\m^k} \Phi $  that maps the vector $r$ to the vector $H_{\m^k}r$ with components
$$(H_{\m^k} r)(\ell)=\sum_{i=1}^nd_{\ell i}\sum_{j=1}^n p_{ij}\bl(\m^k(i)\br)\lf(g\bl(i,\m^k(i),j\br)+\a \sum_{m=1}^q\phi_{jm}\,r_{m}\ri),\qquad  \ell=1,\ldots,q,$$
cf.\ Eq.\ \aggrmap. 
Thus the policy evaluation step finds $r^k=\{r^k_\ell\mid \ell=1,\ldots,q\}$ satisfying 
$$r^k=H_{\m^k} r^k=DT_{\m^k} \Phi r^k=D\big(g_{\m^k}+\a P_{\m^k}\Phi r^k\big),\xdef\fixedpointpoleval{\lab}\eqnum\show{lsmin}$$ 
where $P_{\m^k}$ is the transition probability matrix corresponding to $\m^k$, $g_{\m^k}$ is the expected cost vector of $\m^k$, i.e., the vector whose $i$th component is 
$$\sum_{j=1}^np_{ij}\bl(\m^k(i)\br)g\bl(i,\m^k(i),j\br),\qquad i=1,\ldots,n,$$
and $D$ and $\Phi$  are the matrices with rows the disaggregation and aggregation distributions, respectively. Following the policy evaluation step, the algorithm generates $\m^{k+1}$ by 
$$\m^{k+1}(i)=\arg\min_{u\in U(i)}\sum_{j=1}^n p_{ij}(u)\lf(g(i,u,j)+\a \sum_{m=1}^q\phi_{jm}r^k_m\ri),\qquad i=1,\ldots,n;\xdef\fpolimpr{\lab}\eqnum\show{lsmin}$$
this is the policy improvement step. In the preceding minimization we use one step lookahead, but a multistep lookahead or Monte Carlo tree search can also be used.

It can be shown that this algorithm converges finitely to the unique solution of Eq.\ \rstfixed\ [equivalently the unique fixed point of the mapping $H$ of Eq.\ \aggrmap]. An indirect way to show this is to use the convergence of  PI applied to the aggregate problem to generate a sequence $\{\m^k,r^k, \tl J_0^k,\tl J_1^k\}$. We provide a more  direct proof, which is essentially a special case of a more general convergence proof for PI-type methods given in Prop.\ 3.1 of [Ber11a]. The key fact here is that the linear mappings $DT_\m\Phi$ and $\Phi D T_\m$ are sup-norm contractions, and also have the monotonicity property of DP mappings, which is used in an essential way in the standard convergence proof of ordinary PI.

\xdef\proppiconv{\propn}\propnum\show{myproposition}

\texshopbox{\proposition{\proppiconv:} Let $\m^0$ be any policy and let $\{\m^k,r^k\}$ be a sequence generated by the PI algorithm \fixedpointpoleval-\fpolimpr. Then the sequence $\{r^k\}$ is monotonically nonincreasing (i.e., we have $r_\ell^k\ge r_{\ell}^{k+1}$ for all $\ell$ and $k$) and there exists an index $\bar k$ such that $r^{\bar k}$ is equal to $r^*$, the unique solution of Eq.\ \rstfixed.
}

\proof For each policy $\m$, we consider the linear mapping $\Phi D T_\m:\rn\mapsto\rn$ given by
$$\Phi D T_\m J=\Phi D(g_\m+\a P_\m J),\qquad J\in\rn.$$
This mapping is monotone in the sense that for all vectors $J$ and $J'$ with $J\ge J'$, we have 
$$\Phi D T_\m J\ge \Phi D T_\m J',$$
since the matrix $\a \Phi D P_\m$ of the mapping has nonnegative components. Moreover, the mapping is a contraction of modulus $\a$ with respect to the sup-norm. The reason is that the matrix $\Phi D P_\m$ is a transition probability matrix, i.e., it has nonnegative components and its row sums are all equal to 1. This can be verified by a straightforward calculation, using the fact that the rows of $\Phi$ and $D$ are probability distributions while $P_\m$ is a transition probability matrix. It can also be intuitively verified from the structure of the aggregate problem:  $\Phi D P_\m$ is the matrix of transition probabilities under policy $\m$ for the Markov chain whose $n$ states are the states depicted in the top righthand side of Fig.\ \figsfo.

Since the mapping $D T_{\m^k}\Phi$ has $r^k$ as its unique fixed point [cf.\ Eq.\ \fixedpointpoleval], we have $r^k=D T_{\m^k}\Phi r^k$, so that the vector 
$$\tl J_{\m^k}=\Phi r^k$$
satisfies
$$\tl J_{\m^k}=\Phi D T_{\m^k}\tl J_{\m^k}.$$
It follows that $\tl J_{\m^k}$ is the unique fixed point of the contraction mapping $\Phi D T_{\m^k}$. By using the definition 
$$T_{\m^{k+1}}\tl J_{\m^k}=T\tl J_{\m^k}$$ 
of $\m^{k+1}$ [cf.\ Eq.\ \fpolimpr], we have
$$\tl J_{\m^k}=\Phi D T_{\m^k}\tl J_{\m^k}\ge\Phi D T\tl J_{\m^k} =\Phi D T_{\m^{k+1}}\tl J_{\m^k}.\xdef\ineqstringo{\lab}\eqnum\show{lsmin}$$
Applying repeatedly  the monotone mapping $\Phi D T_{\m^{k+1}}$ to this relation, we have for all $m\ge 1$, 
$$\tl J_{\m^k}\ge (\Phi D T_{\m^{k+1}})^m\tl J_{\m^k}\ge \lim_{m\to\infty}(\Phi D T_{\m^{k+1}})^m\tl J_{\m^k}=\tl J_{\m^{k+1}},\xdef\ineqstringt{\lab}\eqnum\show{lsmin}$$
where the equality follows from the fact that $\tl J_{\m^{k+1}}$ is the fixed point of the contraction mapping $\Phi D T_{\m^{k+1}}$. It follows that $\tl J_{\m^k}\ge \tl J_{\m^{k+1}}$, or equivalently
$$\Phi r^k\ge \Phi r^{k+1},\qquad k=0,1,\ldots.$$
By the definition of the feature-based aggregation architecture [cf.\ Eq.\ \aggrconstr], each column of $\Phi$ has at least one component that is equal to 1. Therefore we have for all $k$
$$r^k\ge r^{k+1},$$
and moreover, the equality $r^k= r^{k+1}$ holds if and only if $\tl J_{\m^k}=\tl J_{\m^{k+1}}.$

The inequality $\tl J_{\m^k}\ge \tl J_{\m^{k+1}}$ implies that as long as $\tl J_{\m^k}\ne \tl J_{\m^{k+1}}$, a policy $\m^k$ cannot be repeated. Since there is only a finite number of policies, it follows that we must eventually have $\tl J_{\m^k}= \tl J_{\m^{k+1}}$. In view of Eqs.\ \ineqstringo-\ineqstringt, we see that 
$$\tl J_{\m^k}=\Phi D T\tl J_{\m^k}$$
or
$$\Phi r^{k}=\Phi D T\Phi r^{k}.$$
Since each column of $\Phi$ has at least one component that is equal to 1, it follows that $r^{k}$ is a fixed point of the mapping $H=DT\Phi$ of Eq.\ \aggrmap, which is $r^*$ by Eq.\ \rstfixed.
\qed

To avoid the $n$-dimensional calculations of the policy evaluation step in the PI algorithm \fixedpointpoleval-\fpolimpr, one may use simulation. In particular, the policy evaluation equation, $r=H_\m r$, is linear of the form
$$r=Dg_\m+\a DP_\m \Phi r,\xdef\polevalaggr{\lab}\eqnum\show{lsmin}$$
 [cf.\ Eq.\ \fixedpointpoleval]. Let us  write this equation as
$Cr=b$, where 
 $$C=I-\a D P_\m \Phi,\qquad b=Dg_\m,$$
and note that it is Bellman's equation for a policy with cost per stage vector equal to $D g_\m$ and transition probability matrix equal to $D P_\m\Phi$. This is the transition matrix under policy $\m$ for the Markov chain whose states are the aggregate states. The solution $r_\m$ of the policy evaluation Eq.\ \polevalaggr\ is the cost vector corresponding to this Markov chain, and can be found by using simulation-based methods with lookup table representation. 

In particular, we may use model-free simulation to  approximate $C$ and $b$, and then solve the system $Cr=b$ approximately. To this end, we obtain a  sequence of sample transitions $\big\{(i_1,j_1),(i_2,j_2),\ldots\big\}$ by first generating a sequence of states $\{i_1,i_2,\ldots\}$ according to some distribution $\{\xi_i\mid i=1,\ldots,n\}$ (with $\xi_i>0$ for all $i$), and then  generate for each $m\ge1$ a sample transition $(i_m,j_m)$ according to the  distribution $\{p_{i_mj}\mid j=1,\ldots,n\}$. 
Given the first $M$ samples, we form the matrix $\widehat C_M$ and vector $\hat b_M$ given by
$$\widehat C_M=I-{\a \over M}\sum_{m=1}^M{1\over \xi_{i_m}}d(i_m)\phi(j_m)',\qquad \hat b_M={1\over M}\sum_{m=1}^M{1\over \xi_{i_m}}d(i_m)g\big(i_m,\m(i_m),j_m\big),\xdef\saggrcandd{\lab}\eqnum\show{lsmin}$$
where $d(i)$ is the $i$th column of $D$ and $\phi(j)'$ is the $j$th row of $\Phi$.
We can then show that $\widehat C_M\to C$ and $\hat b_M\to d$ by using law of large numbers arguments, i.e., writing
$$C=I-\a\sum_{i=1}^n\sum_{j=1}^np_{ij}\big(\m(i)\big)d(i)\phi(j)',\qquad b=\sum_{i=1}^n\sum_{j=1}^np_{ij}\big(\m(i)\big)d(i)g\bl(i,\m(i),j\br),$$
multiplying and dividing $p_{ij}\big(\m(i)\big)$ by $\xi_i$ in order to properly view these expressions as expected values, and using the relation
$$\lim_{M\to\infty}{\hbox{Number of occurrences of the $i$ to $j$ transition from time $m=1$ to $m=M$}\over M}= \xi_i \,p_{ij}\big(\m(i)\big).$$

The corresponding estimates 
$$\hat r_M=\widehat C_M^{-1}\hat b_M$$
converge to the unique solution of the policy evaluation Eq.\ \polevalaggr\ as $M\to\infty$, and provide the estimates  $\Phi\hat r_M$ of the cost vector $J_{\m}$ of $\m$:
$$\tl J_\m=\Phi\hat r_M.$$
This is the aggregation counterpart of the LSTD(0) method. One may also use an iterative simulation-based LSPE(0)-type method or a TD(0)-type method to solve the equation $Cr=b$; see [Ber12].  

Note that instead of using the probabilities $\xi_i$ to sample directly original system states, we may alternatively sample the aggregate states $S_\ell$ according to some distribution $\{\zeta_\ell\mid \ell=1,\ldots,q\}$, generate a sequence of aggregate states $\{S_{\ell_1},S_{\ell_2},\ldots\}$, and then generate a state sequence  $\{i_1,i_2,\ldots\}$ using the disaggregation probabilities. In this case Eq.\ \saggrcandd\ should be modified as follows:
$$\widehat C_M=I-{\a \over M}\sum_{m=1}^M{1\over \zeta_{\ell_m}d_{\ell_m i_m}}d(i_m)\phi(j_m)',\qquad \hat b_M={1\over M}\sum_{m=1}^M{1\over \zeta_{\ell_m}d_{\ell_m i_m}}d(i_m)g\big(i_m,\m(i_m),j_m\big).$$

The main difficulty with the policy improvement step at a given state $i$ is the need to compute the expected value in the  $Q$-factor expression
$$\sum_{j=1}^n p_{ij}(u)\lf(g(i,u,j)+\a \sum_{\ell=1}^q\phi_{j\ell}r^k_\ell\ri)$$
that is minimized over $u\in U(i)$. If the transition probabilities $p_{ij}(u)$ are available and the number of successor states [the states $j$ such that $p_{ij}(u)>0$] is small, this expected value may be easily calculated (an important case where this is so is when the system is deterministic).  Otherwise, one may consider approximating this expected value using one of the model-free schemes described in Section 2.4.

\subsubsection{Simulation-Based Value Iteration and $Q$-Learning}
\vskip-6pt

\pn An exact VI algorithm for obtaining $r^*$ is the fixed point iteration
$$r^{k+1}=Hr^k,$$
starting from some initial guess $r^0$, where $H$ is the contraction mapping of Eq.\ \aggrmap. A stochastic approximation-type algorithm based on this fixed point iteration generates a sequence of aggregate states $\{S_{\ell_0},S_{\ell_1},\ldots\}$ by some probabilistic mechanism, which ensures that all aggregate states are generated infinitely often. Given $r^k$ and $S_{\ell_k}$, it independently generates an original system state $i_k$ according to the probabilities $d_{\ell i}$, and updates the component $r_{\ell_k}$ according to
$$r^{k+1}_{\ell_k}=(1-\g_k)r^k_{\ell_k}+\g_k \min_{u\in U(i)}\sum_{j=1}^n p_{i_k j}(u)\lf(g(i_k,u,j)+\a \sum_{\ell =1}^q\phi_{j\ell}r^k_{\ell}\ri),$$
where $\g_k$ is a diminishing positive stepsize, and leaves all the other components unchanged:
$$r^{k+1}_\ell=r^k_\ell,\qquad \hbox{if }\ell\ne \ell_k.$$
This algorithm can be viewed as an asynchronous stochastic approximation version of VI.  The stepsize $\g_k$ should be diminishing (typically at the rate of $1/k$), and its justification and convergence mechanism are very similar to the ones for the $Q$-learning algorithm. We refer to the paper by Tsitsiklis and VanRoy [TsV96] for further discussion and analysis (see also [BeT96], Section 3.1.2 and 6.7).

A somewhat different algorithm is possible in the case of hard aggregation, assuming that for every $\ell$, the set $U(i)$ is the same for all states $i$ in the disaggregation set $I_\ell$. Then, as discussed in [BeT96], Section 6.7.7, we can introduce $Q$-factors that are constant within each set $I_\ell$ and have the form
$$\tl Q(i,u)=Q(\ell,u),\qquad \ i\in I_\ell,\ u\in U(i).$$
We then obtain an algorithm
that updates the $Q$-factors $ Q(\ell,u)$ one at a time, using a $Q$-learning-type iteration of the form
$$Q(\ell,u):=(1-\gamma)Q(\ell,u)+\gamma \Big(g(i,u,j)+\a \min_{v\in U(j)}Q\big({m(j)},v\big)\Big),$$
where $i$ is a state within $I_\ell$ that is chosen with probability $d_{\ell i}$,
$j$ is the outcome of a transition simulated according to the
transition probabilities $p_{ij}(u)$, the index $m(j)$ corresponds to the aggregate state $S_{m(j)}$ to which $j$ belongs, and $\g$ is the stepsize. It can be seen that this algorithm coincides with
$Q$-learning with lookup table representation, applied to a
lower dimensional aggregate DP problem that involves just the aggregate states. With a suitably decreasing stepsize $\g$
and assuming that each pair $(\ell,u)$ is simulated an infinite number of times,
the standard convergence results for $Q$-learning [Tsi94] apply. 

We note, however, that the $Q$-learning algorithm just described has a substantial drawback. It solves an aggregate problem that differs from the aggregate problem described in Section 4.1, because implicit in the algorithm is the restriction that the same control is applied at all states $i$ that belong to the same disaggregation set. In effect, we are assigning controls to subsets of states (the disaggregation sets) and not to individual states of the original problem. Clearly this is a coarser form of control, which is inferior in terms of performance. However, the $Q$-learning algorithm may find some use in the context of initialization of another algorithm that aspires to better performance.

\subsection{Feature Formation by Using Scoring Functions}

\pn  The choice of the feature mapping $F$ and the method to obtain aggregate states are clearly critical for the success of feature-based aggregation. In the subsequent Section 5 we will discuss how deep neural network architectures can be used for this purpose. In what follows in this section we consider some simple forms of feature mappings that can be used when we already have a reasonable estimate of the optimal cost function $\jstar$ or the cost function $J_\m$ of some policy $\m$, which we can use to group together states with similar estimated optimal cost. Then the aggregation approach can provide an improved piecewise constant or piecewise linear cost approximation. We provide some simple illustrative examples of this approach in Section 4.5. 

In particular, suppose that we have obtained in some way a real-valued {\it scoring function} $V(i)$ of the state $i$, which serves as an index of undesirability of state $i$ as a starting state (smaller values of $V$ are assigned to more desirable states, consistent with the view of $V$ as some form of ``cost" function). One possibility is to use as $V$ an approximation  of the cost function of some ``good" (e.g., near-optimal) policy. Another possibility is to obtain  $V$ as the cost function of some reasonable policy applied to an approximation of the original problem (e.g., a related problem that can be solved more easily either analytically or computationally; see [Ber17], Section 6.2). Still another possibility is to obtain $V$ by  training a neural network or other architecture using samples of state-cost pairs obtained by using  a software or human expert, and some supervised learning technique, such as for example Tesauro's comparison learning scheme [Tes89b], [Tes01].
Finally, one may compute $V$ using some form of policy evaluation algorithm like TD($\l$). 

\xdef\figscoreaggr{\figr}\figrnum\show{myfigure}

Given the scoring function $V$, we will construct a feature mapping that groups together states $i$ with roughly equal scores $V(i)$. In particular, we let $R_1,\ldots, R_q$ be $q$ disjoint intervals that form a partition of the set of possible values of $V$ [i.e., are such that for any state $i$, there is a unique interval $R_\ell$ such that  $V(i)\in R_\ell$]. We define a  feature vector $F(i)$ of the state $i$ according to
$$F(i)=\ell,\qquad \forall\ i \hbox{ such that }V(i)\in R_\ell,\quad \ell=1,\ldots,q.\xdef\featuredef{\lab}\eqnum\show{qfortyss}$$
This feature vector in turn defines a partition of the state space into the sets
$$I_\ell=\big\{i\mid F(i)=\ell\big\}=\big\{i\mid V(i)\in R_\ell\big\},\qquad \ell=1,\ldots,q.\xdef\disagrsetdef{\lab}\eqnum\show{qfortyss}$$
Assuming that all the sets $I_\ell$ are nonempty, we thus obtain a hard aggregation scheme, with aggregation probabilities defined by Eq.\ \hardaggrphi; see Fig.\ \figscoreaggr.

\topinsert
\centerline{\includegraphics[width=5in]{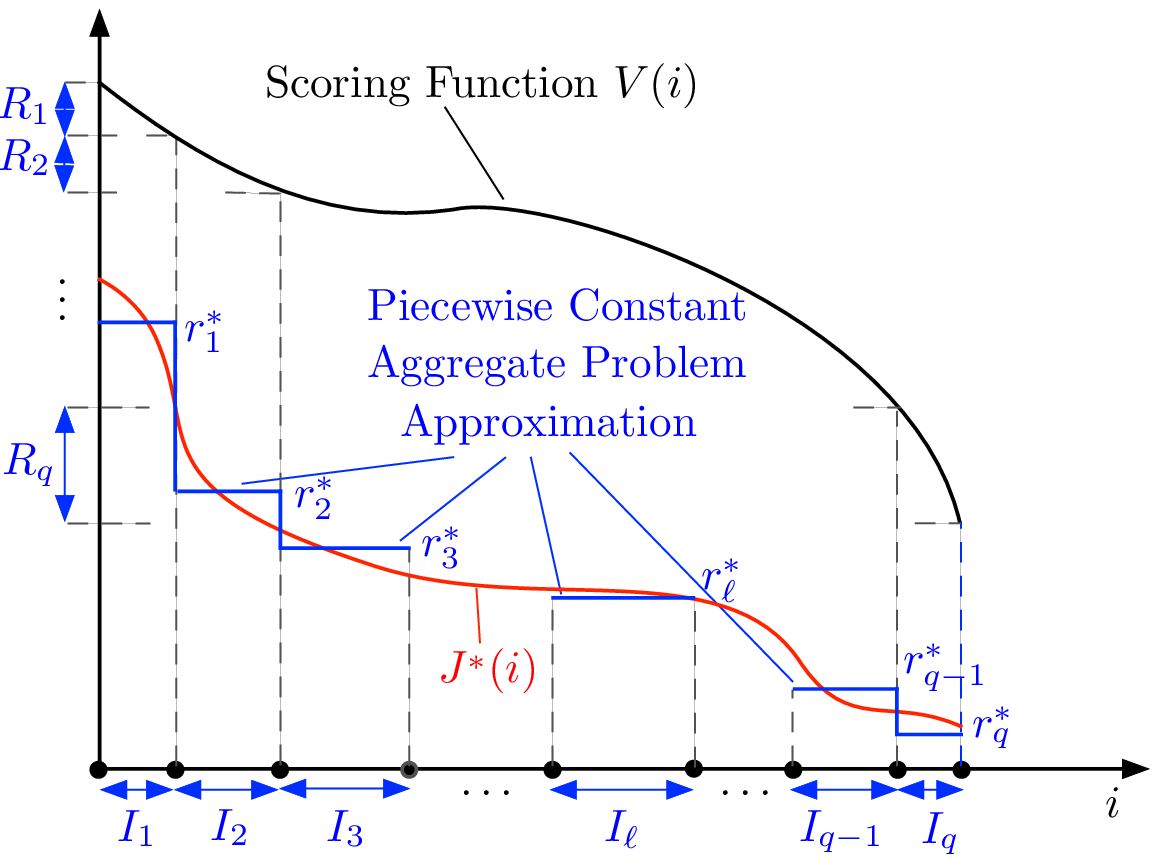}}
\vskip-1.0pc
\hskip-3pc\fig{0pc}{\figscoreaggr.} {Hard aggregation scheme based on a single scoring function. We introduce $q$ disjoint intervals $R_1,\ldots, R_q$ that form a partition of the set of possible values of $V$, and we define a  feature vector $F(i)$ of the state $i$ according to
$$F(i)=\ell,\qquad \forall\ i \hbox{ such that }V(i)\in R_\ell,\ \ell=1,\ldots,q.$$
This feature vector in turn defines a partition of the state space into the sets
$$I_\ell=\big\{i\mid F(i)=\ell\big\}=\big\{i\mid V(i)\in R_\ell\big\},\qquad \ell=1,\ldots,q.$$
The solution of the aggregate problem provides a piecewise constant approximation of the optimal cost function of the original problem.
}\endinsert

\xdef\figscoreaggrrep{\figr}\figrnum\show{myfigure}

A related scoring function scheme may be based on representative states. Here the aggregate states and the disaggregation probabilities are obtained by forming a fairly large sample set of states 
$\{i_m\mid m=1,\ldots,M\},$
 by computing their corresponding scores 
$$\big\{V(i_m)\mid m=1,\ldots,M\big\},$$
 and by suitably dividing the range of these scores into disjoint intervals  $R_1,\ldots, R_q$ to form the aggregate states, similar to Eqs.\ \featuredef-\disagrsetdef. 
Simultaneously we obtain subsets of sampled states $\hat I_\ell\subset I_\ell$ to which we can assign positive disaggregation probabilities. Figure \figscoreaggrrep\ illustrates this idea for the case where each subset $\hat I_\ell$ consists of a single (representative) state. This is a form of ``discretization" of the original state space based on the score values of the states. As the figure indicates, the role of the scoring function is to assist in forming a set of states that is small (to keep the aggregate DP problem computations manageable) but representative (to provide sufficient detail in the approximation of $\jstar$, i.e., be dense in the parts of the state space where $\jstar$ varies a lot, and sparse in other parts).

\topinsert
\centerline{\includegraphics[width=4.9in]{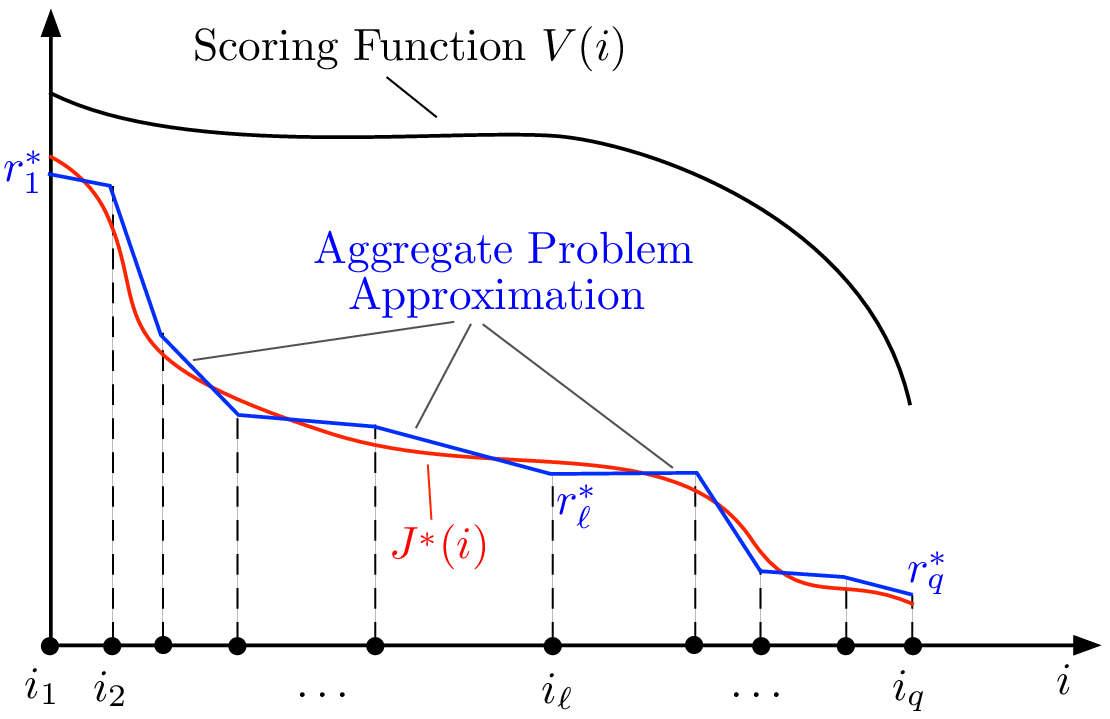}}
\vskip-1.0pc
\hskip-3pc\fig{0pc}{\figscoreaggrrep.} {Schematic illustration of aggregation based on sampling states and using a scoring function $V$ to form a representative set $i_1,\ldots,i_q$. A piecewise linear approximation of $J^*$ is obtained by using the corresponding aggregate costs $r^*_1,\ldots,r^*_q$ and the aggregation probabilities.}\endinsert

The following proposition illustrates the important role of the {\it quantization error\/}, defined as
$$\d=\max_{\ell=1,\ldots,q}\max_{i,j\in I_\ell}\big|V(i)-V(j)\big|.\xdef\quantization{\lab}\eqnum\show{qfortyss}$$
It represents the maximum error that can be incurred by approximating $V$ within each set $I_\ell$ with a single value from its range within the subset.

\xdef\propexactapprox{\propn}\propnum\show{myproposition}

\texshopbox{\proposition{\propexactapprox:}Consider the hard aggregation scheme defined by a scoring function $V$ as described above. Assume that the variations of $\jstar$ and $V$ over the sets $I_1,\ldots,I_q$ are within a factor $\b\ge0$ of each other, i.e., that
$$\big|\jstar(i)-\jstar(j)\big|\le \b\, \big|V(i)-V(j)\big|,\qquad \forall\ i,j\in I_\ell,\ \ell=1,\ldots,q.$$
\nitem{(a)} We have
$$\big|\jstar(i)-r^*_{\ell}\big|\le {\b \d\over 1-\a},\qquad  \forall\ i\in I_\ell,\ \ell=1,\ldots,q,$$
where
$\d$ is the quantization error of Eq.\ \quantization.
\nitem{(b)} Assume that there is no quantization error, i.e.,  $V$ and $\jstar$ are constant within each set $I_\ell$. Then the aggregation scheme yields the optimal cost function $\jstar$ exactly, i.e.,
$$\jstar(i)=r^*_{\ell},\qquad\forall\ i\in I_\ell,\ \ell=1,\ldots,q.$$
}

\proof (a) Since we are dealing with a hard aggregation scheme, the result of Prop.\ \properrorbd\ applies. By our assumptions, the maximum variation of $\jstar$ over the disaggregation sets $I_\ell$ is bounded by $\e=\b\d$, and the result of part (a) follows from Prop.\ 4.1.
\smskip
\pn (b) This is a special case of part (a) with $\d=\e=0$. \qed

Examples of scoring functions that may be useful in various settings are cost functions of nearly optimal policies, or approximations to such cost functions, provided for example by a neural network or other approximation schemes. Another example, arising in the adaptive aggregation scheme proposed by Bertsekas and Castanon [BeC89], is to use as $V(i)$ the residual vector $(TJ)(i)-J(i)$, where $J$ is some approximation to the optimal cost function $\jstar$, or the residual vector $(T_\m J)(i)-J(i)$, where $J$ is some approximation to the cost function of a policy $\m$; see also Keller, Mannor,  and Precup [KMP06]. 
Note that it is not essential that $V$ approximates well $\jstar$ or $J_\m$. What is important is that states with similar values of $\jstar$ or $J_\m$ also have similar values of $V$.

\subsubsection{Scoring Function Scheme with a State Space Partition}
 
\pn Another useful scheme is based on a scoring function $V$, which is defined separately on each one of a collection of disjoint subsets $C_1,\ldots,C_m$ that form a partition of the state space. We define a feature vector $F(i)$ that depends not only on the value of $V(i)$ but also on the membership of $i$ in the subsets of the partition. In particular, for each $\theta=1,\ldots,m$, let $R_{1\theta},\ldots, R_{q\theta}$ be $q$ disjoint intervals that form a partition of the set of possible values of $V$ over the set $C_\theta$. We then define
$$F(i)=(\ell,\theta),\qquad \forall\ i\in C_\theta\hbox{ such that }V(i)\in R_{\ell\theta}.\xdef\featuredef{\lab}\eqnum\show{qfortyss}$$
This feature vector in turn defines a partition of the state space into the $qm$ sets
$$I_{\ell\theta}=\big\{i\mid F(i)=(\ell,\theta)\big\}=\big\{i\in C_\theta\mid V(i)\in R_{\ell\theta}\big\},\qquad \ell=1,\ldots,q,\ \theta=1,\ldots,m,$$
which represent the disaggregation sets of the resulting hard aggregation scheme. In this scheme the aggregate states depend not only on the values of $V$ but also on the subset $C_\theta$ of the partition.

\subsubsection{Using Multiple Scoring Functions}

\pn The approach of forming features using a single scoring function can  be extended to the case where we have a vector of scoring functions $V(i)=\big(V_1(i),\ldots, V_s(i)\big)$. Then we can partition the set of possible  values of $V(i)$ into $q$ disjoint subsets $R_1,\ldots, R_q$ of the $s$-dimensional space $\re^s$, define a feature vector $F(i)$ according to
$$F(i)=\ell,\qquad \forall\ i \hbox{ such that }V(i)\in R_\ell,\ \ell=1,\ldots,q,\xdef\featuredefmulti{\lab}\eqnum\show{qfortyss}$$
and proceed as in the case of a scalar scoring function, i.e., construct a hard aggregation scheme with disaggregation sets given by
$$I_\ell=\big\{i\mid F(i)=\ell\big\}=\big\{i\mid V(i)\in R_\ell\big\},\qquad \ell=1,\ldots,q.$$

One possibility to obtain multiple scoring functions is to start with a single fairly simple scoring function, obtain aggregate states as described earlier, solve the corresponding aggregate problem, and use the optimal cost function of that problem as an additional scoring function. This is reminiscent of {\it feature iteration\/}, an idea that has been suggested in several approximate DP works. 

A related and complementary possibility is to somehow construct multiple policies, evaluate each of these policies (perhaps approximately, using a neural network), and use the policy cost function evaluations as scoring functions. This possibility may be particularly interesting in the case of a deterministic discrete optimization problem. The reason is that the deterministic character of the problem may obviate the need for expensive simulation and neural network training, as we discuss in the next section.

\subsection{Using Heuristics to Generate Features - Deterministic Optimization and Rollout}

\pn An important context where it is natural to use multiple scoring functions is general deterministic optimization problems with a finite search space. For such problems  simple heuristics are often available to obtain suboptimal solutions from various starting conditions, e.g., greedy algorithms of various kinds. The cost of each heuristic can then be used as a scoring function after the problem is converted to a finite horizon DP problem. The formulation that we will use in this section is very general and for this reason the number of states of the DP problem may be very large. Alternative DP reformulations with fewer states may be obtained by exploiting the structure of the problem. For example shortest path-type problems and discrete-time finite-state deterministic optimal control control problems 
can be naturally posed as DP problems with a simpler and more economical formulation than the one given here. In such cases the methodology to be described can be suitably adapted to exploit the problem-specific structural characteristics. 

The general discrete optimization problem that we consider in this section is
$$\eqalign{&\hbox{minimize\ \ }G(u)\cr &\hbox{subject to\ \
}u\in U,\cr}\xdef\dcost{\lab}\eqnum\show{hcost}$$
where $U$ is a finite set of feasible solutions and $G(u)$ is a cost function.  We assume that each solution $u$ has $N$ components; i.e., it has the form $u=(u_1,\ldots,u_N)$, where $N$ is a positive integer.  We can then view the problem as
a sequential decision problem, where the components $u_1,\ldots,u_N$ are selected one-at-a-time. An
$m$-tuple $(u_1,\ldots,u_m)$  consisting of the first $m$ components of a solution is called
an {\it
$m$-solution\/}. We associate $m$-solutions with the $m$th stage of a finite horizon DP problem.\footnote\dag{\ninepoint Our aggregation framework of Section 4.1 extends in a straightforward manner to finite-state finite-horizon problems. The main difference is that optimal cost functions, feature vectors, and scoring functions are not only state-dependent but also stage-dependent. In effect the states are the $m$-solutions for all values of $m$.} In
particular,  for $m=1,\ldots,N$, the states of the $m$th stage are of the form
$(u_1,\ldots,u_m)$. The initial state is a dummy (artificial) state. From this state
we may move to any state
$(u_1)$, with
$u_{1}$ belonging to the  set 
$$U_{1}=\bl\{\tl u_1\mid \hbox{there exists a solution of the form }(\tl
u_1,\tl u_2,\ldots,\tl u_N)\in U\br\}.$$ 
Thus $U_1$ is the set of choices of $u_1$ that are consistent with feasibility.

\xdef\figdiscropt{\figr}\figrnum\show{myfigure}

More generally, from
a state
$(u_1,\ldots,u_{m}),$
we may move to any state of the form
$(u_1,\ldots,u_{m},u_{m+1}),$
with $u_{m+1}$ belonging to the set
$${U_{m+1}(u_1,\ldots,u_{m})=\big\{\tl u_{m+1}\mid\ \hbox{there exists a solution of the form
}(u_1,\ldots,u_{m},\tl u_{m+1},\ldots,\tl u_N)\in U\big\}.}\xdef\arcs{\lab}\eqnum\show{arcs}$$
The choices available at state
$(u_1,\ldots,u_{m})$ are
$u_{m+1}\in U_{{m+1}}(u_1,\ldots,u_{m})$. These are the choices of $u_{m+1}$ that are consistent with the
preceding choices $u_1,\ldots,u_{m}$, and are also consistent with feasibility. The terminal states
correspond to the $N$-solutions $u=(u_1,\ldots,u_N)$, and the only nonzero cost is the terminal cost
$G(u)$. This terminal cost is incurred upon transition from $u$ to an artificial termination state; see Fig.\ \figdiscropt.

\topinsert
\centerline{\includegraphics[width=5.8in]{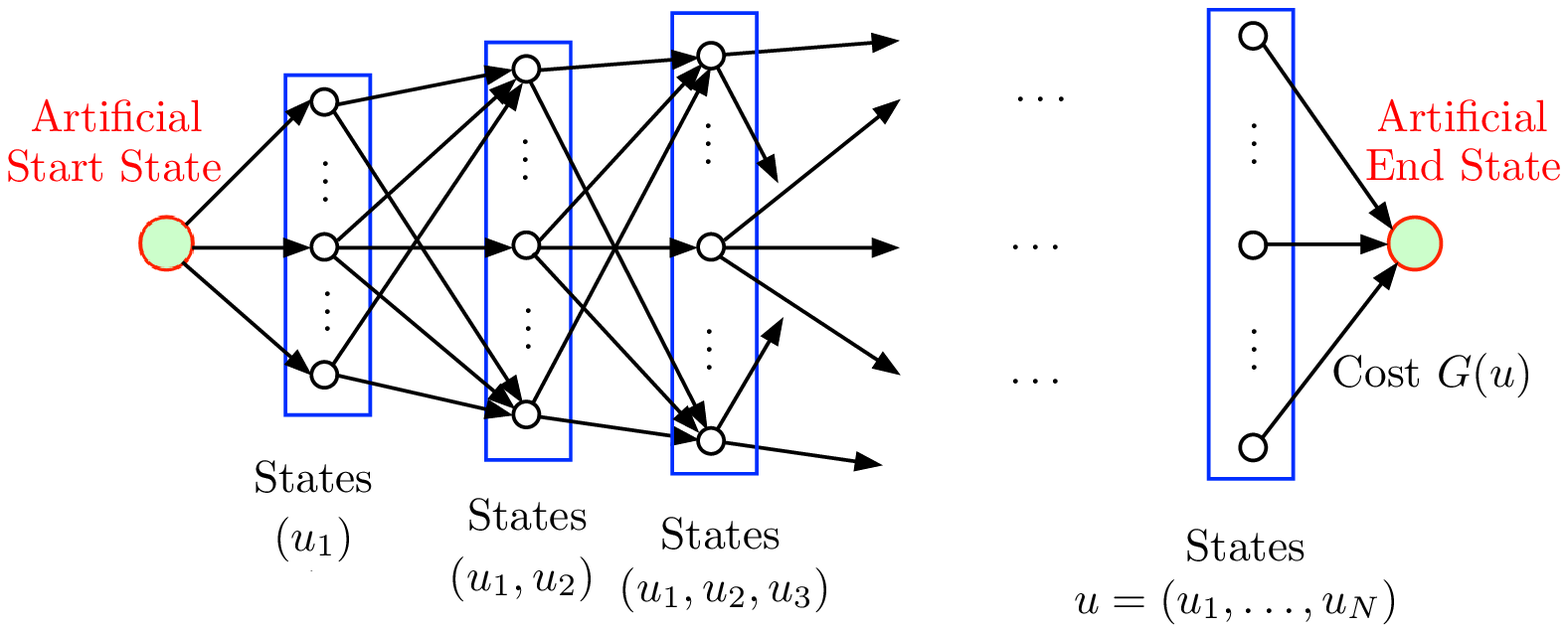}}
\vskip-0.3pc
\hskip-3pc\fig{0pc}{\figdiscropt.} {Formulation of a discrete optimization problem as a DP problem. There is a cost $G(u)$ only at the terminal stage on the arc connecting an $N$-solution $u=(u_1,\ldots,u_N)$ to the artificial terminal state. Alternative formulations may use fewer states by taking advantage of the problem's structure.}\endinsert

Let $\jstar(u_1,\ldots,u_{m})$ denote the optimal cost starting from the $m$-solution
$(u_1,\ldots,u_{m})$, i.e., the optimal cost of the problem over solutions
whose first $m$ components are constrained to be equal to $u_i$, $i=1,\ldots,m$,
respectively. If we knew the optimal cost-to-go functions $\jstar(u_1,\ldots,u_{m})$, we
could construct an optimal solution by a sequence of $N$ single component
minimizations. In particular, an optimal solution
$(u_1^*,\ldots,u_N^*)$ could be obtained sequentially, starting with $u_1^*$ and proceeding forward to $u_N^*$, through the algorithm
$$u_{m+1}^*\in\arg\min_{u_{m+1}\in
U_{m+1}(u_1^*,\ldots,u_{m}^*)}\jstar(u_1^*,\ldots,u_{m}^*,u_{m+1}),\qquad
m=0,\ldots,N-1.$$
Unfortunately, this is seldom viable, because of the prohibitive
computation required to obtain the functions
$\jstar(u_1,\ldots,u_{m})$. 

Suppose that we have $s$ different heuristic algorithms, which we can apply for suboptimal solution. We assume that each of these algorithms can start from  any
$m$-solution $(u_1,\ldots,u_m)$ and
produce an $N$-solution $(u_1,\ldots,u_m,u_{m+1},\ldots,u_N)$. The costs thus generated by the $s$ heuristic algorithms are denoted by
$V_1(u_1,\ldots,u_m),\ldots,V_s(u_1,\ldots,u_m),$
 respectively, and the corresponding vector of heuristic costs is denoted by
 $$V(u_1,\ldots,u_m)=\big(V_1(u_1,\ldots,u_m),\ldots,V_s(u_1,\ldots,u_m)\big).$$
 Note that the heuristic algorithms can be quite sophisticated, and at a given partial solution $(u_1,\ldots,u_m)$, may involve multiple component choices from $(u_{m+1},\ldots,u_N)$ and/or suboptimizations that may depend on the previous choices $u_1,\ldots,u_m$ in complicated ways. In fact, the heuristic algorithms may require some preliminary experimentation and training, using for example, among others, neural networks.

\xdef\figdiscroptagr{\figr}\figrnum\show{myfigure}

The main idea now is to  use the heuristic cost functions as scoring functions to construct a feature-based hard aggregation framework.\footnote{\dag}{\ninepoint There are several variants of this scheme, involving for example a state space partition as in Section 4.3. Moreover, the method of partitioning the decision vector $u$ into its components $u_1,\ldots,u_N$ may be critically important in specific applications.} In particular, for each  $m=1,\ldots,N-1$, we partition the set of possible  values of $V(u_1,\ldots,u_m)$ into $q$ disjoint subsets $R_1^m,\ldots, R_q^m$, we define a feature vector $F(u_1,\ldots,u_m)$ according to
$$F(u_1,\ldots,u_m)=\ell,\qquad \forall\ (u_1,\ldots,u_m)\hbox{ such that }V(u_1,\ldots,u_m)\in R_\ell^m,\ \ell=1,\ldots,q,\xdef\featuredefmulti{\lab}\eqnum\show{qfortyss}$$
and we construct a hard aggregation scheme with disaggregation sets for each  $m=1,\ldots,N-1$, given by
$$I_\ell^m=\big\{(u_1,\ldots,u_m)\mid V(u_1,\ldots,u_m)\in R_\ell^m\big\},\qquad \ell=1,\ldots,q.$$
Note that
the number of aggregate states is roughly similar for each of the $N-1$ stages.  By contrast the number of states of the original problem may increase very fast (exponentially) as $N$ increases; cf.\ Fig.\ \figdiscropt.  

The aggregation scheme is illustrated in Fig.\ \figdiscroptagr. It involves $N-1$ successive transitions between  $m$-solutions to $(m+1)$-solutions ($m=1,\ldots,N-1$), interleaved with transitions to and from the corresponding aggregate states. The aggregate problem is completely defined once the aggregate states and the disaggregation probabilities have been chosen. The transition mechanism of stage $m$ involves the following steps. 
\nitem{(1)} From an aggregate state $\ell$ at stage $m$, we generate some state $(u_1,\ldots,u_m)\in I_\ell^m$ according to the disaggregation probabilities.
\nitem{(2)} We transition to the next state $(u_1,\ldots,u_m,u_{m+1})$ by selecting the control $u_{m+1}$.
\nitem{(3)} We run the $s$ heuristics from the $(m+1)$-solution $(u_1,\ldots,u_m,u_{m+1})$ to determine the next aggregate state, which is the index of the set of the partition of stage $m+1$ to which the vector
$$V(u_1,\ldots,u_m,u_{m+1})=\big(V_1(u_1,\ldots,u_m,u_{m+1}),\ldots,V_s(u_1,\ldots,u_m,u_{m+1})\big)$$
belongs.
\smskip
\pn A key issue is the selection of the disaggregation probabilities for each stage. This requires, for each value of $m$, the construction of a suitable sample of $m$-solutions, where the disaggregation sets $I_\ell^m$ are adequately represented.

\topinsert
\centerline{\includegraphics[width=6.0in]{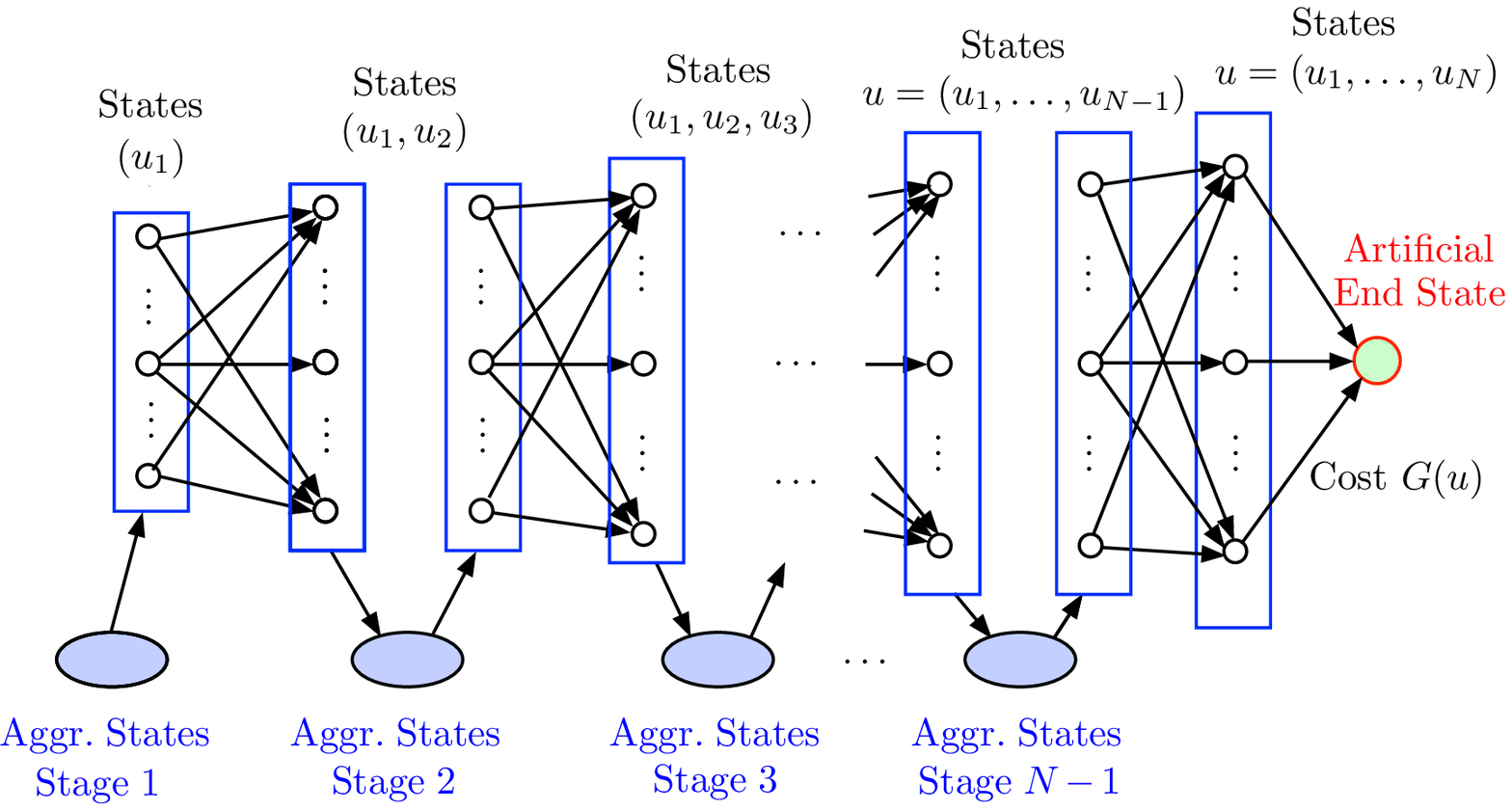}}
\vskip-0.3pc
\hskip-3pc\fig{0pc}{\figdiscroptagr.} {Schematic illustration of the heuristics-based aggregation scheme for  discrete optimization. The aggregate states are defined by the scoring functions/heuristics, and the optimal aggregate costs are obtained by DP stating from the last stage and proceeding backwards.}\endinsert

The solution of the aggregate problem by DP starts at the last stage to compute the corresponding aggregate costs $r^*_{\ell (N-1)}$ for each of the aggregate states $\ell$, using $G(u)$ as terminal cost function. Then it proceeds with the next-to-last stage to compute the corresponding aggregate costs  $r^*_{\ell (N-2)}$, using  the previously computed aggregate costs $r^*_{\ell (N-1)}$, etc.

The optimal cost function $\jstar(u_1,\ldots,u_m)$ for stage $m$ is approximated by a piecewise constant function, which is derived by solving the aggregate problem. This is the function 
$$\tl J(u_1,\ldots,u_m)=r^*_{\ell m}, \qquad\forall\ (u_1,\ldots,u_m)\hbox{ with }V(u_1,\ldots,u_m)\in R_\ell^m,\xdef\aggrcostapprox{\lab}\eqnum\show{qfortyss}$$
where   $r^*_{\ell m}$ is the optimal cost of aggregate state $\ell$ at stage $m$ of the aggregate problem. 

\xdef\figdiscroptpolicy{\figr}\figrnum\show{myfigure}

\topinsert
\centerline{\includegraphics[width=6.0in]{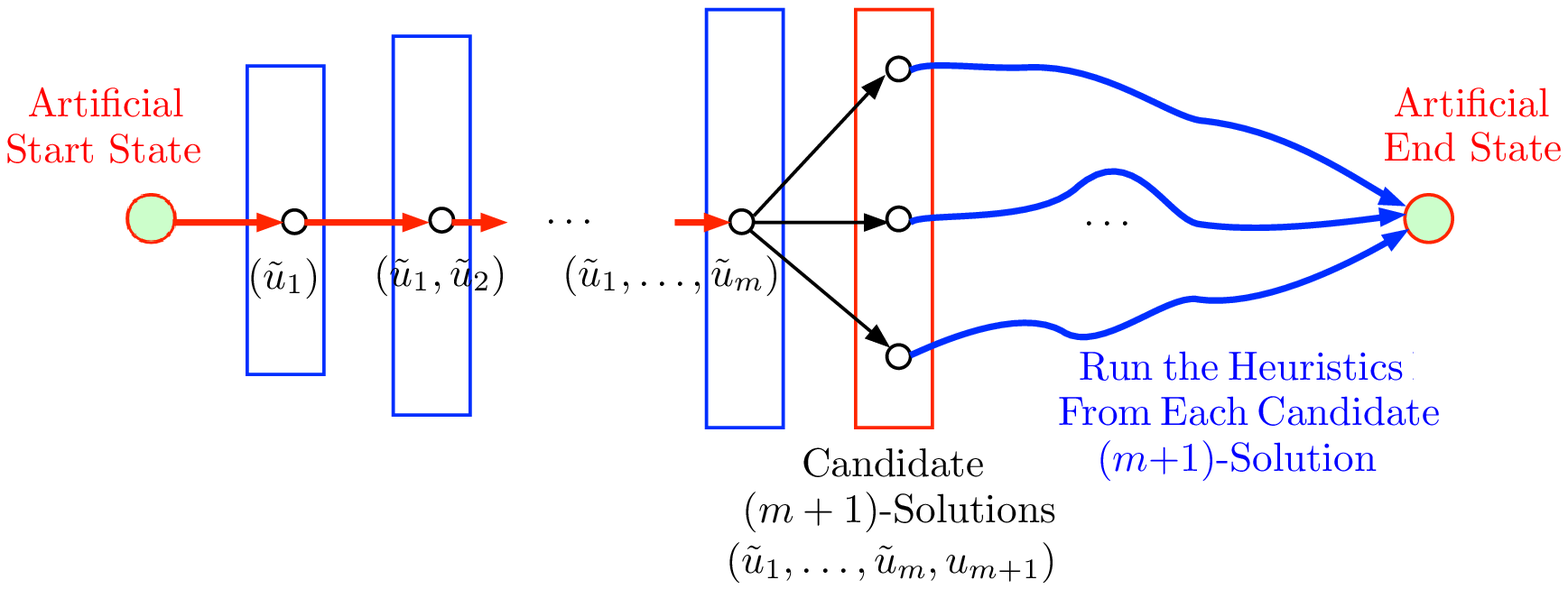}}
\vskip-0pc
\hskip-3pc\fig{0pc}{\figdiscroptpolicy.} {Sequential construction of a  suboptimal $N$-solution $(\tl u_1,\ldots,\tl u_N)$ for the original problem, after the aggregate problem has been solved. Given the $m$-solution $(\tl u_1,\ldots,\tl u_{m})$, we run the $s$ heuristics from each of the candidate $(m+1)$-solution $(\tl u_1,\ldots,\tl u_{m},u_{m+1})$, and compute the aggregate state and aggregate cost of this candidate $(m+1)$-solution. We then select as $\tl u_{m+1}$ the one that corresponds to the candidate $(m+1)$-solution with minimal aggregate cost.}\endinsert

Once the aggregate problem has been solved for the costs $r^*_{\ell m}$, a suboptimal $N$-solution $(\tl u_1,\ldots,\tl u_N)$ for the original problem is obtained sequentially, starting from stage 1 and proceeding to stage $N$, through the minimizations
$$\tl u_1\in\arg\min_{u_1}\tl J(u_1),\xdef\heuristicmino{\lab}\eqnum\show{qfortyss}$$
$$\tl u_{m+1}\in\arg\min_{u_{m+1}\in
U_{m+1}(\tl u_1,\ldots,\tl u_{m})}\tl J(\tl u_1,\ldots,\tl u_{m},u_{m+1}),\quad
m=1,\ldots,N-1.\xdef\heuristicmin{\lab}\eqnum\show{qfortyss}$$

 Note that to evaluate each of the costs $\tl J(\tl u_1,\ldots,\tl u_{m},u_{m+1})$ needed for this minimization, we need to do the following (see Fig.\ \figdiscroptpolicy):
\nitem{(1)} Run the $s$ heuristics from the $(m+1)$-solution $(\tl u_1,\ldots,\tl u_{m},u_{m+1})$ to evaluate the scoring vector of heuristic costs 
$$V(\tl u_1,\ldots,\tl u_m,u_{m+1})=\big(V_1(\tl u_1,\ldots,\tl u_m,u_{m+1}),\ldots,V_s(\tl u_1,\ldots,\tl u_m,u_{m+1})\big).$$
\nitem{(2)} Set $\tl J(\tl u_1,\ldots,\tl u_{m},u_{m+1})$ to the aggregate cost $r^*_{\ell (m+1)}$ of the aggregate state $S_{\ell (m+1)}$ corresponding to this scoring vector, i.e., to the set $R_\ell^{(m+1)}$ such that
$$V(\tl u_1,\ldots,\tl u_m,u_{m+1})\in R_\ell^{(m+1)}.$$
\smskip
\pn Once $\tl J(\tl u_1,\ldots,\tl u_{m},u_{m+1})$ has been computed for all  ${u_{m+1}\in
U_{m+1}(\tl u_1,\ldots,\tl u_{m})}$, we select $\tl u_{m+1}$ via the minimization  \heuristicmin, and repeat starting from the $(m+1)$-solution $(\tl u_1,\ldots,\tl u_{m},\tl u_{m+1})$. Note that even if there is only one heuristic, $\tl u_{m+1}$ minimizes the aggregate cost $r^*_{\ell (m+1)}$, which is not the same as the cost corresponding to the heuristic.

We finally mention a simple improvement of the scheme just described for constructing an $N$-solution. In the course of the algorithm many other $N$-solutions are obtained, during the training and final solution selection processes. It is possible that some of these solutions are actually better [have lower cost $G(u)$] than the final $N$-solution $(\tl u_1,\ldots,\tl u_N)$ that is constructed by using the aggregate problem formulation. This can happen because the aggregation scheme is subject to quantization error. Thus it makes sense to maintain the best of the $N$-solutions generated in the course of the algorithm, and compare it at the end with the $N$-solution obtained through the aggregation scheme.  This is similar to the so-called ``fortified"  version of the rollout algorithm (see [BTW97] or [Ber17]).

\subsubsection{Relation to the Rollout Algorithm}

\pn 
The idea of using one or more heuristic algorithms as a starting point for generating an improved solution of a discrete optimization problem is shared by other suboptimal control approaches. A prime example is the rollout algorithm, which in some contexts can be viewed as a single policy iteration; see [BTW97] for an analysis of  rollout  for discrete optimization problems, and the textbook [Ber17] for an extensive discussion and many references to applications, including the important model predictive control methodology for control system design. 

Basically the rollout algorithm uses the scheme of Fig.\ \figdiscroptpolicy\ to construct a suboptimal solution $(\tl u_1,\ldots,\tl u_{N})$ in $N$ steps, one component at a time, but adds a new decision $\tl u_{m+1}$ to the current $m$-solution  $(\tl u_1,\ldots,\tl u_{m})$ in a simpler way. It runs the $s$ heuristics from each candidate $(m+1)$-solution $(\tl u_1,\ldots,\tl u_{m},u_{m+1})$ and computes the corresponding heuristic costs 
$$V_1(\tl u_1,\ldots,\tl u_m,u_{m+1}),\ldots,V_s(\tl u_1,\ldots,\tl u_m,u_{m+1}).$$
It then selects as the next decision $\tl u_{m+1}$ the one that minimizes over 
${u_{m+1}\in
U_{m+1}(\tl u_1,\ldots,\tl u_{m})}$
the best heuristic cost
$$\hat V(\tl u_1,\ldots,\tl u_m,u_{m+1})=\min\big\{V_1(\tl u_1,\ldots,\tl u_m,u_{m+1}),\ldots,V_s(\tl u_1,\ldots,\tl u_m,u_{m+1})\big\},$$
i.e., it uses $\hat V$ in place of $\tl J$ in Eqs.\ \heuristicmino-\heuristicmin. In practice, the rollout algorithm's heuristics may involve sophisticated suboptimizations that may make sense in the context of the problem at hand.

Note that the construction of the final $N$-solution is similar and equally complicated in the rollout and the scoring vector-based aggregation approach. However, the aggregation approach requires an extra layer of computation {\it prior to constructing the $N$-solution\/}, namely the solution of an aggregate problem. This may be a formidable problem, because it is stochastic (due to the use of disaggregation probabilities) and  must be solved exactly (at least in principle). Still, the number of  states of the aggregate problem may be quite reasonable, and its solution is well suited for parallel computation. 

On the other hand, setting aside the issue of computational solution of the aggregate problem, the heuristics-based aggregation algorithm has the potential of being far superior to the rollout algorithm, for the same reason that approximate policy improvement based on aggregation can be far superior to policy improvement based on one-step lookahead. In particular, with sufficiently large number of aggregate states to eliminate the effects of the quantization error, feature-based aggregation will find an optimal solution, regardless of the quality of the heuristics used. By contrast, policy iteration and rollout can only aspire to produce a solution that is better than the one produced by the heuristics.

\subsubsection{Using Multistep Lookahead and Monte Carlo Tree Search}

\pn Once the aggregate problem that is based on multiple scoring functions has been solved, the final $N$-solution can be constructed in more sophisticated ways than the one described in Fig.\ \figdiscroptpolicy.  It can be seen that the scheme of Eqs.\ \heuristicmino-\heuristicmin\ and Fig.\ \figdiscroptpolicy\ is based on one-step lookahead. It is possible instead to use multistep lookahead or randomized versions such as Monte Carlo tree search.

\xdef\figdiscroptmultistep{\figr}\figrnum\show{myfigure}

\topinsert
\centerline{\includegraphics[width=6in]{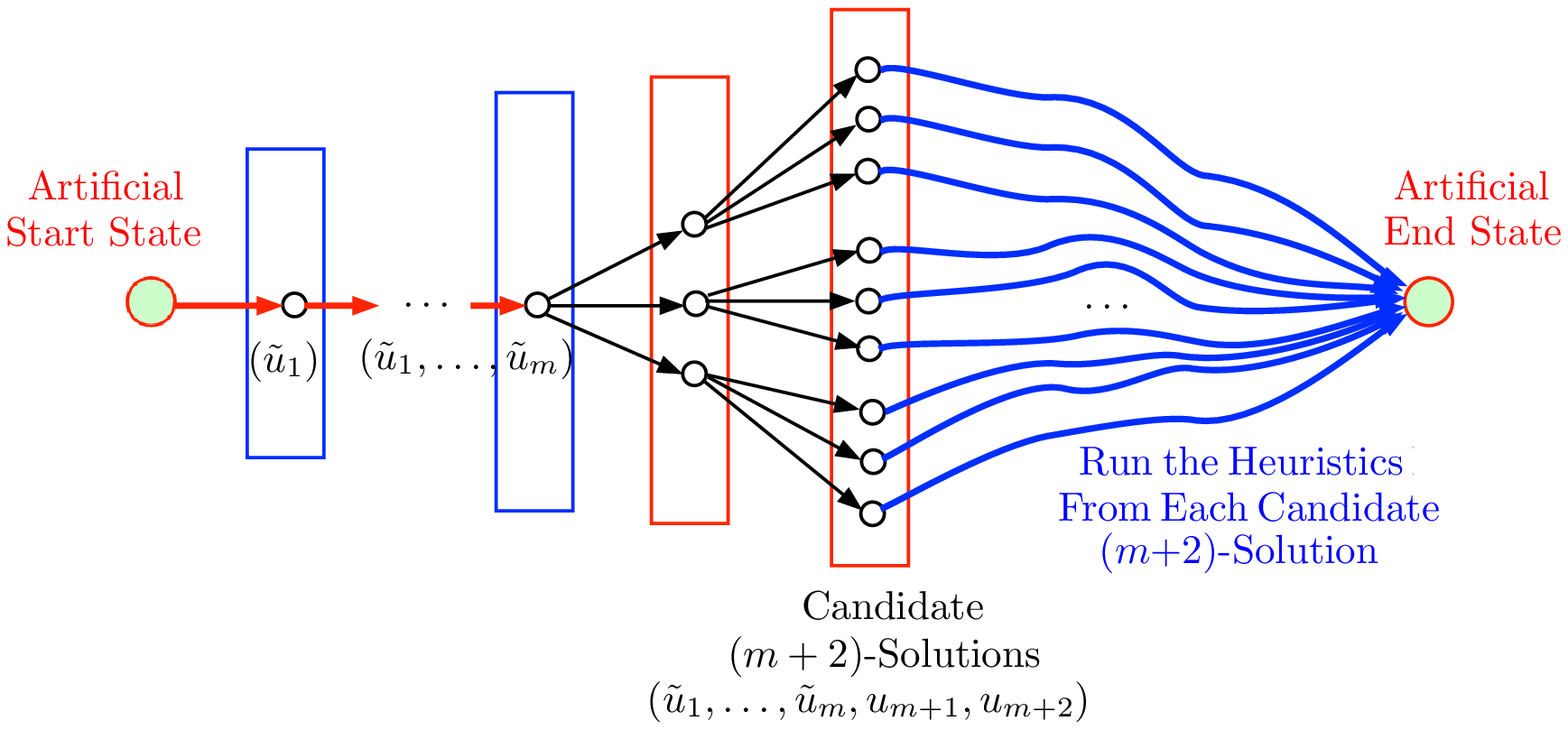}}
\vskip-1pc
\hskip-3pc\fig{0pc}{\figdiscroptmultistep.} {Sequential construction of a  suboptimal $N$-solution $(\tl u_1,\ldots,\tl u_N)$ by using two-step lookahead, after the aggregate problem has been solved. Given the $m$-solution $(\tl u_1,\ldots,\tl u_{m})$, we run the $s$ heuristics from all the candidate $(m+2)$-solutions $(\tl u_1,\ldots,\tl u_{m},u_{m+1},u_{m+2})$, and select as $\tl u_{m+1}$ the first component of the two-step sequence that corresponds to minimal aggregate cost.}\endinsert

As an example, in a two-step lookahead scheme, we again obtain a suboptimal solution $(\tl u_1,\ldots,\tl u_N)$ for the original problem in $N$ stages, starting from stage 1 and proceeding to stage $N$. At stage 1, we carry out the two-step minimization
$$(\tl u_1,\tl u_2)\in\arg\min_{u_1,u_2}\tl J(u_1,u_2),\xdef\heuristicminomul{\lab}\eqnum\show{qfortyss}$$
and fix the first component $\tl u_1$ of the result, cf.\ Fig.\ \figdiscroptmultistep. We then proceed sequentially: for $m=1,\ldots,N-2$, given the current $m$-solution $(\tl u_1,\ldots,\tl u_{m})$, we carry out the two-step minimization
$$(\tl u_{m+1},\tl u_{m+2})\in\arg\min_{u_{m+1},u_{m+2}}\tl J(\tl u_1,\ldots,\tl u_{m},u_{m+1},u_{m+2}),\quad
m=1,\ldots,N-2,\xdef\heuristicminmul{\lab}\eqnum\show{qfortyss}$$
and fix the first component $\tl u_{m+1}$ of the result, cf.\ Fig.\ \figdiscroptmultistep. At the final stage, given the $(N-1)$-solution $(\tl u_1,\ldots,\tl u_{N-1})$, we carry out the one-step minimization
$$\tl u_{N}\in\arg\min_{u_N}\tl J(\tl u_1,\ldots,\tl u_{N-1},u_{N}),\xdef\heuristicminlast{\lab}\eqnum\show{qfortyss}$$
and obtain the final $N$-solution $(\tl u_1,\ldots,\tl u_{N})$.

Multistep lookahead generates a tree of fixed depth that is rooted at the last node $\tl u_m$ of the current $m$-solution, and then runs the heuristics from each of the leaf nodes of the tree. We can instead select only a subset of these leaf nodes from which to run the heuristics, thereby economizing on computation. The selection may be based on some heuristic criterion.
 Monte Carlo tree search similarly uses multistep lookahead but selects only a random sample of leaf nodes to search based on some criterion. 
 
 In a more general version of Monte Carlo tree search, instead of a single partial solution, we maintain multiple partial solutions, possibly of varying length. At each step, a one-step or multistep lookahead tree is generated from the most ``promising" of the current partial solutions, selected by using a randomization mechanism.  The heuristics are run from the leafs of the lookahead trees similar to Fig.\ \figdiscroptmultistep. Then some of the current partial solutions are expanded with an additional component based on the results produced by the heuristics. This type of Monte Carlo tree search has been suggested for use in conjunction with rollout (see the paper [RSS12]), and it can be similarly used with feature-based aggregation.

\subsection{Stochastic Shortest Path Problems - Illustrative Examples}

\pn Our aggregation framework extends straightforwardly to stochastic shortest path (SSP for short) problems, where there is no discounting and in addition to the states $1,\ldots,n$, there is an additional cost-free and absorbing termination state, denoted $0$ (the text references given earlier discuss in detail such problems). The principal change needed is to account for the termination state by introducing an additional aggregate state with corresponding disaggregation set $\{0\}$. As before there are also other aggregate states $S_1,\ldots,S_q$ whose disaggregation sets $I_1,\ldots,I_q$ are subsets of $\{1,\ldots,n\}$. With this special handling of the termination state, the aggregate problem becomes a standard  SSP problem whose termination state is the aggregate state corresponding to $0$. 
The Bellman equation of the aggregate problem is given by
$$r_\ell=\sum_{i=1}^nd_{\ell i}\min_{u\in U(i)}\sum_{j=1}^n p_{ij}(u)\lf(g(i,u,j)+\sum_{m=1}^q\phi_{jm}\,r_{m}\ri),\qquad \ell=1,\ldots,q,\xdef\aggrmapssp{\lab}\eqnum\show{lsmin}$$
[cf.\ Eq.\ \aggrmap]. It has a unique solution under some well-known conditions that date to the paper by Bertsekas and Tsitsiklis [BeT91] (there exists at least one proper policy, i.e., a stationary policy that guarantees eventual termination from each initial state with probability 1; moreover all stationary policies that are not proper have infinite cost starting from some initial state). In particular, these conditions are satisfied if all stationary policies are proper.

We will now provide two simple illustrative SSP examples, which were presented in the author's paper [Ber95] as instances of poor performance of TD($\l$) and other methods that are based on projected equations and temporal differences (see also the book [BeT96], Section 6.3.2).
In these examples the cost function of a policy will be approximated by using feature-based aggregation and a scoring function obtained using either the TD(1) or the TD(0) algorithms. The approximate cost function computed by aggregation will be compared with the results of TD(1) and TD(0). We will show that aggregation provides a much better approximation, suggesting better policy improvement results in a PI context.

Our examples involve a problem with a single policy $\m$ where the corresponding Markov chain is deterministic with $n$ states plus a termination state $0$. Under $\m$, when at state $i=1,\ldots,n$, we move to state $i-1$ at a cost $g_i$. Thus starting at state $i$ we traverse each of the states $i-1,\ldots,1$ and terminate at state 0 at costs $g_i,g_{i-1},\ldots,g_1$, respectively, while accumulating the total cost
$$J_\m(i)=g_i+\cdots+g_1,\qquad i=1,\ldots,n,$$
with $J_\m(0)=0$. We consider a linear approximation to this cost function, which we denote by $V$:
$$V(i)=ri,\qquad i=1,\ldots,n,$$
where $r$ is a scalar parameter. This parameter may be obtained by using any of the  simulation-based methods that are available for training linear architectures, including TD($\l$). In the subsequent discussion we will assume that TD($\l$) is applied in an idealized form where the simulation samples contain no noise.

The TD(1) algorithm is based on minimizing the sum of the squares of the differences between $J_\m$ and $V$ over all states, yielding the approximation
$$\hat V_1(i)=\hat r_1 i,\qquad i=0,1,\ldots,n,$$
where 
$$\hat r_1\in\arg\min_{r\in\re}\sum_{i=1}^n\big(J_\m(i)-ri\big)^2.\xdef\leastsqone{\lab}\eqnum\show{lsmin}$$
Here, consistent with our idealized setting of noise-free simulation, we assume that $J_\m(i)$ is computed exactly for all $i$. 
The TD(0) algorithm is based on minimizing the sum of the squares of the errors in satisfying the Bellman equation $V(i)=g_i+V(i-1)$ (or temporal differences) over all states, yielding the approximation
$$\hat V_0(i)=\hat r_0 i,\qquad i=0,1,\ldots,n,$$
where 
$$\hat r_0=\in\arg\min_{r\in\re}\sum_{i=1}^n\big(g_i+r(i-1)-ri\big)^2.\xdef\leastsqtwo{\lab}\eqnum\show{lsmin}$$
Again, we assume that the temporal differences $\big(g_i+r(i-1)-ri\big)$ are computed exactly for all $i$.

The straightforward solution of the minimization problems in Eqs.\ \leastsqone\ and \leastsqtwo\ yields
$$\hat r_1={n(g_1+\cdots+g_n)+(n-1)(g_1+\cdots+g_{n-1})+\cdots+g_1\over n^2+(n-1)^2+\cdots+1},$$
and
$$\hat r_0={n g_n+(n-1) g_{n-1}+\cdots+g_1\over n+(n-1)+\cdots+1}.$$

\xdef\plotscasea{\figr}\figrnum\show{myfigure}
\xdef\plotscaseb{\figr}\figrnum\show{myfigure}

Consider now two different choices of the one-stage costs $g_i$:
\nitem{(a)} $g_1=1$, and $g_i=0$ for all $i\ne 1$. 
\nitem{(b)} $g_n=-(n-1)$, and $g_i=1$ for all $i\ne n$.
\smskip
\pn Figures \plotscasea\ and \plotscaseb\ provide plots of $J_\m(i)$, and the approximations $\hat V_1(i)$ and $\hat V_0(i)$ for these two cases (these plots come from [Ber95] where the number of states used was $n=50$). It can be seen that $\hat V_1(i)$ and particularly $\hat V_0(i)$ are poor approximations of $J_\m(i)$, suggesting that if used for policy improvement, they may yield a poor successor policy.

\topinsert
\centerline{\includegraphics[width=4.9in]{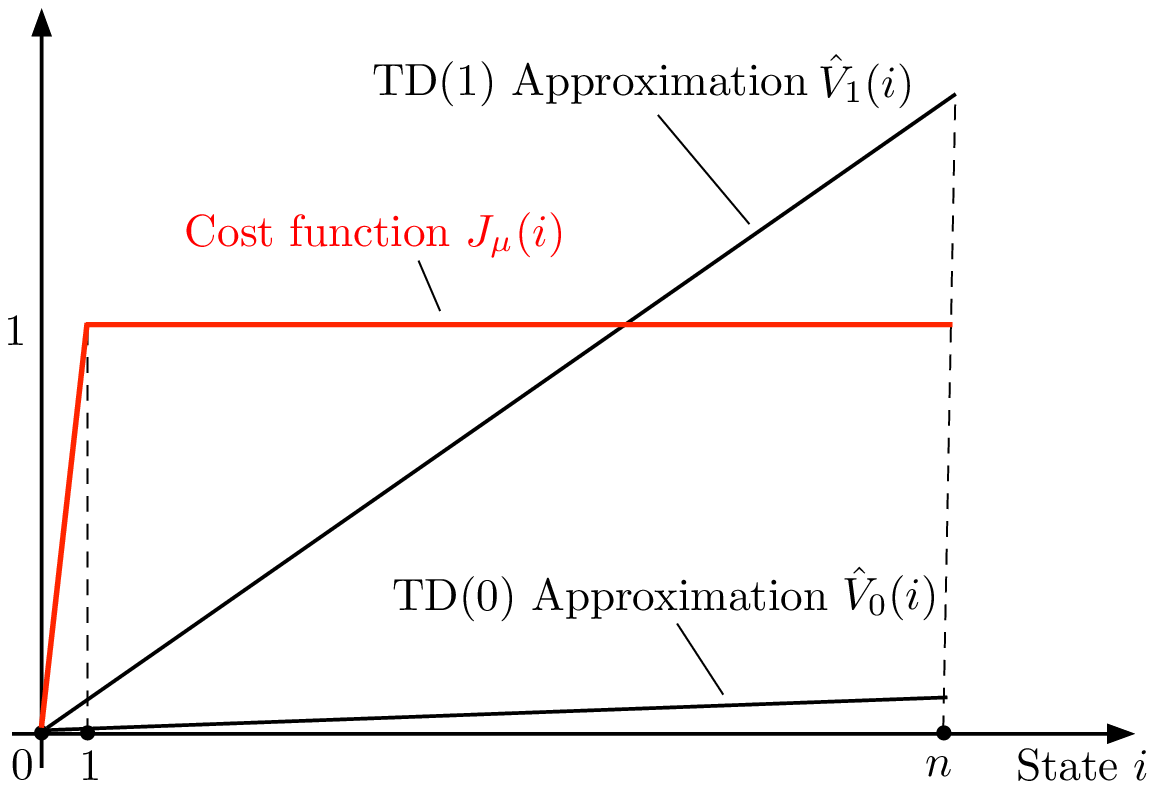}}
\vskip-1.0pc
\hskip-3pc\fig{0pc}{\plotscasea.} {Form of $J_\m(i)$ and the linear approximations $\hat V_1(i)$ and $\hat V_0(i)$ for case (a): $g_1=1$, and $g_i=0$ for all $i=2,\ldots,n$.}\endinsert

\midinsert
\centerline{\includegraphics[width=4.2in]{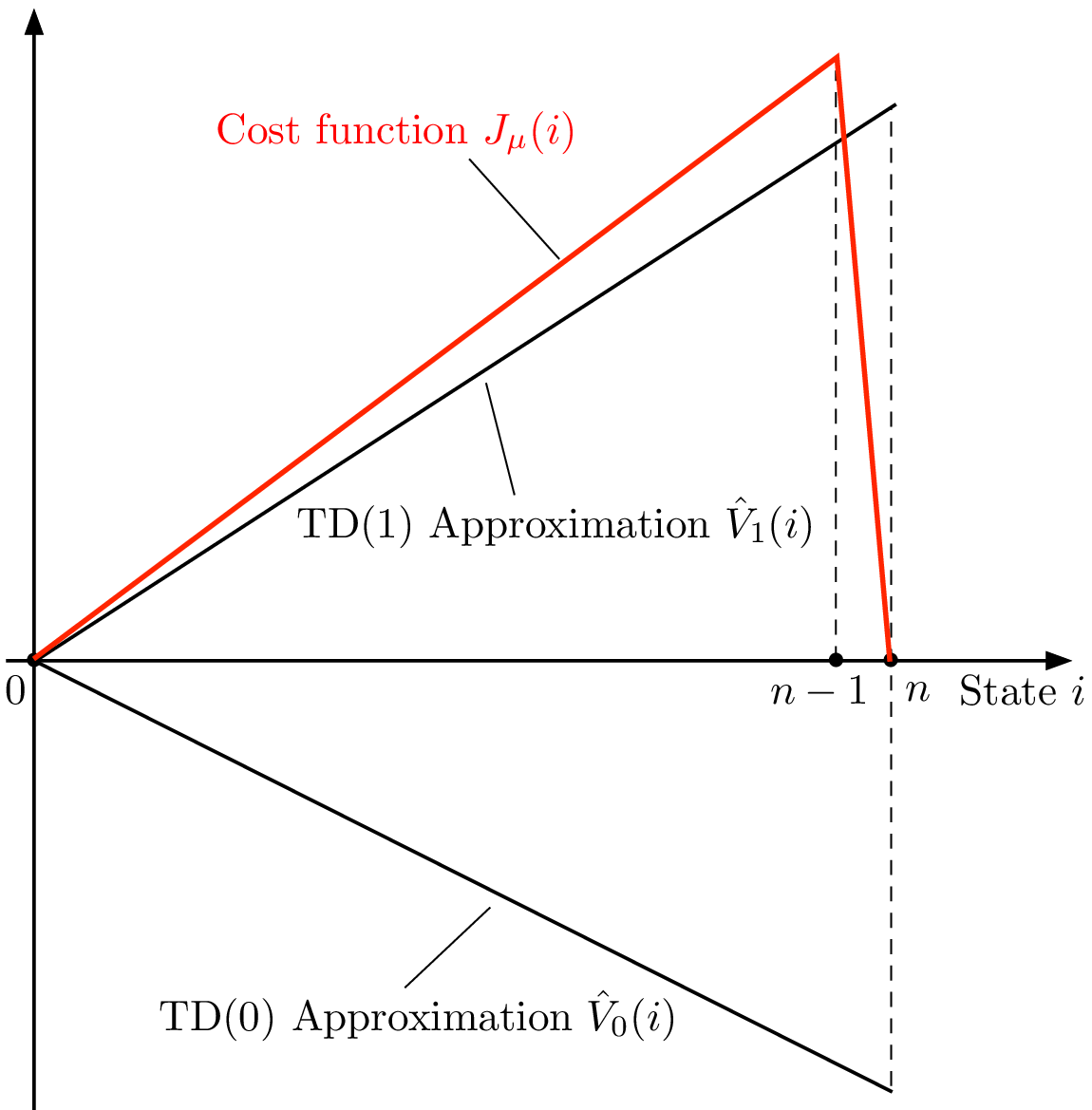}}
\vskip-1.0pc
\hskip-3pc\fig{0pc}{\plotscaseb.} {Form of $J_\m(i)$ and the linear approximations $\hat V_1(i)$ and $\hat V_0(i)$ for case (b): $g_n=-(1-n)$, and $g_i=1$ for all $i=1,\ldots,n-1$.}\endinsert

We will now consider a hard aggregation scheme based on using $\hat V_1$ and $\hat V_0$ as scoring functions. The aggregate states of such a scheme in effect consist of disaggregation subsets $I_1,\ldots,I_q$ with $\cup_{\ell=1}^q I_\ell=\{1,\ldots,n\}$ plus the subset $\{0\}$ that serves as the termination state of the aggregate problem. With either $\hat V_1$ or $\hat V_0$ as the scoring function, the subsets $I_1,\ldots,I_q$ consist of contiguous states. In order to guarantee that the termination state is eventually reached in the aggregate problem, we assume that the disaggregation probability of the smallest state within each of the subsets $I_1,\ldots,I_q$ is strictly positive; this is a mild restriction, which is naturally satisfied in typical schemes that assign equal probability to all the states in a disaggregation set. 

Consider first case (a) (cf.\ Fig.\ \plotscasea). Then, because the policy cost function $J_\m$ is constant within each of the subsets $I_1,\ldots,I_q$, the scalar $\e$ in Prop.\ \properrorbd\ is equal to 0, implying that the hard aggregation scheme yields the optimal cost function, i.e., $r^*_\ell=J_\m (i)$ for all $i\in I_\ell$. To summarize, in case (a) the TD(0) approach yields a very poor linear cost function approximation, the TD(1) approach yields a poor linear cost function approximation, but the aggregation scheme yields the nonlinear policy cost function $J_\m$ exactly.

\xdef\piecelin{\figr}\figrnum\show{myfigure}

Consider next case (b) (cf.\ Fig.\ \plotscaseb). Then, the hard aggregation scheme yields a piecewise constant approximation to the optimal cost function. The quality of the approximation is degraded by quantization effects. In particular, as the variations of $J_\m$, and $\hat V_1$ or $\hat V_0$ increase over the disaggregation sets $I_1,\ldots,I_q$, the quality of the approximation deteriorates, as predicted by Prop.\ \propexactapprox. Similarly, as the number of states  in the disaggregation sets $I_1,\ldots,I_q$ is reduced, the quality of the approximation improves, as illustrated in Fig.\ \piecelin. In the extreme case where there is only one state in each of the disaggregation sets, the aggregation scheme yields  exactly $J_\m$.

To summarize, in case (b) the TD(0) approach yields a very poor linear cost function approximation, the TD(1) approach yields a reasonably good linear cost function approximation, while the aggregation scheme yields a piecewise constant approximation whose quality depends on the coarseness of the quantization that is implicit in the selection of the number $q$ of disaggregation subsets. 
The example of case (b) also illustrates how the quality of the scoring function affects the quality of the approximation provided by the aggregation scheme. Here both $\hat V_1$ and $\hat V_0$ work well as scoring functions, despite their very different form, because states with similar values of $J_\m$ also have similar values of $\hat V_1$ as well as $\hat V_0$ (cf.\ Prop.\ \propexactapprox).

\topinsert
\centerline{\includegraphics[width=4.7in]{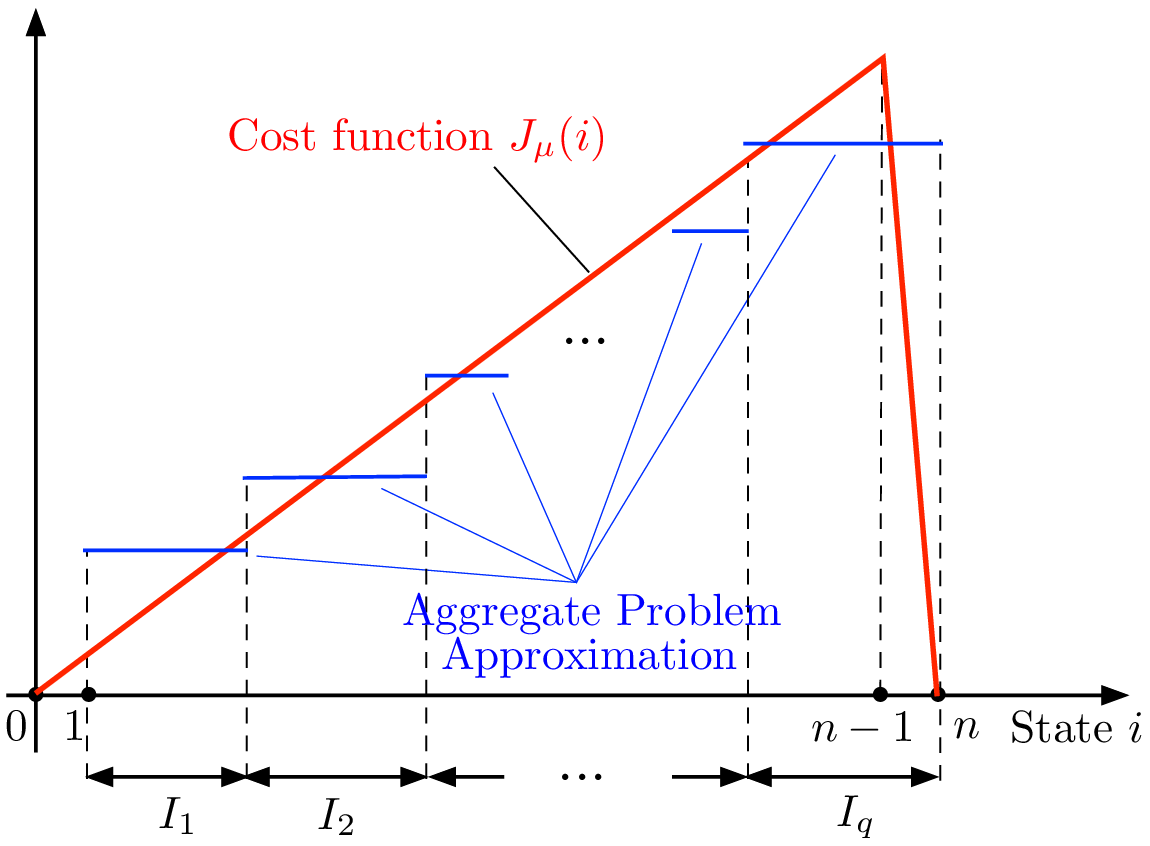}}
\vskip-1.5pc
\hskip-3pc\fig{0pc}{\piecelin.} {Schematic illustration of the piecewise constant approximation of $J_\m$ that is provided by hard aggregation based on the scoring functions $\hat V_1$ and $\hat V_0$ in case (b).}\endinsert

\subsection{Multistep Aggregation}

\xdef\figsftm{\figr}\figrnum\show{myfigure}
\vskip-1pc

\pn The aggregation methodology discussed so far is based on the aggregate problem Markov chain of Fig.\ \figsfo, which returns to an aggregate state after a single transition of the original chain. We may obtain alternative aggregation frameworks by considering a different Markov chain, which starting from an aggregate state, involves multiple original system state transitions before return to an aggregate state. We discuss two possibilities:

\nitem{(a)} {\it $k$-Step Aggregation\/}: Here we require a fixed number $k$ of transitions between original system states before returning to an aggregate state.
\nitem{(b)} {\it $\l$-Aggregation\/}: Here the number $k$ of transitions prior to returning to an aggregate state is controlled by some randomization mechanism. In the case where $k$ is geometrically distributed with parameter $\l\in(0,1)$, this method involves multistep mappings that arise in temporal difference contexts and facilitate the use of temporal differences methodology.
\smskip

Another related possibility, which we do not discuss in this paper, is to introduce {\it temporal abstractions} (faster/multistep ``macro-actions" and transitions between selected states with suitably computed transition costs) into the upper (original system) portion of the aggregate problem Markov chain of Fig.\ \figsfo. There have been many proposals of this type in the reinforcement learning literature, under various names; for some representative works, see Hauskrecht et al.\ [HMK98],  Sutton, Precup, and Singh [SPS99], Parr and Russell [PaR98], Dietterich [Die00], Konidaris and Barto [KoB09], Ciosek and Silver [CiS15], Mann, Mannor, and Precup [MMP15], Serban et al.\ [SSP18], and the references cited there. It is likely that some of these proposals can be fruitfully adapted to our feature-based aggregation context, and this is an interesting subject for further research.

\subsubsection{$k$-Step Aggregation}

\pn This scheme, suggested in [Ber11a] and illustrated in Fig.\ \figsftm, is specified by disaggregation and aggregation probabilities as before, but involves $k>1$ transitions between original system states in between transitions from and to aggregate states.
The aggregate DP problem for this scheme involves $k+1$ copies of the original state space, in addition to the aggregate states. We accordingly introduce  vectors $\tl J_0,\tl J_1,\ldots,\tl J_k$, and $r^*=\{r^*_1,\ldots,r^*_q\}$ where:
\smskip
\nitem{} $r^*_\ell$ is the optimal cost-to-go from aggregate state $S_\ell$.
\nitem{} $\tl J_0(i)$ is the optimal cost-to-go from original system state $i$ that has just been generated from an aggregate state (left side of Fig.\ \figsftm).
\nitem{} $\tl J_1(j_1)$ is the optimal cost-to-go from original system state $j_1$ that has just been generated from an original system state $i$.
\nitem{} $\tl J_m(j_m)$, $m=2,\ldots,k$, is the optimal cost-to-go from original system state $j_m$ that has just been generated from an original system state $j_{m-1}$.
\smskip
\pn These vectors satisfy the following set of Bellman equations:
$$r^*_\ell=\sum_{i=1}^nd_{\ell i}\tl J_0(i),\qquad \ell=1,\ldots,q,$$
$$\tl J_0(i)=\min_{u\in U(i)}\sum_{j_1=1}^n p_{ij_1}(u)\big(g(i,u,j_1)+\a \tl J_1(j_1)\big),\qquad i=1,\ldots,n,\xdef\multibo{\lab}\eqnum\show{lsmin}$$
$$\eqalign{\tl J_m(j_m)=\min_{u\in U(j_m)}\sum_{j_{m+1}=1}^n p_{j_mj_{m+1}}&(u)\big(g(j_m,u,j_{m+1})+\a \tl J_{m+1}(j_{m+1})\big),\cr
&\ \ \ \ \ \ \ \ \ \ \ \ \ j_m=1,\ldots,n,\ m=1,\ldots,k-1,\cr}\xdef\multibt{\lab}\eqnum\show{lsmin}$$
$$\tl J_k(j_k)=\sum_{\ell=1}^q\phi_{j_k\ell}r^*_\ell,\qquad j_k=1,\ldots,n.\xdef\multibth{\lab}\eqnum\show{lsmin}$$
By combining these equations, we obtain an equation for $r^*$:
$$r^*=DT^k(\Phi r^*),$$
where $T$ is the usual DP mapping of the original problem [the case $k=1$ corresponds to Eqs.\ \rstfixed-\aggrmap]. As earlier, it can be seen that the associated mapping $DT^k\Phi$ is a contraction mapping with respect to the sup-norm, but its contraction modulus is $\a^k$ rather than $\a$.

\topinsert
\centerline{\hskip0pc\includegraphics[width=6.4in]{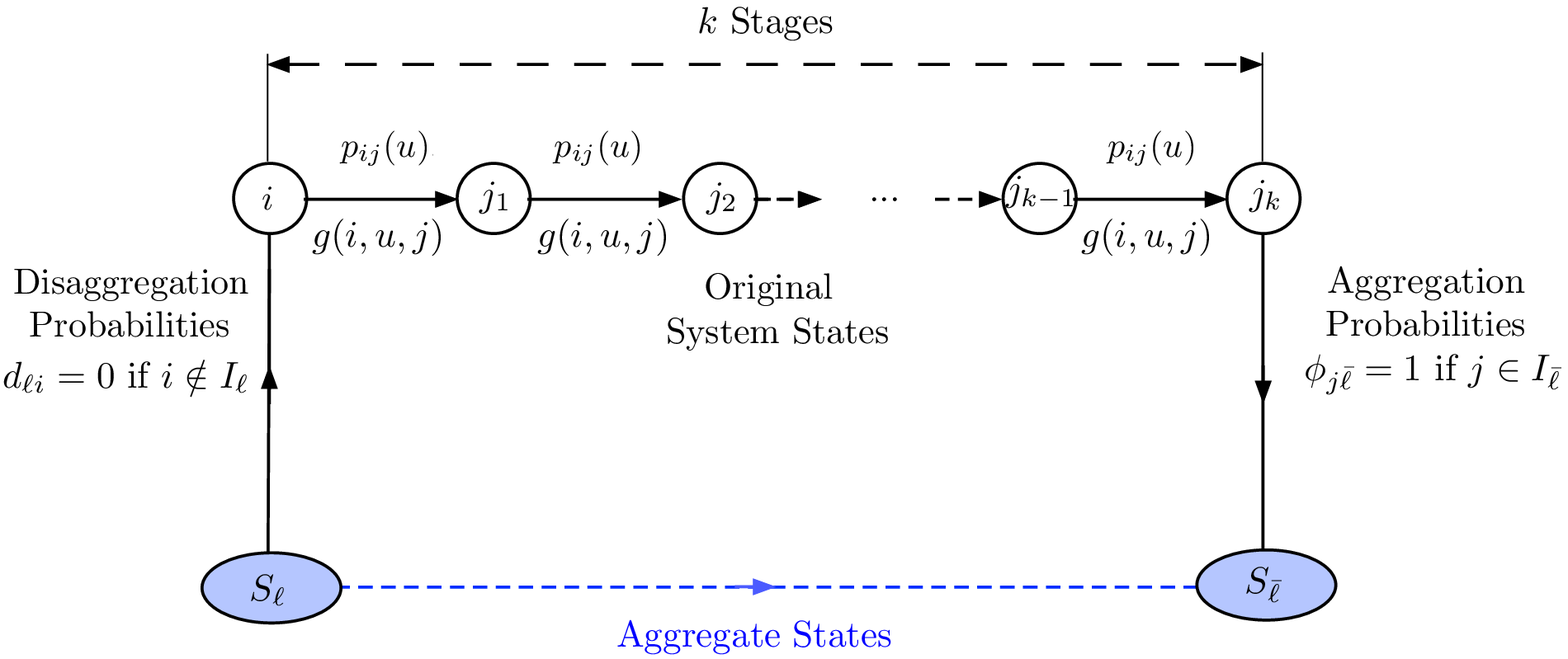}}
\vskip-0pc
\hskip-3pc\fig{0pc}{\figsftm}{The transition mechanism for multistep aggregation. It is based on a dynamical system involving $k$ transitions between original system states interleaved between transitions from and to aggregate states.}\endinsert

There is a similar mapping corresponding to a fixed policy and it can be used to implement a PI algorithm, which evaluates a policy through calculation of a corresponding parameter vector $r$ and then improves it. However, there is a major difference from the single-step aggregation case: a policy involves a set of $k$ control functions $\{\m_0,\ldots,\m_{k-1}\}$, and while a known policy can be easily simulated, its improvement involves multistep lookahead using the minimizations of Eqs.\ \multibo-\multibth, and may be costly. Thus the preceding implementation of multistep aggregation-based PI is a useful idea only for problems where the cost of this multistep lookahead minimization (for a single given starting state) is not prohibitive.

On the other hand, from a theoretical point of view, a multistep aggregation scheme provides a means of better approximation of the true optimal cost vector $\jstar $, independent of the use of a large number of aggregate states. This can be seen from Eqs.\ \multibo-\multibth, which by classical value iteration convergence results, show that $\tl J_0(i)\to \jstar (i)$ as $k\to\infty$, regardless of the choice of aggregate states. Moreover, because the modulus of the underlying contraction is $\a^k$,  we can  verify an improved error bound in place of the bound \aggrbound\ of Prop.\ \properrorbd, which corresponds to $k=1$:
$$\big|\jstar(i)-r^*_\ell\big|\le {\e\over 1-\a^k},\qquad  \forall\ i \hbox{ such that } i\in I_\ell,\ \ell=1,\ldots,q,$$
where $\e$ is given by Eq.\ \epsdef. The proof is very similar to the one of Prop.\ \properrorbd.

\subsubsection{$\l$-Aggregation}

\pn Multistep aggregation need not involve sequences of a fixed number of transitions between original system states. The number of transitions may be state-dependent or may be controlled by some randomized mechanism. 
In one such possibility, called {\it $\l$-aggregation\/}, we introduce a parameter $\l\in(0,1)$ and consider a Markov chain that makes a transition with probability $1-\l$ from an original system state to an aggregate state at each step, rather than with certainty after $k$ steps as in Fig.\  \figsftm. Then it can be shown that the cost vector of a given stationary policy $\m$, may be evaluated approximately by $\Phi r_\m$, where $r_\m$ is the solution of the equation
$$r=DT_\m^{(\l)}(\Phi r),\xdef\polevalaggr{\lab}\eqnum\show{lbcontrtwo}$$
where $T_\m^{(\l)}$ is the mapping given by Eq.\ \tdlambdamap. This equation
has a unique solution because the mapping $DT_\m^{(\l)}\Phi$ can be shown to be a contraction mapping with respect to the sup-norm.  

As noted earlier, the aggregation equation 
$$\Phi r=\Phi DT_\m(\Phi r)$$
 is a projected equation because $\Phi D$ is a projection mapping with respect to a suitable weighted Euclidean seminorm (see [YuB12], Section 4; it is a norm projection in the case of hard aggregation). Similarly, the $\l$-aggregation equation 
 $$\Phi r=\Phi DT_\m^{(\l)}(\Phi r)$$
 is a projected equation, which is related to the proximal algorithm [Ber16a], [Ber18b], and may be solved by using temporal differences. Thus we may use exploration-enhanced versions of the LSTD($\l$) and LSPE($\l$) methods in an approximate PI scheme to solve the $\l$-aggregation equation. We refer to [Ber12] for further discussion.
 
\vskip-1.5pc

\section{Policy Iteration with Feature-Based Aggregation and a Neural Network}

\vskip-0.5pc

\xdef\pineural{\figr}\figrnum\show{myfigure}

\pn We noted in Section 3 that neural networks can be used to construct features at the output of the last nonlinear layer. The neural network training process also yields linear weighting parameters for the feature vector $F(i)$ at the output of the last layer, thus obtaining an approximation $\hat J_\m\big(F(i)\big)$ to the cost function of a given policy $\m$. Thus given the current policy $\m$, the typical PI produces the new policy $\hat \m$ using the approximate policy improvement operation \polimprove\ or a multistep variant, as illustrated in Fig.\ \pineural.

\xdef\piaggregate{\figr}\figrnum\show{myfigure}

A similar PI scheme can be constructed based  on feature-based aggregation with features supplied by the same neural network; see Fig.\ \piaggregate. The main idea is to replace the (approximate) policy improvement  operation with the solution of an aggregate problem, which provides the (approximately) improved policy $\hat \m$. This is a more complicated policy improvement operation, but computes the new policy $\hat \m$ based on a more accurate cost function approximation: one that is a nonlinear function of the features rather than linear. Moreover, {\it $\hat \m$ not only aspires to be an improved policy relative to $\m$, but also to be an optimal policy based on the aggregate problem\/}, an approximation itself of the original DP problem. In particular, suppose that the neural network approximates $J_\m$ perfectly. Then the scheme of Fig.\ \pineural\ will replicate a single step of the PI algorithm starting from $\m$, while the aggregation scheme of Fig.\ \piaggregate, with sufficient number of aggregate states, will produce a policy that is arbitrarily close to optimal. 

Let us now explain each of the steps of the aggregation-based PI procedure of Fig.\ \piaggregate, starting with the current policy $\m$.

\nitem{(a)} {\it Feature mapping construction\/}: We train the neural network using a training set of state-cost pairs that are generated using the current policy $\m$. This provides a feature vector $F(i)$ as described in Section 3.

\nitem{(b)} {\it Sampling to obtain the disaggregation sets\/}: We sample the state space, generating a subset of states $I\subset \{1,\ldots,n\}$. We partition the corresponding set of state-feature pairs 
$$\big\{(i,F(i))\mid i\in I\big\}$$
 into a collection of subsets $S_1,\ldots,S_q$. We then consider the aggregation framework with 
$S_1,\ldots,S_q$ as the aggregate states, and the corresponding aggregate problem as described in Section 4. The sampling to obtain the set of states $I$ may be combined with exploration to ensure that a sufficiently representative set of states is included. 

\topinsert
\centerline{\includegraphics[width=5.8in]{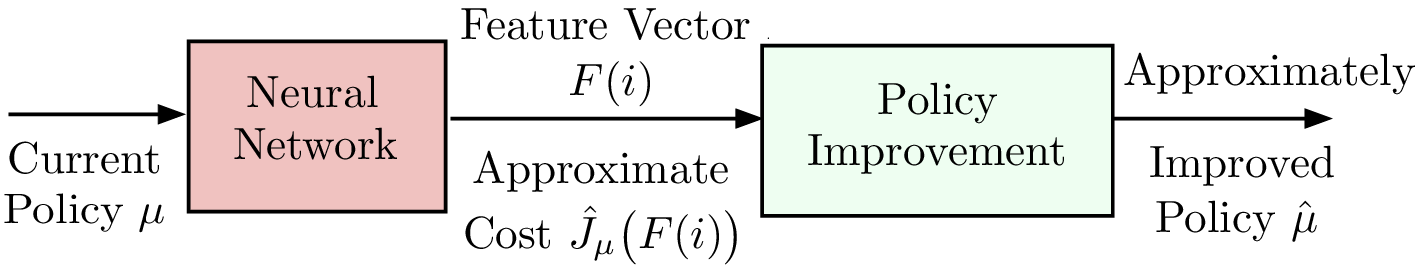}}
\vskip-0.0pc
\hskip-3pc\fig{0pc}{\pineural.} {Schematic illustration of PI using a neural network-based cost approximation. Starting with a training set of state-cost pairs generated using the current policy $\m$, the neural network yields a set of features and an approximate cost evaluation $\skew5 \hat J_\m$ using a linear combination of the features. This is followed by policy improvement using $\skew5 \hat J_\m$ to generate the new policy $\hat \m$.}\endinsert

 \topinsert
\centerline{\includegraphics[width=6.2in]{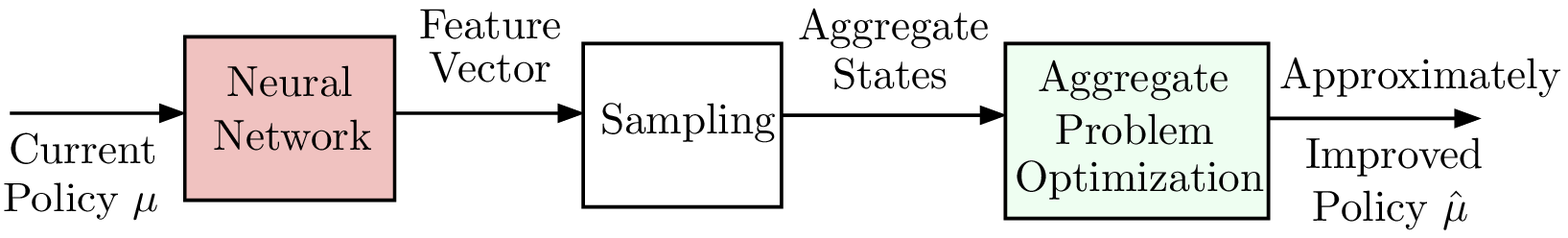}}
\vskip-0.0pc
\hskip-3pc\fig{0pc}{\piaggregate.} {Illustration of PI using feature-based aggregation with features supplied by a neural network. Starting with a training set of state-cost pairs generated using the current policy $\m$, the neural network yields a set of features, which are used to construct a feature-based aggregation framework. The optimal policy of the corresponding aggregate problem is used as the new policy $\hat \m$.}\endinsert

\nitem{(c)} {\it Aggregate problem solution\/}: The aggregate DP problem is solved by using a simulation-based method to yield  (perhaps approximately) the aggregate state optimal costs $r^*_\ell$, $\ell=1,\ldots,q$ (cf.\ Section 4.2). 
\nitem{(d)} {\it Definition of the improved policy\/}: The ``improved" policy is simply the optimal policy of the aggregate problem (or an approximation thereof, obtained for example after a few iterations of approximate simulation-based PI). This policy is defined implicitly by the aggregate costs $r^*_\ell$, $\ell=1,\ldots,q$,  using the one-step lookahead minimization
$$\hat \m(i)\in\arg\min_{u\in U(i)}\sum_{j=1}^n p_{ij}(u)\left(g(i,u,j)+\a \sum_{\ell=1}^q\phi_{j\ell}\,r^*_{\ell}\right),\qquad i=1,\ldots,n,$$
[cf.\ Eq.\ \belleqt] or a multistep lookahead variant. Alternatively, the ``improved" policy can implemented in model-free fashion using a $Q$-factor architecture $\tl Q(i,u,\theta)$, as described in Sections 2.4 and 4.1, cf.\ Eqs.\ \samplesaggr-\qoptpolaggr.

\smskip

Let us also note that there are several options for implementing the algorithmic ideas of this section. 

\nitem{(1)} The neural network-based feature construction process may be performed any number of times, each time followed by an aggregate problem solution that constructs a new policy, which is then used to generate new training data for the neural network. Alternatively, the neural network training and feature construction process may be done only once, followed by the solution of the corresponding feature-based aggregate problem. 

\nitem{(2)} Several deep neural network-based PI cycles may be performed, a subset of the features thus generated may be selected, and  the corresponding aggregate problem is solved just once, as a way of improving the final policy generated by the deep reinforcement learning process.

\nitem{(3)} Following each cycle of neural network-based feature evaluation, the generated features may be supplemented with additional problem-specific handcrafted features, and/or features from previous cycles. This is a form of feature iteration that was noted in the preceding section.

\smskip

Finally, let us mention a potential weakness of using the features obtained at the output of the last nonlinear layer of the neural network in the context of aggregation: the sheer number of these features may be so large that the resulting number of aggregate states may become excessive. To address this situation one may consider pruning some of the features, or reducing their number using some form of regression, at the potential loss of some approximation accuracy. In this connection let us also emphasize a point made earlier in connection with an advantage of deep (rather than shallow) neural networks: {\it because with each additional layer, the generated features tend to be more complex, their number at the output of the final nonlinear layer of the network can be made smaller as the number of layers increases\/}. An extreme case is to use the cost function approximation obtained at the output of the neural network as a single feature/scoring function, in the spirit of Section 4.3. 

\subsubsection{Using Neural Networks in Conjunction with Heuristics}

\pn We noted at the end of Section 4.3 another use of neural networks in conjunction with aggregation: somehow construct multiple policies, evaluate each of these policies using a neural network, and use the policy cost function evaluations as multiple scoring functions in a feature-based aggregation scheme. In Section 4.4, we elaborated on this idea for the case of the deterministic discrete optimization problem
$$\eqalign{&\hbox{minimize\ \ }G(u_1,\ldots,u_N)\cr &\hbox{subject to\ \
}(u_1,\ldots,u_N)\in U,\cr}$$
where $U$ is a finite set of feasible solutions and $G$ is a cost function [cf.\ Eq.\ \dcost]. We described the use of multiple heuristics to construct corresponding scoring functions. At any given $m$-solution, the  scoring function values are computed by running each of the heuristics. A potential time-saving alternative is to approximate these scoring functions using neural networks.

In particular, for each of the heuristics, we may train a separate neural network by using a training set consisting of pairs of $m$-solutions  and corresponding heuristic costs. In this way we can obtain approximate scoring functions  
$$\tl V_1(u_1,\ldots,u_m; \theta_1), \ldots,\tl V_s(u_1,\ldots,u_m; \theta_s),$$
 where $\theta_1, \ldots,\theta_s$ are the corresponding neural network weight vectors. We may then use the approximate scoring functions as features in place of the exact heuristic cost functions to construct an aggregate problem similar to the one described in Section 4.4. The solution of the aggregate problem can be used in turn to define a new policy, which may optionally be added to the current set of heuristics, as discussed earlier.

Note that a separate neural network is needed for each heuristic and stage, so assembling the training data together with the training itself can be quite time consuming. However, both the data collection and the training processes can benefit greatly from parallelization.

Finally, let us note that the approach of using a neural network to obtain approximate scoring functions may also be used in conjunction with a rollout scheme that uses a limited horizon. In such a scheme, starting from an $m$-solution, we may evaluate all possible subsequent $(m+1)$-solutions by running each of the $s$ heuristics up to a certain horizon depth of $d$ steps [rather than the full depth of $(N-m-1)$ steps], and then approximate the subsequent heuristic cost  [from stage $(m+1+d)$ to stage $N$] by using the neural network estimates.

\vskip-1.5pc

\section{Concluding Remarks}

\vskip-0.5pc
\pn We have surveyed some aspects of approximate PI methods with a focus on a new idea for policy improvement: feature-based aggregation that uses features provided by a neural network or a heuristic scheme, perhaps in combination with additional handcrafted features. We have argued that this type of policy improvement, while more time-consuming, may yield more effective policies, owing to the DP character of the aggregate problem and the use of a nonlinear feature-based architecture. The algorithmic idea of this paper seems to work well on small examples. However, tests with challenging problems are needed to fully evaluate its merits, particularly since solving the aggregate DP problem is more time-consuming than the standard one-step lookahead policy improvement scheme of Eq.\ \thrtenea\ or its multistep lookahead variants. 

In this paper we have focused on finite-state discounted Markov decision problems, but our approach clearly extends to other types of finite-state DP involving stochastic uncertainty, including finite horizon, stochastic shortest path, and semi-Markov decision problems. It is also worth considering extensions to infinite-state problems, including those arising in the context of continuous spaces optimal control, shortest path, and partially observed Markov decision problems. Generally, the construction of aggregation frameworks for continuous spaces problems is conceptually straightforward, and follows the pattern  discussed in this paper for finite-state problems. For example a hard aggregation scheme involves a partition of the continuous state space into a finite number of subsets/aggregate states, while a representative states scheme involves discretization of the continuous state space using a finite number of states. Note, however, that from a mathematical point of view, there may be a substantial issue of consistency, i.e., whether the solution of the aggregate problem ``converges" to the solution of the continuous spaces problem as the number of aggregate states increases. Part of the reason has to do with the fact that the Bellman equation of continuous spaces problems need not have a unique solution. The author's monograph [Ber18a], Sections 4.5 and 4.6, provides an analysis of this question for shortest path and optimal control problems with a continuous state space, and identifies classes of problems that are more amenable to approximate DP solution approaches. 

Finally, we note that the key issue of feature construction can be addressed in a number of ways. In this paper we have focused on the use of deep neural networks and heuristics for approximating the optimal cost function or the cost functions of policies. However, we may use instead any methodology that automatically constructs good features at reasonable computational cost.

\vskip-1.5pc

\section{References}
\vskip-0.5pc
\def\ref{\vskip1.pt\pn}

\def\refer{\ref}

\ref[ADB17] Arulkumaran, K., Deisenroth, M.\ P., Brundage, M. and Bharath, A.\ A., 2017. ``A Brief Survey of Deep Reinforcement Learning," arXiv preprint arXiv:1708.05866.

\ref [Abr90] Abramson, B., 1990.\ ``Expected-Outcome: A General Model of Static
Evaluation," IEEE Transactions on Pattern Analysis and Machine Intelligence,
Vol.\ 12, pp.\ 182-193.

\ref[BBD10] Busoniu, L., Babuska, R., De Schutter, B., and Ernst, D.,  2010.\ Reinforcement Learning and Dynamic Programming Using Function Approximators, CRC Press, N.\ Y.

\ref[BBS87] Bean, J.\ C., Birge, J.\ R., and Smith, R.\ L., 1987.\ ``Aggregation in Dynamic Programming," Operations Research, Vol.\ 35, pp.\ 215-220.

\ref[BBS95] Barto, A.\  G., Bradtke, S.\ J., and Singh, S.\ P., 1995.
``Real-Time Learning and Control Using Asynchronous Dynamic Programming," Artificial Intelligence, 
Vol.\ 72, pp.\ 81-138.

\ref[BPW12] Browne, C.,  Powley, E.,  Whitehouse, D.,  Lucas, L., Cowling, P.\ I.,  Rohlfshagen, P.,  Tavener, S.,  Perez, D.,  Samothrakis, S., and  Colton, S., 2012.\ ``A Survey of Monte Carlo Tree Search Methods," IEEE Trans.\ on Computational Intelligence and AI in Games, Vol.\ 4, pp.\ 1-43.

\ref[BSA83] Barto, A.\ G., Sutton, R.\ S., and Anderson, C.\ W., 1983.\ 
``Neuronlike Elements that Can Solve Difficult Learning Control
Problems,'' IEEE Trans.\ on Systems, Man, and Cybernetics,
Vol.\ 13, pp.\ 835-846.

\ref [BTW97] Bertsekas, D.\ P., Tsitsiklis, J.\ N., and Wu, C., 1997.\ ``Rollout Algorithms for
Combinatorial Optimization,'' Heuristics, Vol.\ 3, pp.\ 245-262.

\ref [BeC89] Bertsekas, D.\ P., and Castanon, D.\ A., 1989.\ 
``Adaptive  Aggregation Methods for Infinite Horizon Dynamic Programming," IEEE
Trans.\ on Aut.\  Control, Vol.\ AC-34, pp.\ 589-598.

\ref[BeC99] Bertsekas, D.\ P., and  Castanon, D.\ A., 1999.\ ``Rollout Algorithms for
Stochastic Scheduling Problems," Heuristics, Vol.\ 5, pp.\ 89-108. 

\ref [BeT91]  Bertsekas, D.\ P., and Tsitsiklis, J.\ N., 1991.\ ``An Analysis of
Stochastic Shortest Path Problems,"
Math.\ Operations Research, Vol.\ 16, pp.\ 580-595.

\ref [BeT96]  Bertsekas, D.\ P., and Tsitsiklis, J.\ N., 1996.\ Neuro-Dynamic
Programming, Athena Scientific, Belmont, MA.

\ref [BeT00]  Bertsekas, D.\ P., and Tsitsiklis, J.\ N., 2000.\ ``Gradient Convergence in Gradient
Methods," SIAM J.\ on Optimization,
Vol.\ 10, pp. 627-642.

\ref[Ber95] Bertsekas, D.\ P., 1995.\ ``A Counterexample to Temporal Differences
Learning,''  Neural Computation, Vol.\ 7, pp.\ 270-279.

\ref[Ber11a] Bertsekas, D.\ P., 2011.\
``Approximate Policy Iteration: A Survey and Some New Methods," J.\ of Control Theory and Applications, Vol.\ 9, pp.\ 310-335. 

\ref[Ber11b] Bertsekas, D.\ P., 2011.\
``Temporal Difference Methods for General Projected Equations," IEEE Trans.\ on Aut.\  Control, Vol.\ 56, pp.\ 2128-2139.

\ref[Ber11c] Bertsekas, D.\ P., 2011.\
``$\l$-Policy Iteration: A Review and a New Implementation,"  Lab.\ for Information and Decision Systems Report LIDS-P-2874, MIT; in {\it Reinforcement Learning and Approximate Dynamic Programming for Feedback Control\/}, by F.\ Lewis and D.\ Liu (eds.), IEEE Press, Computational Intelligence Series, 2012.

\ref[Ber12] Bertsekas, D.\ P., 2012.\ Dynamic Programming and Optimal Control, Vol.\ II: Approximate Dynamic Programming, 4th edition, Athena Scientific, Belmont, MA.

\ref[Ber13] Bertsekas, D.\ P., 2013.\ ``Rollout Algorithms for Discrete Optimization: A Survey," Handbook of Combinatorial Optimization, Springer.

\ref[Ber15] Bertsekas, D.\ P., 2015.\ Convex Optimization Algorithms, Athena Scientific, Belmont, MA.

\ref[Ber16a] Bertsekas, D.\ P., 2016.\ ``Proximal Algorithms and Temporal Differences for Large Linear Systems: Extrapolation, Approximation, and Simulation," Report LIDS-P-3205, MIT; arXiv preprint arXiv:1610.05427.

\ref[Ber16b] Bertsekas, D.\ P., 2016.\ Nonlinear Programming, 3rd edition, Athena Scientific, Belmont, MA.

\ref[Ber17] Bertsekas, D.\ P., 2017.\ Dynamic Programming and Optimal Control, Vol.\ I, 4th edition, Athena Scientific, Belmont, MA.

\ref[Ber18a] Bertsekas, D.\ P., 2018.\ Abstract Dynamic Programming, Athena Scientific, Belmont, MA.

\ref[Ber18b] Bertsekas, D.\ P., 2018.\ ``Proximal Algorithms and Temporal Difference Methods for Solving Fixed Point Problems," Computational Optimization and Applications J., Vol.\ 70, pp.\ 709-736.

\ref[Bis95] Bishop, C.\ M, 1995.\ Neural Networks for Pattern 
Recognition,
Oxford University Press, N.\ Y.

\ref [CFH05] Chang, H.\ S., Hu, J., Fu, M.\ C., and Marcus, S.\ I., 2005.\  ``An Adaptive Sampling Algorithm for Solving Markov Decision Processes," Operations Research, Vol.\ 53, pp. 126-139.

\ref [CFH13] Chang, H.\ S., Hu, J., Fu, M.\ C., and Marcus, S.\ I., 2013.\  Simulation-Based Algorithms for Markov Decision Processes, (2nd Ed.), Springer, N.\ Y.

\ref[Cao07] Cao, X.\ R., 2007.\ Stochastic Learning and Optimization: A Sensiti\-vity-Based Approach, Springer, N.\ Y.

\ref[ChK86] Christensen, J., and Korf, R.\ E., 1986.\
``A Unified Theory of Heuristic
Evaluation Functions and its Application to Learning,'' in Proceedings 
AAAI-86,  pp.\ 148-152.

\ref[ChM82] Chatelin, F., and Miranker, W.\ L., 1982.\  ``Acceleration by Aggregation of Successive Approximation Methods," Linear Algebra and its Applications, Vol.\ 43, pp.\ 17-47.

\ref[CiS15] Ciosek, K., and Silver, D., 2015.\ ``Value Iteration with Options and State Aggregation," Report, Centre for Computational Statistics and Machine Learning University College London.

\ref[Cou06] Coulom, R., 2006.\ ``Efficient Selectivity and Backup Operators in Monte-Carlo Tree Search," International Conference on Computers and Games, Springer, pp.\ 72-83.

\ref[Cyb89] Cybenko, 1989.\ ``Approximation by Superpositions of a
Sigmoidal Function," Math.\ of Control, Signals, and Systems, Vol.\ 2,
pp.\ 303-314.

\ref[DNW16] David, O.\ E., Netanyahu, N.\ S., and Wolf, L., 2016.\ ``Deepchess: End-to-End Deep Neural Network for Automatic Learning in Chess," in International Conference on Artificial Neural Networks, pp.\ 88-96, Springer.

\ref[DiM10] Di Castro, D., and Mannor, S., 2010.\ ``Adaptive Bases for Reinforcement Learning," Machine Learning and Knowledge Discovery in Databases, Vol.\ 6321, pp.\ 312-327.

\ref[Die00] Dietterich, T., 2000.\ ``Hierarchical Reinforcement Learning with the
MAXQ Value Function Decomposition," J.\ of Artificial Intelligence Research, Vol.\ 13, pp.\ 227-303.

\ref[FYG06] Fern, A., Yoon, S. and Givan, R., 2006.\ ``Approximate Policy Iteration with a Policy Language Bias: Solving Relational Markov Decision Processes," J.\ of Artificial Intelligence Research, Vol.\ 25, pp.\ 75-118.

\ref[DoD93] Douglas, C.\ C., and Douglas, J., 1993.\ ``A Unified Convergence Theory for Abstract Multigrid or Multilevel Algorithms, Serial and Parallel," SIAM J.\ Num.\ Anal., Vol.\ 30, pp.\ 136-158.

\ref[Fle84] Fletcher, C.\ A.\ J., 1984.\ Computational Galerkin Methods, Springer, N.\ Y.

\ref[Fun89] Funahashi, K., 1989.\ ``On the Approximate Realization of Continuous Mappings by
Neural Networks," Neural Networks, Vol.\ 2, pp.\ 183-192.

\ref[GBC16] Goodfellow, I., Bengio, J., and Courville, A., Deep Learning, MIT Press, Cambridge, MA.

\ref[GGS13] Gabillon, V., Ghavamzadeh, M., and Scherrer, B., 2013.\ ``Approximate Dynamic Programming Finally Performs Well in the Game of Tetris," in Advances in Neural Information Processing Systems, pp.\ 1754-1762.

\ref[Gor95] Gordon, G.\ J., 1995.\ ``Stable Function Approximation in  Dynamic Programming,''
in Machine Learning: Proceedings of the 12th International Conference, Morgan Kaufmann, San Francisco, CA.

\ref[Gos15] Gosavi, A., 2015.\ Simulation-Based Optimization:
Parametric Optimization Techniques and Reinforcement Learning, 2nd Edition, Springer, N.\ Y.

\ref[HMK98] Hauskrecht, M., Meuleau, N., Kaelbling, L.\ P., Dean, T., and Boutilier, C., 1998.\ ``Hierarchical Solution of Markov Decision Processes Using Macro-Actions," in Proceedings of the 14th Conference on Uncertainty in Artificial Intelligence, pp.\ 220-229.

\ref[HOT06]  Hinton, G.\ E., Osindero, S., and Teh, Y.\ W., 2006. ``A Fast Learning Algorithm for Deep Belief Nets," Neural Computation, Vol.\ 18, pp.\ 1527-1554.

\ref[HSW89] Hornick, K., Stinchcombe, M., and White, H., 1989.\ ``Multilayer Feedforward Networks
are Universal Approximators," Neural Networks, Vol.\ 2, pp.\ 359-159.

\ref[Hay08] Haykin, S., 2008.\ Neural Networks and Learning Machines, (3rd Edition), Prentice-Hall, Englewood-Cliffs, N.\ J.

\ref[Hol86] Holland, J.\ H., 1986.\ ``Escaping Brittleness: the Possibility
of General-Purpose Learning Algorithms Applied to Rule-Based Systems,"
in Machine Learning: An Artificial Intelligence Approach,
Michalski, R.\ S., Carbonell, J.\ G., and Mitchell, T.\ M., (eds.),
Morgan Kaufmann, San Mateo, CA, pp.\ 593-623.

\ref[Iva68] Ivakhnenko, A.\ G., 1968.\ ``The Group Method of Data Handling: A Rival of the Method of Stochastic Approximation," Soviet Automatic Control, Vol.\ 13, pp.\ 43-55.

\ref[Iva71] Ivakhnenko, A.\ G., 1971.\ ``Polynomial Theory of Complex Systems," IEEE Transactions on Systems, Man and Cybernetics, Vol.\ 4, pp.\ 364-378.

\ref[Jon90] Jones, L.\ K., 1990.\ ``Constructive Approximations for 
Neural Networks by Sigmoidal Functions," Proceedings of the 
IEEE, Vol.\ 78, pp.\ 1586-1589.

\ref[KMP06] Keller, P.\ W., Mannor, S., and Precup, D., 2006.\ ``Automatic Basis Function Construction for Approximate Dynamic Programming and Reinforcement Learning," in Proc.\ of the 23rd International Conference on Machine Learning, ACM, pp.\ 449-456.

\ref [KVZ72] Krasnoselskii, M.\ A.,  Vainikko, G.\ M., Zabreyko, R.\ P.,  and Ruticki, Ya.\ B., 1972.\ Approximate Solution of Operator Equations, Translated by D.\ Louvish, Wolters-Noordhoff Pub., Groningen.

\ref[Kir11] Kirsch, A., 2011.\ An Introduction to the Mathematical Theory of Inverse Problems, (2nd Edition), Springer, N.\ Y.

\ref[KoB09] Konidaris, G., and Barto, A., 2009.\ ``Efficient Skill Learning Using Abstraction Selection," in 21st International Joint Conference on Artificial Intelligence.

\ref[LLL08] Lewis, F.\ L., Liu, D., and Lendaris, G.\ G., 2008.\ Special Issue on Adaptive Dynamic Programming and Reinforcement Learning in Feedback Control, IEEE Trans.\ on Systems, Man, and Cybernetics, Part B, Vol.\ 38, No.\ 4.

\ref[LLP93] Leshno, M., Lin, V.\ Y., Pinkus, A., and Schocken, S., 1993.\ ``Multilayer Feedforward Networks with a Nonpolynomial Activation Function can Approximate any Function," Neural Networks, Vol.\ 6, pp.\ 861-867.

\ref[LWL17] Liu, W., Wang, Z., Liu, X., Zeng, N., Liu, Y., and Alsaadi, F.\ E., 2017. ``A Survey of Deep Neural Network Architectures and their Applications," Neurocomputing, Vol.\ 234, pp.\ 11-26.

\ref[LeL12] Lewis, F.\ L., and Liu, D., 2012.\ Reinforcement Learning and Approximate Dynamic Programming for Feedback Control, IEEE Press Computational Intelligence Series, N.\ Y. 

\ref[Li17] Li, Y., 2017.\ ``Deep Reinforcement Learning: An Overview," arXiv preprint ArXiv: 1701.07274v5.

\ref[MMP15] Mann, T.A., Mannor, S. and Precup, D., 2015.\ ``Approximate Value Iteration with Temporally Extended Actions," J.\ of Artificial Intelligence Research, Vol.\ 53, pp.\ 375-438.

\ref [MMS06] Menache, I., Mannor, S., and Shimkin, N., 2005.\ ``Basis Function Adaptation in Temporal Difference Reinforcement Learning," Ann.\ Oper.\ Res., Vol. 134, pp.\ 215-238.

\ref[Men82] Mendelssohn, R., 1982.\ ``An Iterative Aggregation Procedure for Markov Decision Processes," Operations Research, Vol.\ 30, pp.\ 62-73.

\ref[PaR98] Parr, R., and Russell, S.\ J., 1998.\ ``Reinforcement Learning with Hierarchies of Machines," in Advances in Neural Information Processing Systems, pp.\ 1043-1049.

\ref [Pow11] Powell, W.\ B., 2011.\  Approximate Dynamic Programming: Solving the Curses of Dimensionality, 2nd Edition, J.\ Wiley and Sons, Hoboken, N.\ J.

\ref[RPW91] Rogers, D.\ F., Plante, R.\ D., Wong, R.\ T.,  and Evans, J.\ R., 1991.\ ``Aggregation and Disaggregation Techniques and Methodology in Optimization," Operations Research, 
Vol.\ 39, pp.\ 553-582.

\ref[SBP04] Si, J., Barto, A., Powell, W., and Wunsch, D., (Eds.) 2004.\ Learning and Approximate Dynamic
Programming, IEEE Press, N.\ Y.

\ref[SGG15] Scherrer, B., Ghavamzadeh, M., Gabillon, V., Lesner, B., and Geist, M., 2015.\ ``Approximate Modified Policy Iteration and its Application to the Game of Tetris," J.\ of Machine Learning Research, Vol.\ 16, pp.\ 1629-1676.

\ref[SHS17] Silver, D., Hubert, T., Schrittwieser, J., Antonoglou, I., Lai, M., Guez, A., Lanctot, M., Sifre, L., Kumaran, D., Graepel, T. and Lillicrap, T., 2017.\ ``Mastering Chess and Shogi by Self-Play with a General Reinforcement Learning Algorithm," arXiv preprint arXiv:1712.01815.

\ref[SJJ95] Singh, S.\ P., Jaakkola, T., and Jordan, M.\ I., 1995.\
``Reinforcement Learning with Soft State Aggregation,'' in Advances in Neural
Information Processing Systems 7,  MIT Press, Cambridge, MA.

\ref[SPS99] Sutton, R., Precup, D., and Singh, S., 1999.\ ``Between MDPs and Semi-MDPs: A Framework for Temporal Abstraction in Reinforcement Learning," Artificial Intelligence, Vol.\ 112, pp.\ 181-211.

\ref[SSP18] Serban, I.\ V., Sankar, C., Pieper, M., Pineau, J., Bengio., J., 2018.\ ``The Bottleneck Simulator:
A Model-based Deep Reinforcement Learning Approach," arXiv preprint arXiv:1807.04723.v1. 

\ref[Saa03] Saad, Y., 2003.\ Iterative Methods for Sparse Linear Systems, SIAM, Phila., Pa.

\ref[Sam59] Samuel, A.\ L., 1959.\ ``Some Studies in Machine Learning
Using the Game of Checkers,''  IBM J.\ of Research and Development,
pp.\ 210-229.

\ref[Sam67] Samuel, A.\ L., 1967.\ ``Some Studies in Machine Learning
Using the Game of Checkers.\ II -- Recent Progress,'' 
IBM J.\ of Research and Development,
pp.\ 601-617.

\ref[Sch13] Scherrer, B., 2013.\ ``Performance Bounds for Lambda Policy Iteration and Application 
to the Game of Tetris," J.\ of Machine Learning Research, Vol.\ 14, pp.\ 1181-1227.

\ref[Sch15] Schmidhuber, J., 2015.\ ``Deep Learning in Neural Networks: An Overview," Neural Networks, Vol.\ 61, pp.\ 85-117.

\ref [Sha50] Shannon, C., 1950.\  ``Programming a Digital Computer for Playing
Chess," Phil.\ Mag., Vol.\ 41, pp.\ 356-375.

\ref[SuB98] Sutton, R.\  S., and Barto, A.\ G., 1998.\ Reinforcement Learning, MIT
Press, Cambridge, MA. (A draft 2nd edition is available on-line.)

\ref[Sut88] Sutton, R.\  S., 1988.\ ``Learning to Predict by the Methods of
Temporal Differences," Machine Learning, Vol.\ 3, pp.\ 9-44.

\ref [Sze10] Szepesvari, C., 2010.\ Algorithms for Reinforcement Learning, Morgan and Claypool Publishers, San Franscisco, CA.

\ref[TeG96] Tesauro, G., and Galperin, G.\ R., 1996.\ ``On-Line Policy Improvement
Using Monte Carlo Search,'' presented at the 1996 Neural Information
Processing Systems Conference, Denver, CO; also in M. Mozer et al.\ (eds.), Advances in
Neural Information Processing Systems 9, MIT Press (1997).

\ref[Tes89a] Tesauro, G.\ J., 1989. ``Neurogammon Wins Computer
Olympiad,'' Neural Computation, Vol.\ 1, pp.\ 321-323.

\ref[Tes89b] Tesauro, G.\ J.,  1989.\ ``Connectionist Learning of Expert Preferences by Comparison Training," in Advances in Neural Information Processing Systems, pp.\ 99-106.

\ref[Tes92] Tesauro, G.\ J., 1992.\ ``Practical Issues in Temporal 
Difference Learning,"
Machine Learning, Vol.\ 8, pp.\ 257-277.

\ref[Tes94] Tesauro, G.\ J., 1994.\ ``TD-Gammon, a Self-Teaching 
Backgammon
Program, Achieves Master-Level Play,'' Neural Computation, Vol.\ 6, pp.\ 
215-219.

\ref[Tes95] Tesauro, G.\ J., 1995.\ 
``Temporal Difference Learning and TD-Gammon,'' Communications of the 
ACM,
Vol.\ 38, pp.\ 58-68.

\ref[Tes01] Tesauro, G.\ J., 2001.\ ``Comparison Training of Chess Evaluation Functions," in Machines that Learn to Play Games, Nova Science Publishers,  pp.\ 117-130.

\ref[Tes02] Tesauro, G.\ J., 2002.\ ``Programming Backgammon Using Self-Teaching Neural Nets," Artificial Intelligence, Vol.\ 134, pp.\ 181-199.

\ref[TsV96] Tsitsiklis, J.\ N., and Van Roy, B., 1996. ``Feature-Based Methods for
Large-Scale Dynamic Programming," Machine Learning, Vol.\ 22, pp.\ 59-94.

\ref[Tsi94] Tsitsiklis, J.\ N., 1994.\ ``Asynchronous Stochastic Approximation and
Q-Learning," Machine Learning, Vol.\ 16, pp.\ 185-202.

\ref[VVL13] Vrabie, V., Vamvoudakis, K.\ G., and Lewis, F.\ L., 2013.\ Optimal Adaptive Control and Differential Games by Reinforcement Learning Principles, The Institution of Engineering and Technology, London.

\ref[Van06] Van Roy, B., 2006.\ ``Performance Loss Bounds for Approximate Value Iteration with State Aggregation," Mathematics of Operations Research, Vol.\ 31, pp.\ 234-244.

\ref[Wer77] Werb\"{o}s, P.\ J., 1977.\ ``Advanced Forecasting Methods
for Global Crisis Warning and Models of Intelligence,''
General Systems Yearbook, Vol.\ 22, pp.\ 25-38.

\ref[YuB04] Yu, H., and Bertsekas, D.\ P., 2004.\ ``Discretized Approximations for POMDP with Average Cost," Proc.\ of the 20th Conference
on Uncertainty in Artificial Intelligence, Banff, Canada.

\ref[YuB09] Yu, H., and Bertsekas, D.\ P., 2009.\ ``Basis Function Adaptation Methods for Cost Approximation in MDP," Proceedings of 2009 IEEE Symposium on Approximate Dynamic Programming and Reinforcement Learning (ADPRL 2009), Nashville, Tenn. 

\ref[YuB10] Yu, H., and Bertsekas, D.\ P., 2010.\ ``Error Bounds for Approximations from Projected Linear Equations," Mathematics of Operations Research, Vol.\ 35, pp.\ 306-329.

\ref[YuB12] Yu, H., and Bertsekas, D.\ P., 2012.\ ``Weighted Bellman Equations and their Applications in Dynamic Programming," Lab.\ for Information and Decision Systems Report LIDS-P-2876, MIT. 

\end